\documentclass[10pt,twocolumn]{article}
\usepackage[letterpaper,margin=0.72in,columnsep=0.27in]{geometry}
\usepackage[T1]{fontenc}
\usepackage{lmodern}
\usepackage[expansion=false]{microtype}
\usepackage{amsmath,amssymb}
\usepackage{graphicx}
\usepackage{booktabs}
\usepackage{array}
\usepackage{tabularx}
\usepackage{multirow}
\usepackage[table]{xcolor}
\usepackage{makecell}
\usepackage{caption}
\usepackage{enumitem}
\usepackage[round,sort]{natbib}
\usepackage[colorlinks=true,linkcolor=blue!50!black,citecolor=blue!50!black,urlcolor=blue!50!black]{hyperref}
\usepackage{url}
\hypersetup{pdftitle={HubEgo-100K: In-the-Wild Egocentric Video of Skilled Work for Robot Learning},pdfauthor={Javad Amirian, Armin Kiani, Loic Schaeffer, Tim Sprecher, Mohamed Taha, Karthik Yadati},pdfsubject={Dataset paper},pdfkeywords={egocentric video, robot learning, dataset, skilled work, imitation learning}}

\setlist{nosep,leftmargin=*}
\newcommand{\hub}{HubEgo-100K}
\newcolumntype{Y}{>{\raggedright\arraybackslash}X}
\newcolumntype{P}[1]{>{\raggedright\arraybackslash}p{#1}}
\newcolumntype{R}[1]{>{\raggedleft\arraybackslash}p{#1}}
\newcommand{\thinrule}{\arrayrulecolor{black!25}\hline}
\newcommand{\thickrule}{\arrayrulecolor{black}\Xhline{0.8pt}\arrayrulecolor{black!25}}
\newcommand{\headrow}{\rowcolor{black!6}}
\newcommand{\headrule}{\thinrule}
\definecolor{hubrow}{HTML}{E8ECFB}
\newcolumntype{I}{!{\color{black!25}\vrule width 0.4pt}}
\newcommand{\dsq}[1]{\par\vspace{3.5pt}\noindent\textbf{#1}\par\nobreak\vspace{1pt}\noindent\ignorespaces}
\title{\textbf{HubEgo-100K: In-the-Wild Egocentric Video\\ of Skilled Work for Robot Learning}}
\author{
\begin{tabular}{@{}*{3}{p{0.24\textwidth}<{\centering}}@{}}
Javad Amirian & Armin Kiani & Lo\"ic Schaeffer\\[4pt]
Tim Sprecher & Mohamed Taha & Karthik Yadati\\[12pt]
\multicolumn{3}{c}{\normalsize Hub Labs, San Francisco}\\[2pt]
\multicolumn{3}{c}{\small \href{https://hub.xyz}{hub.xyz}}
\end{tabular}}
\date{}
\begin{document}
\onecolumn
\maketitle
\vspace{-1.2em}
\vspace{0pt plus 0.5fill}
{\centering\captionsetup{type=figure}%
\includegraphics[width=\textwidth]{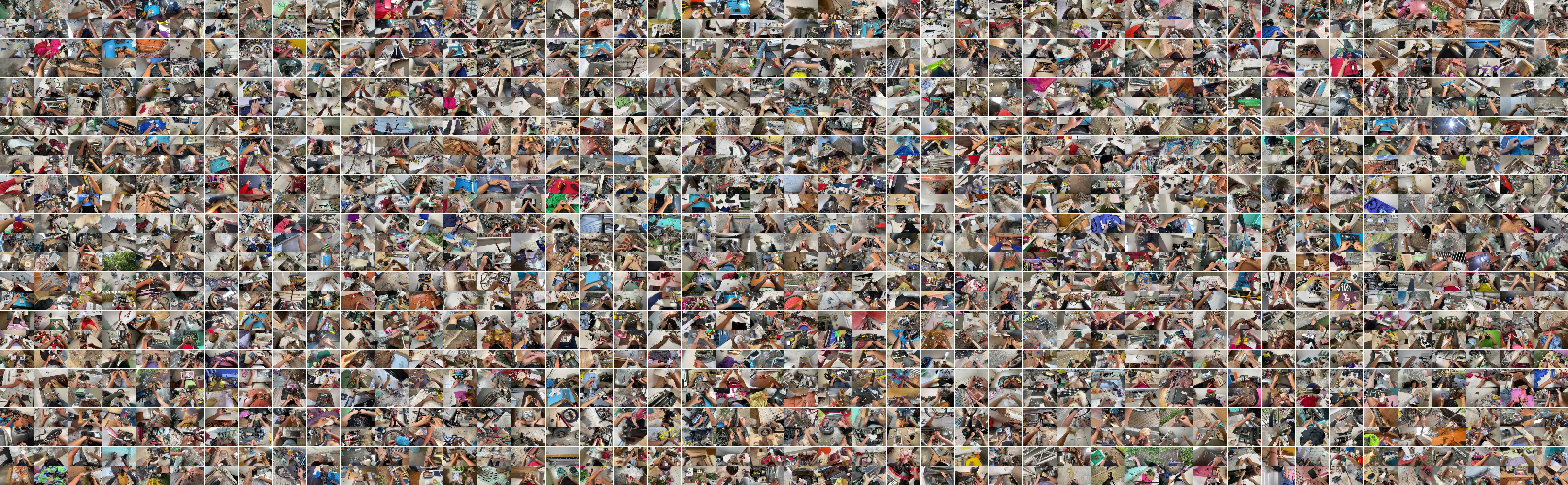}\par
\caption{Diversity in \hub{} v1.0: one frame each from 1,150 of the 12,227 contributors, in their own workplaces and homes, across all 30 work categories.}
\label{fig:teaser}
\par}
\vspace{15pt plus 1fill minus 5pt}
{\centering
\begin{minipage}[b]{0.86\textwidth}
{\centering\bfseries\Large Abstract\par}\vspace{0.9em}
\fontsize{11}{13.7}\selectfont\itshape Robot policies trained by imitation generalize poorly outside the environments and objects seen in training. Published data-scaling studies find that generalization grows with the diversity of training environments and objects rather than with more demonstrations per environment. Robot data is costly to collect across many environments, whereas egocentric human video can be recorded in the wild at scale; existing egocentric video, however, mostly shows daily-life activity and contains little skilled work in real workplaces. We present \hub{}, a dataset built for diversity. This version of the paper covers the first release, v1.0: 10,124 hours of 1080p egocentric video with synchronized 100 Hz inertial (IMU) data from one head-mounted phone camera. In v1.0, 12,227 contributors record their own unscripted work in 15,548 environments (each one contributor's own setting of one type, such as their kitchen or workshop) across 11 setting types, 30 work categories, 56 task types and 35 action families, with 83\% of hours in workplace settings. The release comprises 72,754 episodes with machine-generated action and language annotations, in LeRobot and MCAP formats under CC BY 4.0. We measure diversity along each axis, show that inside v1.0, with episodes in random order, new contributors and words keep appearing, and relate the measurements to published evidence on six data properties tied to policy performance, including hand visibility and inertial sensing. We also document collection, curation, privacy measures and data format, and provide a datasheet. Later versions of this paper will add benchmarks and cover further releases toward 100,000 hours.
\end{minipage}\par}

\twocolumn[{%
\centering\captionsetup{type=figure}%
\includegraphics[width=\textwidth]{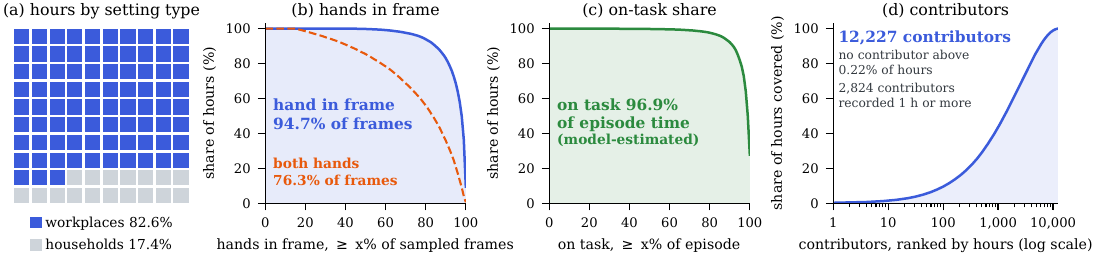}
\caption{Settings, hands, task focus and contributors in \hub{} v1.0. (a) Hours by setting type: each square is 1\% of hours; 82.6\% of hours are recorded in workplace settings, which include trades done in homes (26.9\% of hours), and 17.4\% in households. (b) Model-estimated hand visibility: share of hours in episodes in which at least one hand (solid) or both hands (dashed) appear in at least $x$\% of the sampled frames; over all hours, at least one hand appears in 94.7\% of sampled frames and both hands in 76.3\%. (c) Model-estimated on-task share: share of hours in episodes that are on task for at least $x$\% of the episode; the hours-weighted on-task share is 96.9\%. (d) Share of hours covered by contributors ranked by hours (log scale): of the 12,227 contributors, none accounts for more than 0.22\% of hours, and 2,824 contributors recorded one hour or more.}
\label{fig:glance}
\par\vspace{\dbltextfloatsep}
}]

\section{Introduction}
\label{sec:intro}

Robot manipulation policies trained by imitation generalize poorly outside the environments and objects seen in training \citep{kim2024openvla,khazatsky2024droid}. Published data-scaling studies find that generalization grows with the diversity of the training data rather than with its volume alone. \citet{lin2025scaling} report that generalization follows a power law in the number of training environments and objects, while additional demonstrations per environment saturate; experiments with DROID \citep{khazatsky2024droid} and $\pi_{0.5}$ \citep{black2025pi05} show the same dependence on scene diversity (Section~\ref{sec:why-env}). Robot data, however, is costly to collect across many environments. DROID gathered 350 hours across 564 scenes over 12 months with 50 data collectors \citep{khazatsky2024droid}, and generalist policies are still trained on pooled robot data: 970k demonstrations from Open X-Embodiment for OpenVLA \citep{kim2024openvla}, and data from 7 robot configurations and 68 tasks together with Open X-Embodiment for $\pi_0$ \citep{black2025pi0}.

Egocentric (first-person) human video can be recorded at scale without robot hardware \citep{hoque2026egodex}, and robot foundation models have begun to use it: GR00T N1 trains on human egocentric video alongside synthetic and real-robot data, and learns from this action-free video through latent actions \citep{nvidia2025gr00t}. Egocentric human video improves visual representations for manipulation \citep{nair2022r3m,radosavovic2022mvp,majumdar2023vc1} and, when co-trained with a small amount of robot data, improves policy generalization to new scenes, objects and tasks \citep{kareer2025egomimic,qiu2025humanoid,tao2025dexwild,pi2025humantransfer}.

Existing egocentric datasets were mostly collected to study daily-life activity \citep{grauman2022ego4d}, kitchens \citep{damen2022epic100,li2018egtea}, or scripted and instrumented tasks, mostly in lab or tabletop settings \citep{sener2022assembly101,liu2022hoi4d,wang2023holoassist,hoque2026egodex}. Skilled manual work in real workplaces, such as bricklaying, wiring, plumbing, garment sewing, microsoldering, hair cutting or engine repair, involves tools, materials and multi-step procedures, and is not the primary focus of these collections. Ego-Exo4D \citep{grauman2024egoexo4d} targets skilled activity explicitly, with a multi-view capture rig in 123 scene contexts. Egocentric-10K and Egocentric-100K \citep{buildai2025egocentric10k,buildai2025egocentric100k} release head-mounted video of factory and manual labor at 10,000 and 100,405 hours. \hub{} spans 30 work categories, of which manufacturing and assembly is one (5.4\% of hours), and pairs every episode with per-frame data from the phone's inertial measurement unit (IMU) and with task types and machine-generated action and language annotations.

\begin{table*}[tp]
\centering
\footnotesize
\caption{Egocentric and manipulation datasets, with the diversity each source reports: where the data was recorded (Setting) and how many people recorded it (People: participants, camera wearers, data collectors or contributors, as each source counts them). n/a: not reported. Hours as reported by each source. Places are counted as each source reports them (kitchens, locations, scenes, rooms); \hub{} counts 15,548 environments, each one contributor's own setting of one type (their kitchen, their workshop), a close stand-in for a count of sites (contributors who share a workplace count once each).}
\label{tab:positioning}
\renewcommand{\arraystretch}{1.1}
\begin{tabularx}{\textwidth}{P{3.35cm}IR{1.3cm}IP{2.3cm}IYIYIR{1.3cm}IP{1.75cm}}
\thickrule
\headrow \textbf{Dataset} & \textbf{Hours} & \textbf{Viewpoint} & \textbf{Setting} & \textbf{Modalities} & \textbf{People} & \textbf{License} \\
\headrule
Ego4D \citep{grauman2022ego4d} & 3,670 & head-mounted (7 camera types) & daily life; 74 locations, 9 countries & RGB; audio, IMU (836 h), gaze, 3D scans, stereo on subsets & 931$^{d}$ & Ego4D license agreement \\
\thinrule
Ego-Exo4D \citep{grauman2024egoexo4d} & 1,286 (221 ego)$^{a}$ & Aria glasses + 4 to 5 static cameras & skilled activities; 123 scenes, 13 cities & ego and exo RGB, audio, IMU, gaze, 3D, poses, language & 740 & Ego-Exo4D license agreement \\
\thinrule
EPIC-KITCHENS-100 \citep{damen2022epic100} & 100 & head-mounted & participants' own kitchens (45) & RGB, audio, narrations & 37 & CC BY-NC 4.0 \\
\thinrule
EGTEA Gaze+ \citep{li2018egtea} & 28 & eye-tracking glasses & one kitchen, 7 recipes & RGB, audio, gaze, hand masks & 32 & n/a \\
\thinrule
Assembly101 \citep{sener2022assembly101} & 513 (167 ego)$^{a}$ & 8 static + 4 headset (monochrome) & lab desk, toy assembly & RGB, monochrome ego, 3D hand pose & 53 & CC BY-NC 4.0 \\
\thinrule
HOI4D \citep{liu2022hoi4d} & n/a (2.4M frames) & helmet-mounted RGB-D & 610 indoor rooms & RGB-D, 3D hand and object pose, segmentation & 9 & n/a \\
\thinrule
HoloAssist \citep{wang2023holoassist} & 166$^{b}$ & HoloLens 2 & 20 guided object-centric tasks & RGB, depth, head and hand pose, gaze, audio, IMU & 222 & CDLA-Permissive-2.0 \\
\thinrule
EgoDex \citep{hoque2026egodex} & 829 & Apple Vision Pro & tabletop, 194 tasks & RGB, 3D hand, finger and upper-body pose & n/a & CC BY-NC-ND \\
\thinrule
EgoVerse \citep{egoverse2026} & 1,362$^{c}$ & Aria glasses, custom rigs, phones & lab and in the wild; 240 scenes & RGB, camera and hand pose, language & 2,087 & n/a \\
\thinrule
Egocentric-10K \citep{buildai2025egocentric10k} & 10,000 & head-mounted, monocular & factories & RGB (1080p) & n/a$^{e}$ & Apache 2.0 \\
\thinrule
Egocentric-100K \citep{buildai2025egocentric100k} & 100,405 & head-mounted fisheye & manual labor & RGB ($456\times256$) & n/a$^{e}$ & Apache 2.0 \\
\thinrule
DROID \citep{khazatsky2024droid} & 350 & robot: 2 external + wrist stereo & 564 scenes, 52 buildings & stereo RGB, proprioception, actions, language & 50 collectors & CC BY 4.0 \\
\arrayrulecolor{black!60}\hline\arrayrulecolor{black!25}
\rowcolor{hubrow} \textbf{\hub{} (v1.0)} & 10,124 & head-mounted phone, monocular & skilled work in the wild; 11 setting types, 30 work categories; 83\% of hours in workplace settings & RGB (1080p), IMU; magnetometer and barometer for most episodes & 12,227 & CC BY 4.0 \\
\thickrule
\end{tabularx}
\\[2pt]
{\footnotesize $^{a}$ All views combined; egocentric-only hours in parentheses. $^{b}$ The project website states 169 hours. $^{c}$ Number from the paper; the project website reports 4,003 hours for its current release. $^{d}$ Number from the paper; the project website states 923. $^{e}$ The dataset cards state a mean of 4.68 and 7.06 hours per worker, respectively, but no worker count.}
\end{table*}

\hub{} is an egocentric video dataset of skilled work recorded in the wild, and it is built for diversity (Fig.~\ref{fig:teaser}). Contributors record a named, unscripted task with a head-mounted phone during their own work, in their own workplace or home, with their own tools and materials. By skilled work we mean trade, craft and service tasks done as an occupation. This paper describes the first release, v1.0 (10,124 hours of 1080p video with synchronized inertial data from one head-mounted phone camera), and measures its diversity axis by axis (Fig.~\ref{fig:diversity}, Table~\ref{tab:div}): 12,227 contributors; 15,548 environments across 11 setting types, with 83\% of hours in workplace settings and 17\% in households; 30 work categories and 56 task types; 35 action families; machine-generated descriptions that use 1,429 verb and 3,709 noun lemmas; and 56 phone models. An environment is one contributor's own setting of one type (their kitchen, their workshop), so environments of the same type are separate physical places recorded by different contributors, and their count is a close stand-in for the number of sites (contributors who share a workplace count once each). Hours are spread along these axes: no contributor accounts for more than 0.22\% of hours. Inside v1.0, with episodes in random order, new contributors and words keep appearing: the last 10\% of hours adds 356 contributors and about 84 noun lemmas. The footage is curated for manipulation: by model estimate, at least one hand appears in 94.7\% of sampled frames and both hands in 76.3\%, and the model-estimated on-task share is 96.9\% (Fig.~\ref{fig:glance}). Subsequent releases will extend the dataset to 100,000 hours. We make the following contributions.
\begin{itemize}
\item \textbf{Diversity along measured axes.} An in-the-wild dataset of unscripted skilled work whose diversity is measured on every axis: contributors (12,227), environments (15,548), setting types (11), work categories (30), task types (56) and action families (35). For each axis we report the count and how evenly hours are spread over it; for contributors and words we also report how new ones keep appearing inside v1.0 when episodes are taken in random order (Section~\ref{sec:diversity}).
\item \textbf{Dataset.} \hub{} v1.0: 10,124 hours of 1080p egocentric video in 72,754 episodes, with per-frame inertial data and machine-generated action, difficulty and language annotations per episode, under CC BY 4.0 (Section~\ref{sec:dataset}).
\item \textbf{Evidence review tied to measurements.} A review of published findings on which data properties matter for robot learning from human video, with each property mapped to a quantity we measure on \hub{} (Section~\ref{sec:why}, Table~\ref{tab:evidence}).
\item \textbf{Quantitative analysis.} The spread of hours over contributors, coverage of action families across work categories, the long tail of the description vocabulary and the difficulty mix (Section~\ref{sec:analysis}). Every statistic is computed from the released catalog and annotation tables and can be recomputed without downloading video.
\item \textbf{Format and documentation.} A LeRobot v3 layout with native MCAP sensor streams and per-episode catalog and annotation tables (Section~\ref{sec:format}), a datasheet (Appendix~\ref{app:datasheet}) and versioned releases toward 100,000 hours (Section~\ref{sec:release}).
\end{itemize}

\section{Related Work}
\label{sec:related}

\paragraph{Egocentric video datasets.} EPIC-KITCHENS \citep{damen2018epic,damen2022epic100} recorded unscripted activity in participants' own kitchens with head-mounted cameras and introduced verb and noun classes for actions. EGTEA Gaze+ \citep{li2018egtea} paired cooking video with gaze. Ego4D \citep{grauman2022ego4d} collected 3,670 hours of daily-life video from 931 camera wearers in 74 locations and defined benchmarks over episodic memory, hands and objects, and forecasting. Ego-Exo4D \citep{grauman2024egoexo4d} captured skilled activities from synchronized first- and third-person views. Assembly101 \citep{sener2022assembly101}, HOI4D \citep{liu2022hoi4d} and HoloAssist \citep{wang2023holoassist} focus on procedural assembly, category-level hand-object interaction and guided tasks, respectively, with 3D hand annotations. EgoDex \citep{hoque2026egodex} recorded 829 hours of tabletop manipulation with Apple Vision Pro hand tracking. EgoVerse \citep{egoverse2026} pools egocentric demonstrations from academic and industry groups for robot learning, and Egocentric-10K and Egocentric-100K \citep{buildai2025egocentric10k,buildai2025egocentric100k} release head-mounted video of manual labor.

\begin{figure*}[t]
\centering
\setlength{\tabcolsep}{1.5pt}\renewcommand{\arraystretch}{0.9}\footnotesize
\begin{tabular}{@{}*{6}{>{\centering\arraybackslash}p{0.161\textwidth}}@{}}
\includegraphics[width=\linewidth]{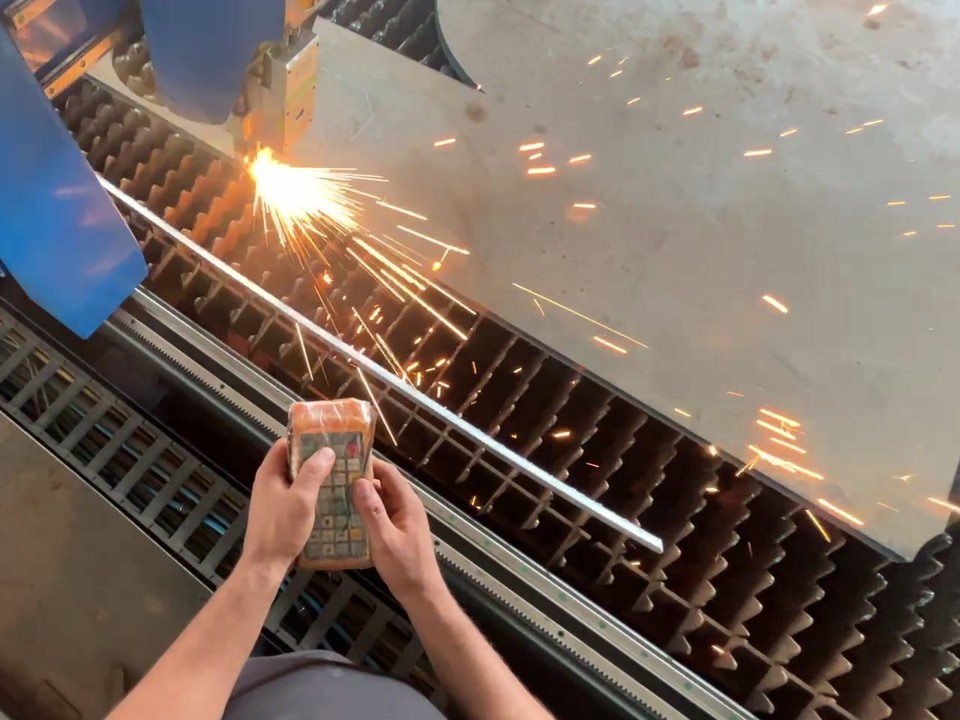} & \includegraphics[width=\linewidth]{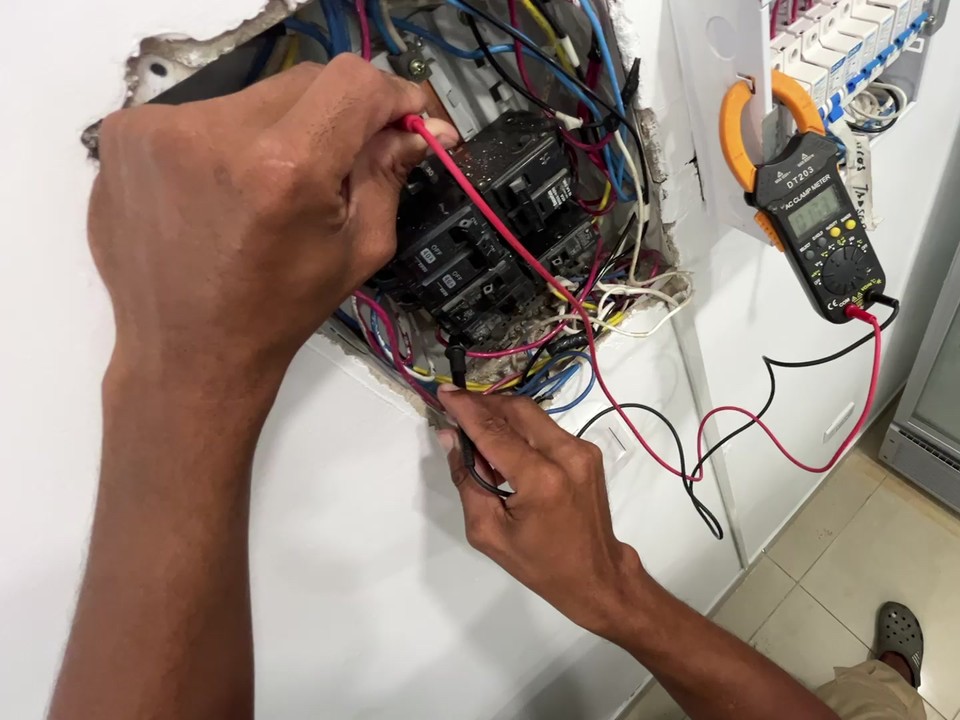} & \includegraphics[width=\linewidth]{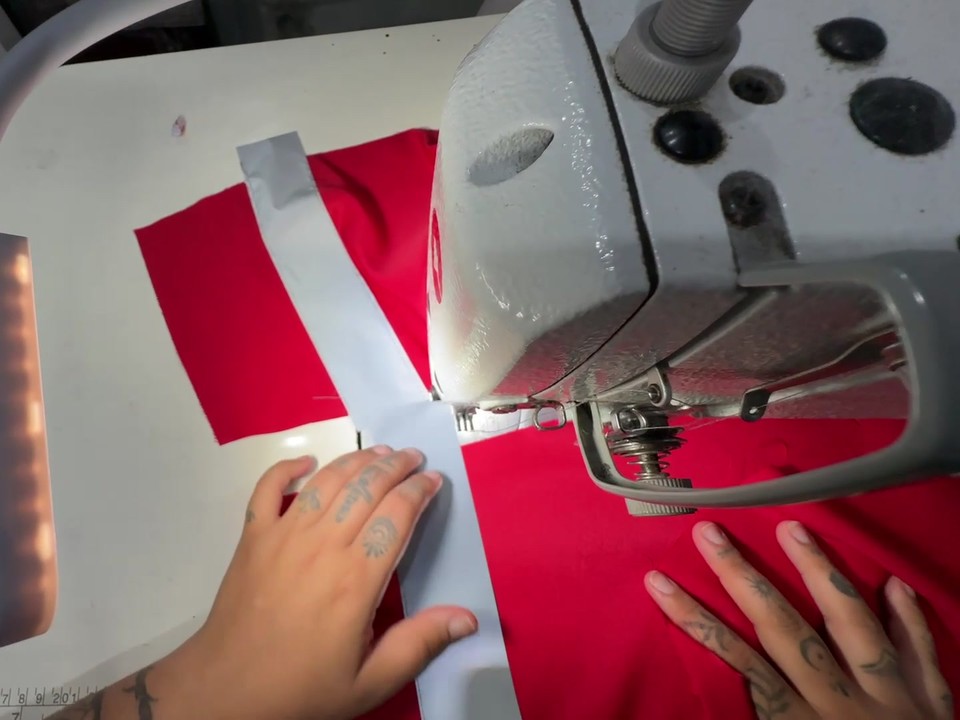} & \includegraphics[width=\linewidth]{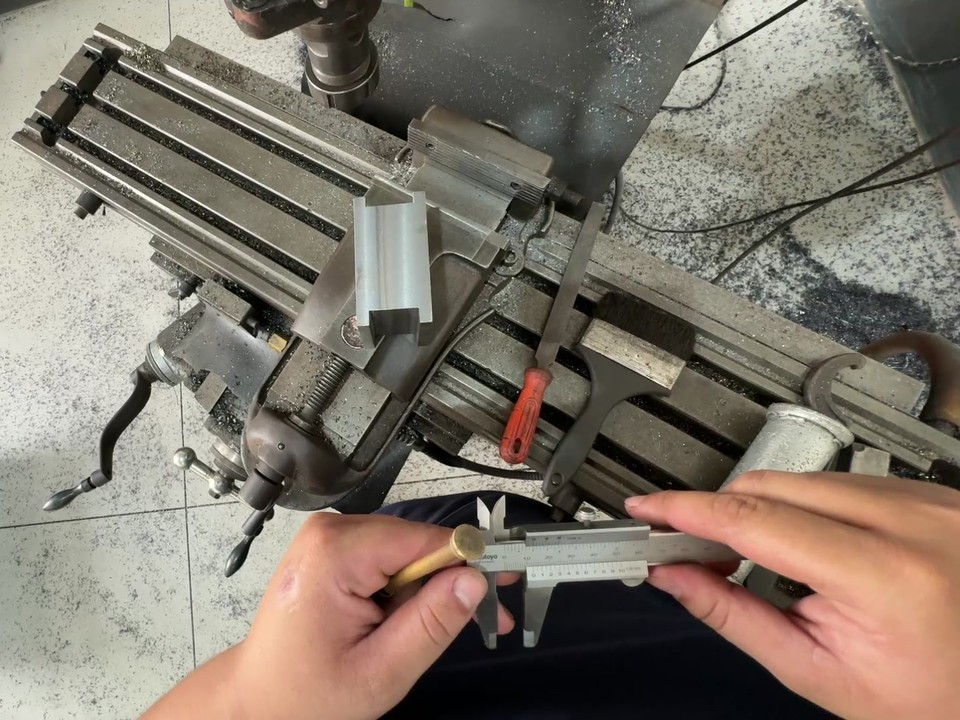} & \includegraphics[width=\linewidth]{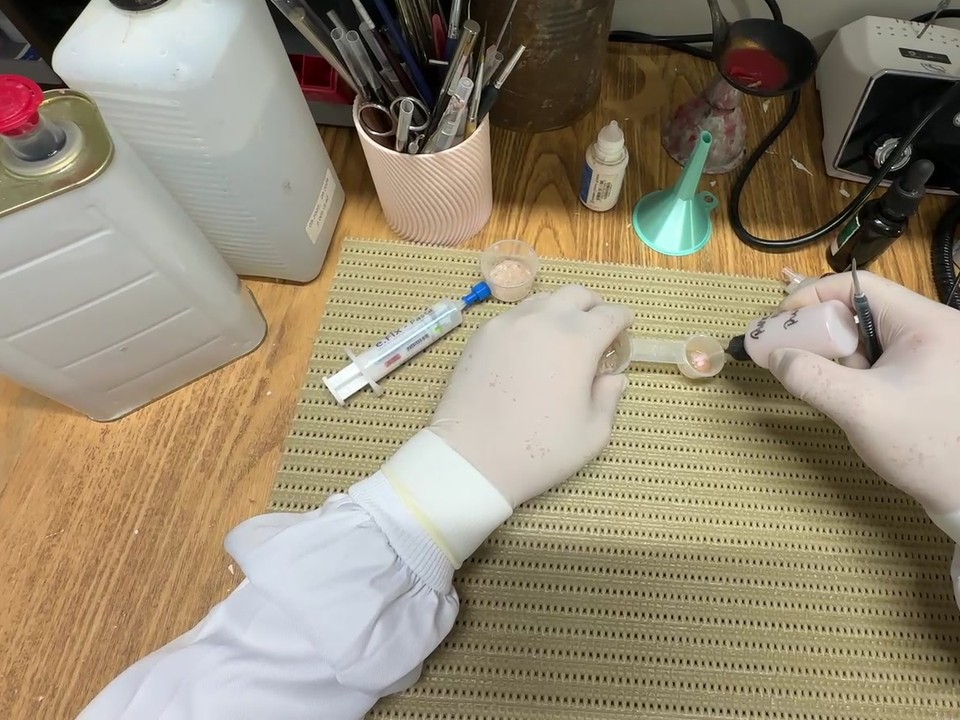} & \includegraphics[width=\linewidth]{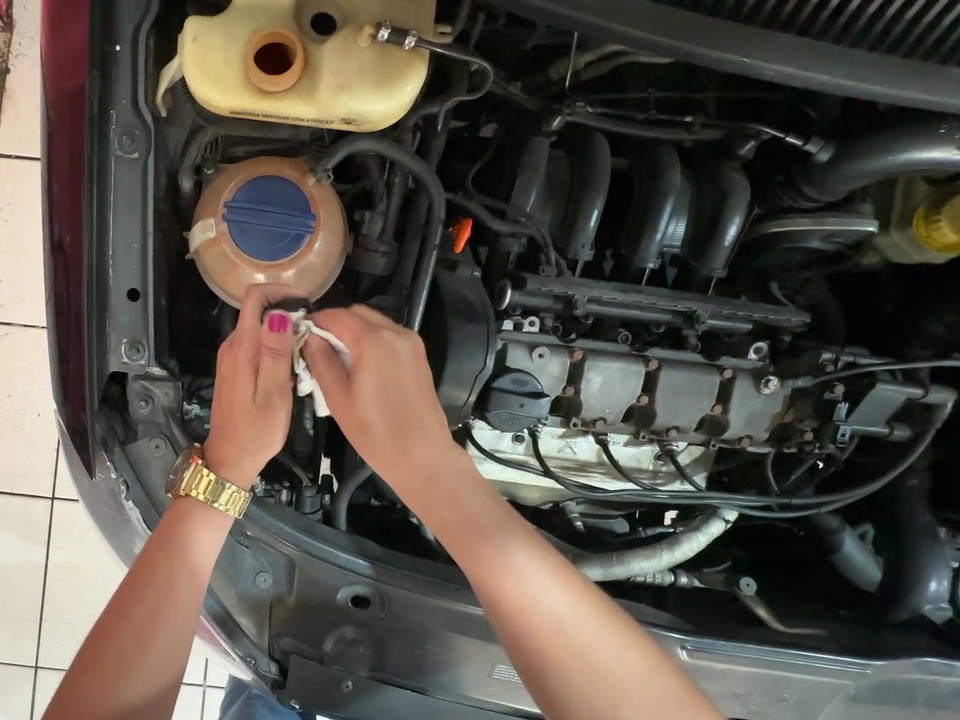} \\
CNC plasma cutting & Breaker panel testing & Industrial sewing & Milling machine setup & Dental lab work & Engine inspection \\[4pt]
\includegraphics[width=\linewidth]{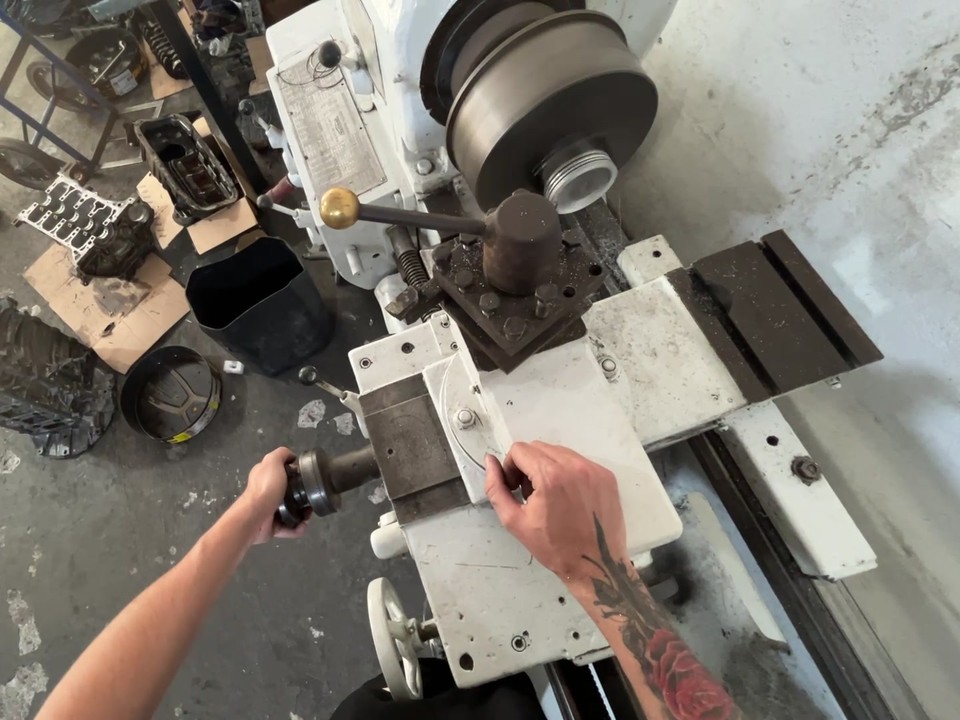} & \includegraphics[width=\linewidth]{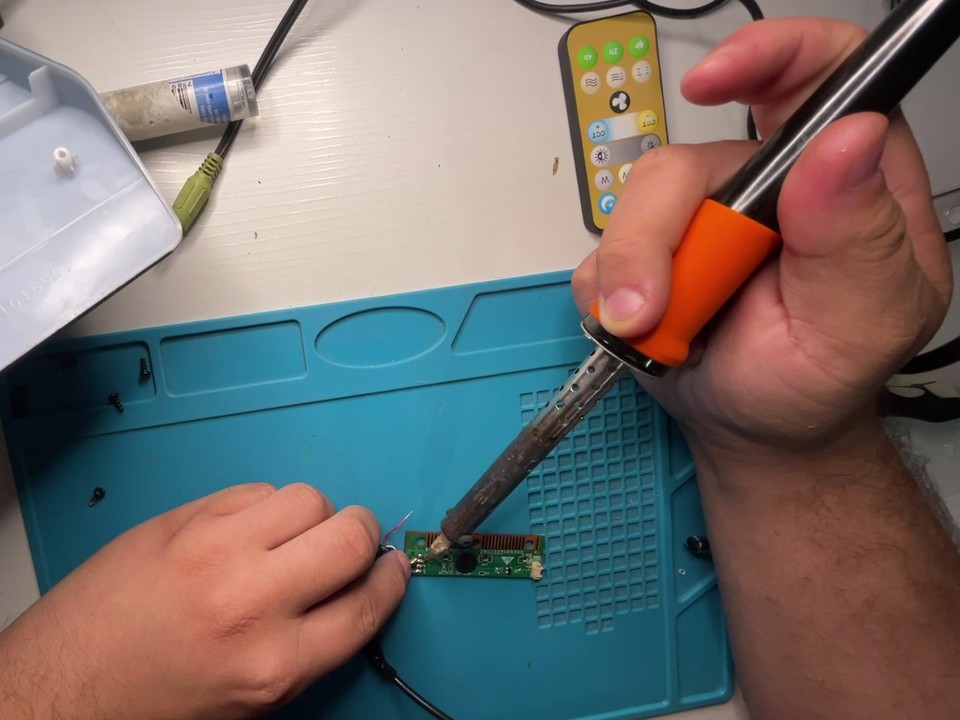} & \includegraphics[width=\linewidth]{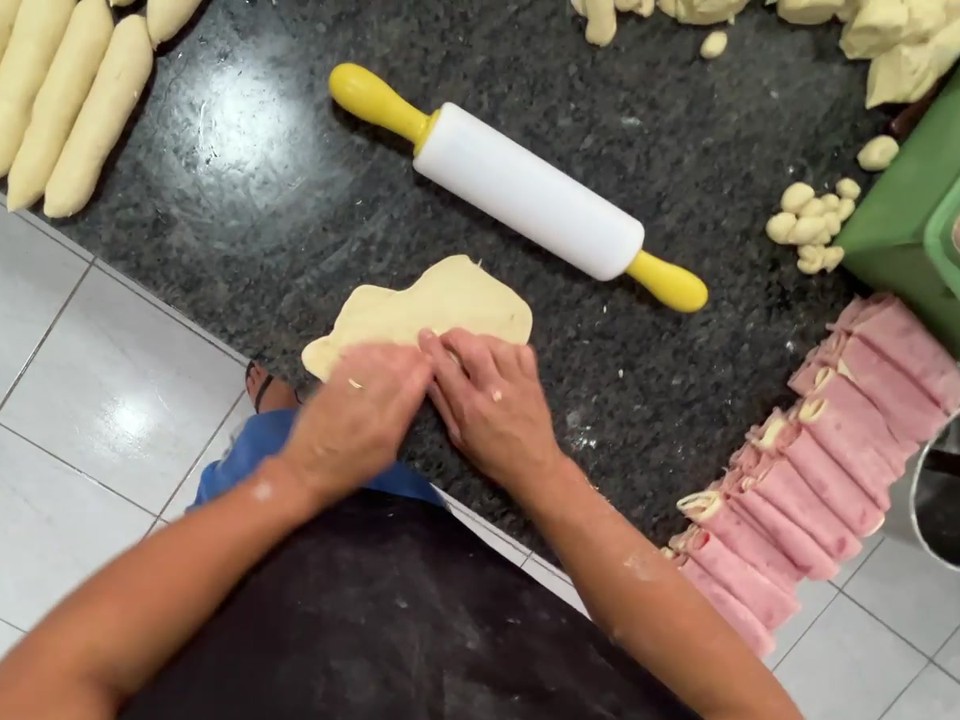} & \includegraphics[width=\linewidth]{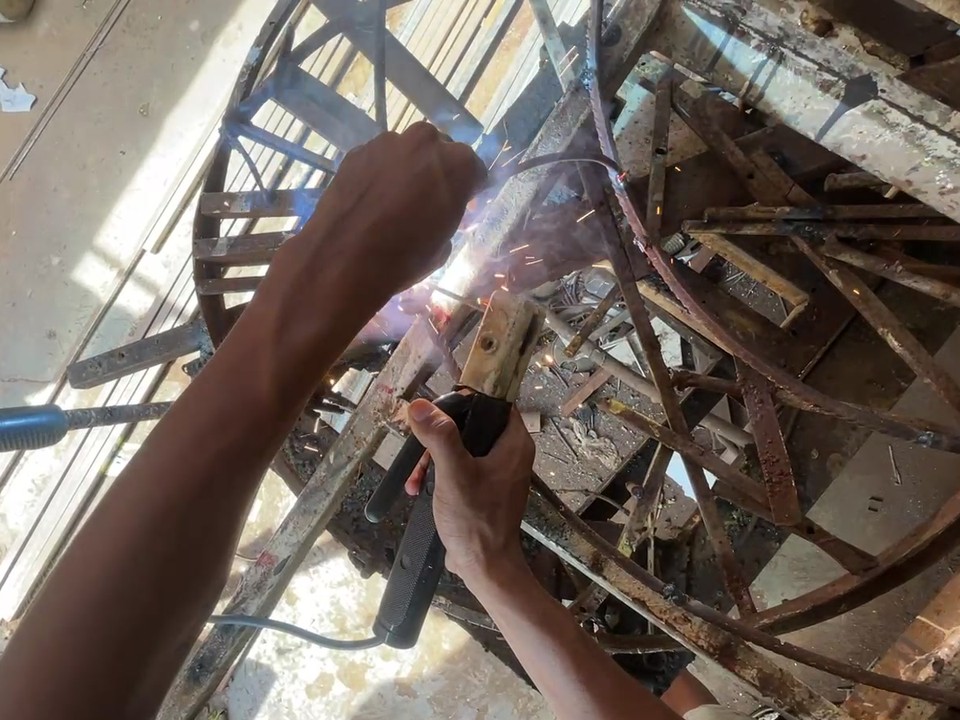} & \includegraphics[width=\linewidth]{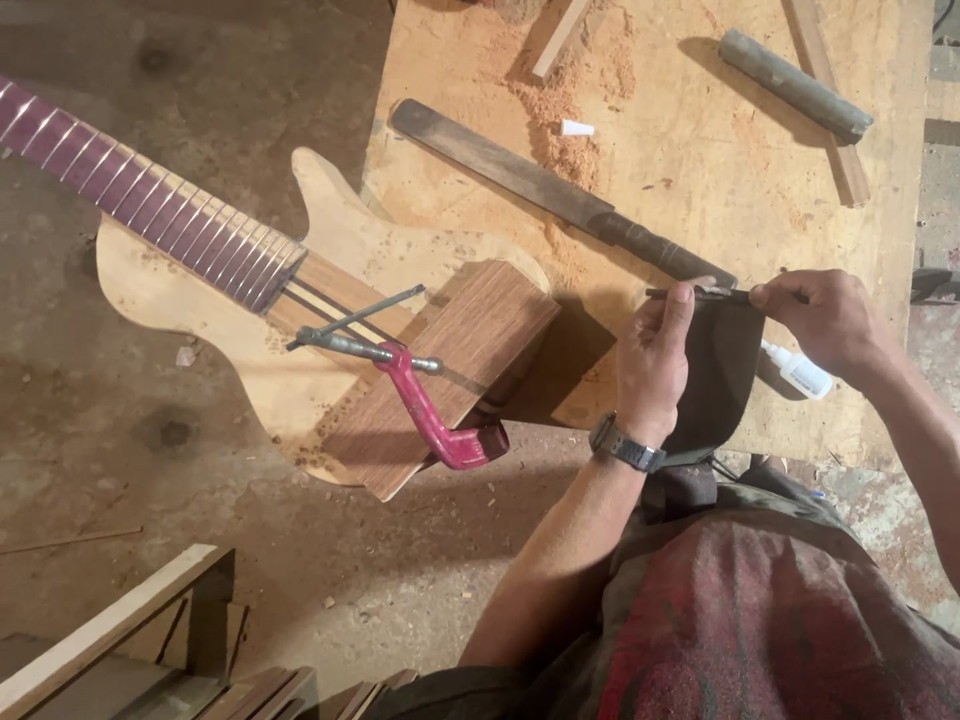} & \includegraphics[width=\linewidth]{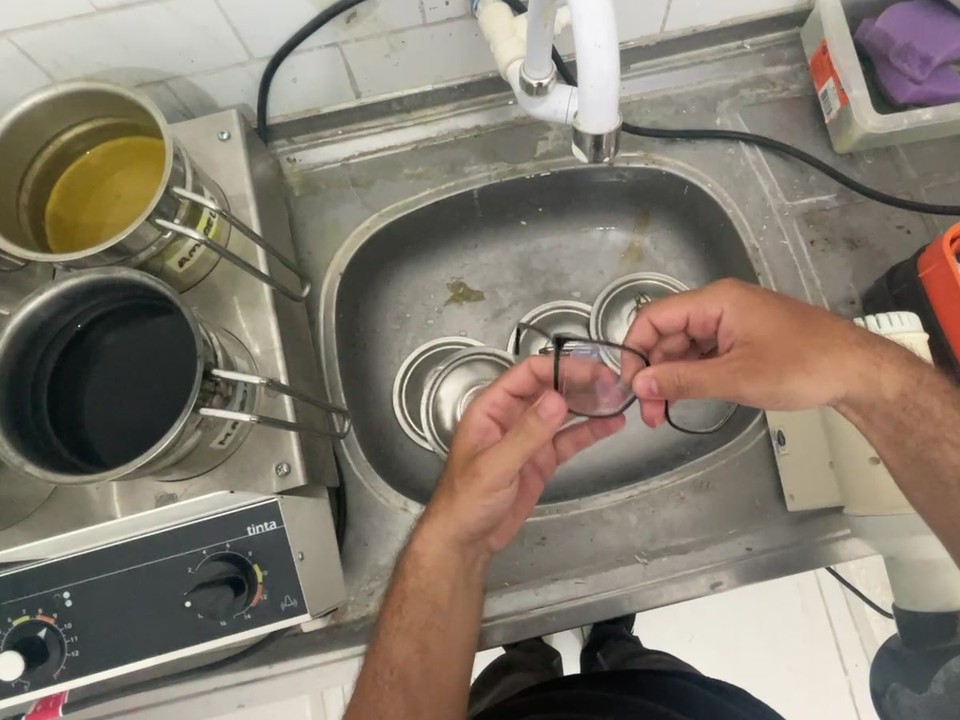} \\
Lathe machining & Keypad board soldering & Bread roll shaping & Rebar welding & Guitar building & Eyeglass lens fitting \\
\end{tabular}
\caption{Examples of skilled work in \hub{}: frames from twelve episodes labeled hard, recorded with a head-mounted phone by twelve contributors in their own workplaces.}
\label{fig:examples}
\end{figure*}

\paragraph{Robot datasets and generalist policies.} Large robot datasets \citep{brohan2023rt1,oxe2024,khazatsky2024droid} report the effect of data diversity on generalization, as does $\pi_{0.5}$ \citep{black2025pi05}. \citet{lin2025scaling} measured how imitation learning generalization scales with environments, objects and demonstrations, using Diffusion Policy \citep{chi2023diffusion}, the policy class also used in DROID. Generalist policies such as OpenVLA \citep{kim2024openvla} and $\pi_0$ \citep{black2025pi0} are trained on pooled robot data, and GR00T N1 \citep{nvidia2025gr00t} adds synthetic data and human video to the mixture.

\paragraph{Human video for robot learning.} Human video serves as a source of supervision in several ways: pretrained visual representations \citep{nair2022r3m,radosavovic2022mvp,majumdar2023vc1,dasari2023unbiased}, affordances and hand-object contact mined from video \citep{bahl2023vrb,srirama2024hrp}, retargeting of human hand motion \citep{qin2022dexmv,bahl2022whirl}, co-training of policies on human and robot data \citep{wang2023mimicplay,kareer2025egomimic,qiu2025humanoid,tao2025dexwild}, latent-action pretraining \citep{ye2025lapa}, and vision-language-action (VLA) models pretrained on human video \citep{yang2025egovla,luo2025beingh0,nvidia2025gr00t,pi2025humantransfer,long2026egoguidance}. We review their quantitative findings in Section~\ref{sec:why}.

\begin{figure}[t]
\centering
\includegraphics[width=\columnwidth]{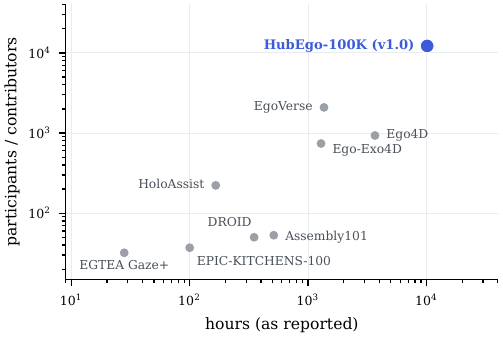}
\caption{Hours against number of people who recorded the data (log scales), among the datasets in Table~\ref{tab:positioning} that report both; each source counts people in its own way. Hours as reported (all views for Ego-Exo4D and Assembly101); DROID counts data collectors.}
\label{fig:posscatter}
\end{figure}

\paragraph{Comparison with existing datasets.} Table~\ref{tab:positioning} compares \hub{} v1.0 with egocentric and manipulation datasets on the two axes of diversity that most sources report, the people who recorded the data and the places where it was recorded, using the numbers of each dataset's paper, website or dataset card; where sources count hours over all camera views or disagree, the table notes it. Among the datasets in Table~\ref{tab:positioning} that report the number of people, the others list 9 to 2,087 and v1.0 has 12,227 contributors (Fig.~\ref{fig:posscatter}). Each contributor records in their own workplace or home, which gives 15,548 environments across 11 setting types and 30 work categories; the other datasets in Table~\ref{tab:positioning} report 45 to 610 kitchens, locations, scenes or rooms. Hours are spread over many contributors: the median contributor recorded 21.3 minutes (mean 49.7), and among the datasets in Table~\ref{tab:positioning}, Ego4D averages about 3.9 hours per camera wearer. v1.0 pairs this diversity with 1080p video, inertial data, task types and per-episode action and language annotations under CC BY 4.0. Section~\ref{sec:diversity} goes through the axes one by one.

\section{The \hub{} Dataset}
\label{sec:dataset}

This section describes how v1.0 was collected, curated, annotated, packaged and released. Table~\ref{tab:stats} lists its headline statistics, and a datasheet \citep{gebru2021datasheets} is given in Appendix~\ref{app:datasheet}.

\begin{figure*}[t]
\centering
\includegraphics[width=\textwidth]{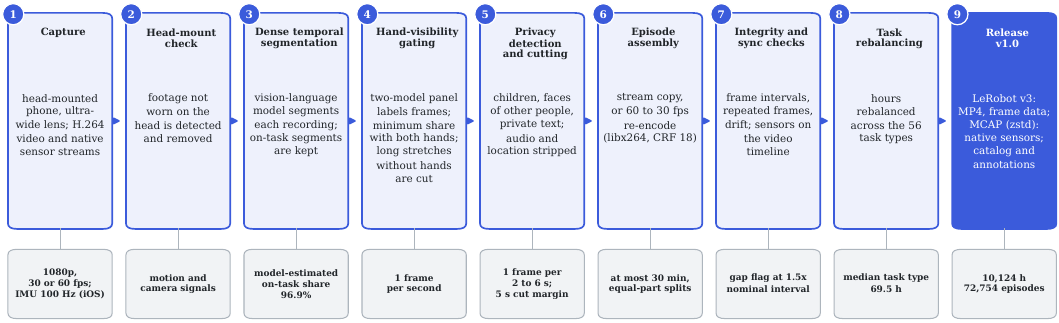}
\caption{Curation pipeline from capture to release, with the main parameter of each stage (Section~\ref{sec:curation}). Human review decisions, where available, take precedence over the automated ones. The signals computed per episode (model-estimated on-task share and hand visibility, capture integrity) are released with the data.}
\label{fig:pipeline}
\end{figure*}

\begin{figure*}[t]
\centering
\setlength{\tabcolsep}{1.5pt}\renewcommand{\arraystretch}{0.9}\footnotesize
\begin{tabular}{@{}*{6}{>{\centering\arraybackslash}p{0.161\textwidth}}@{}}
\includegraphics[width=\linewidth]{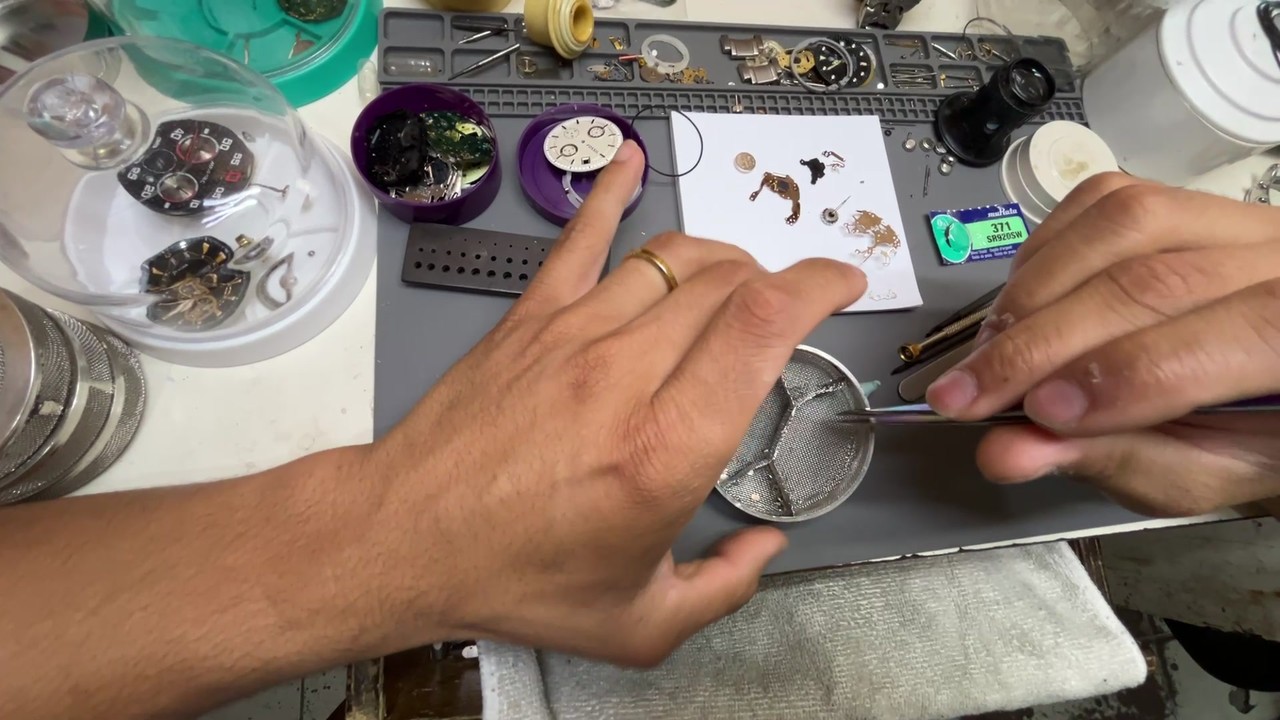} & \includegraphics[width=\linewidth]{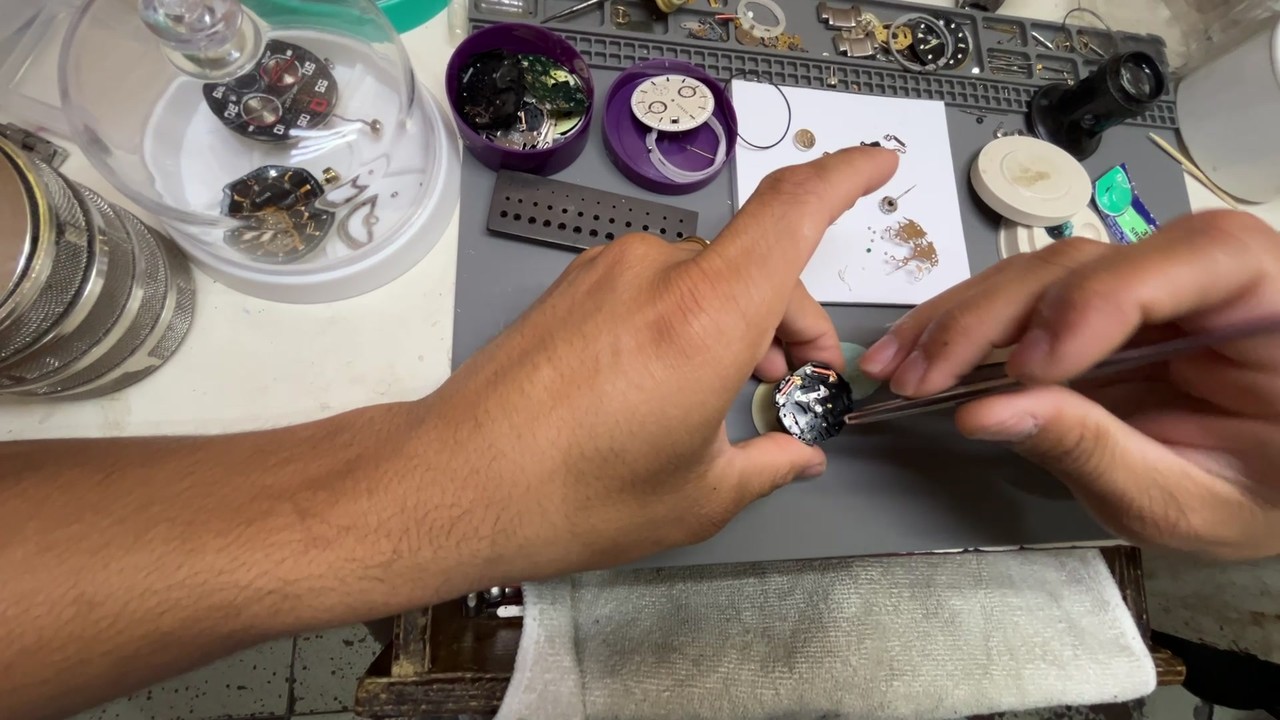} & \includegraphics[width=\linewidth]{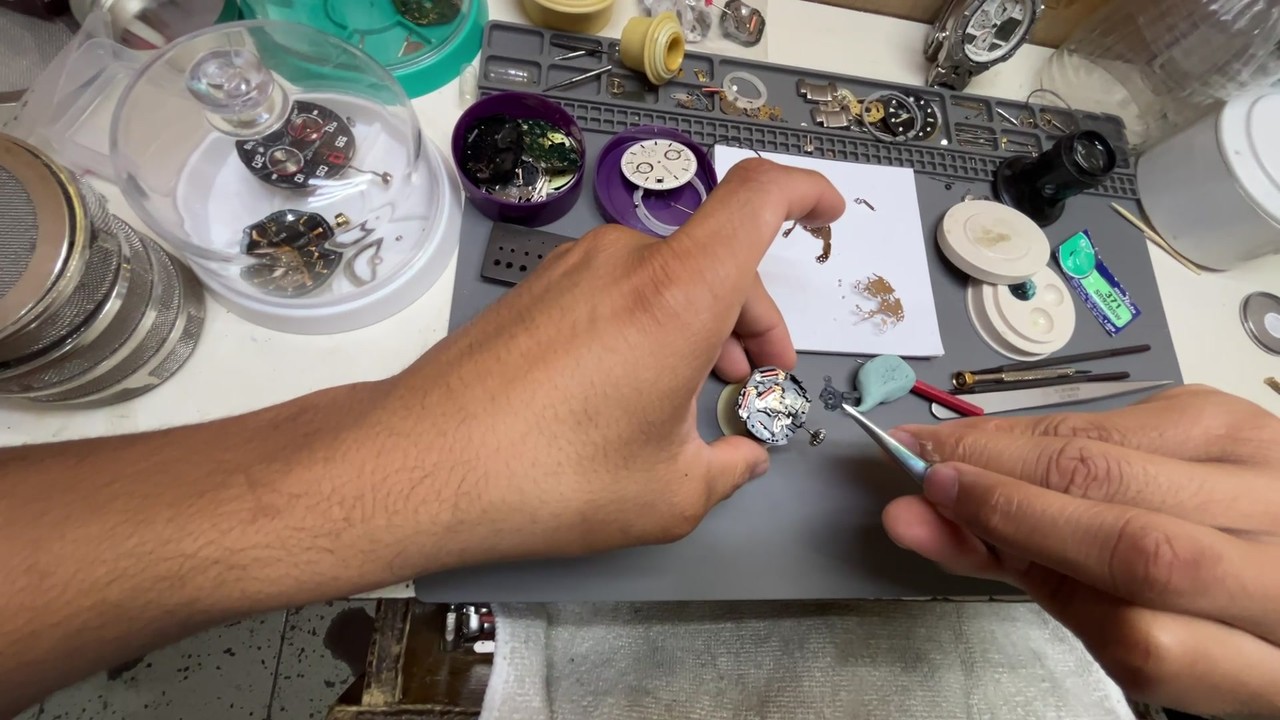} & \includegraphics[width=\linewidth]{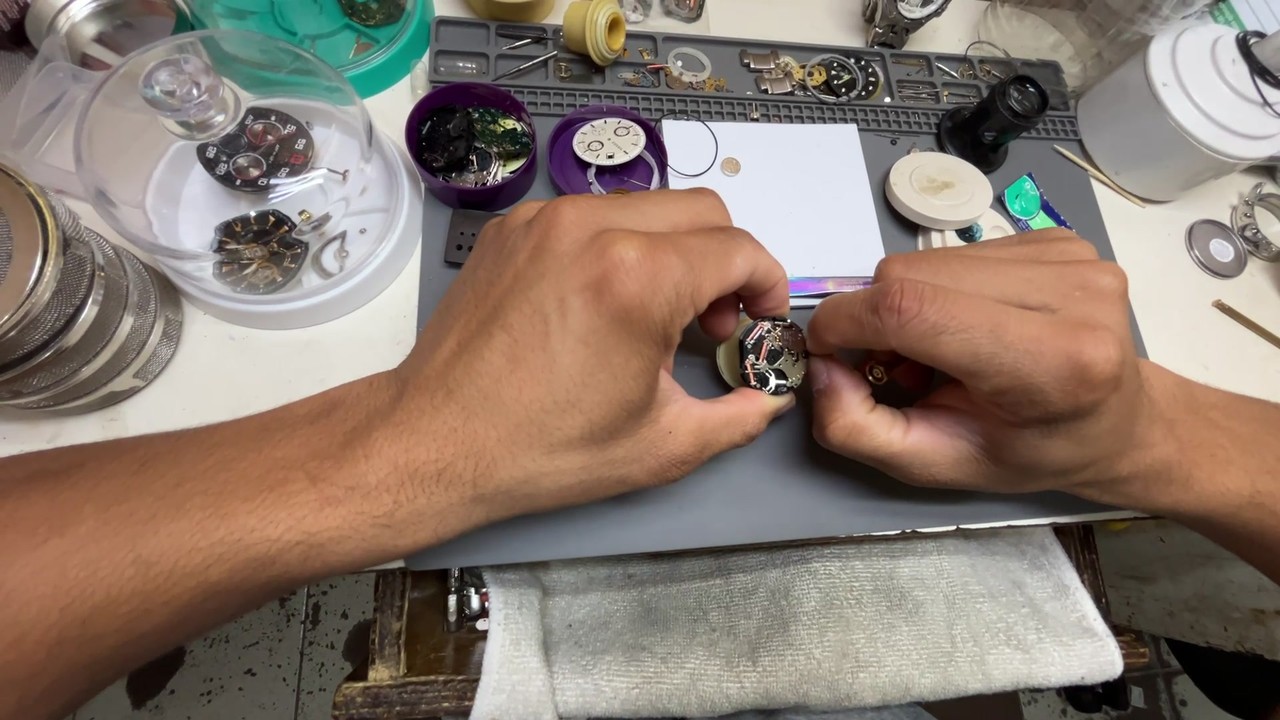} & \includegraphics[width=\linewidth]{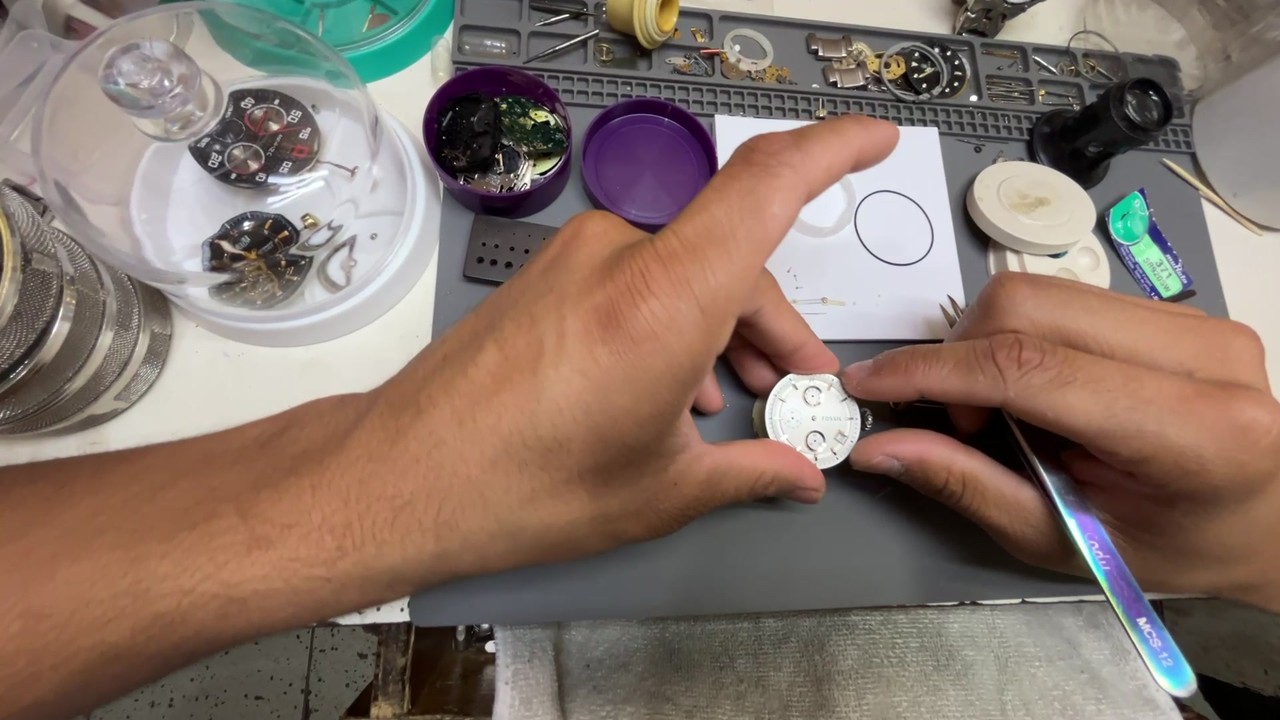} & \includegraphics[width=\linewidth]{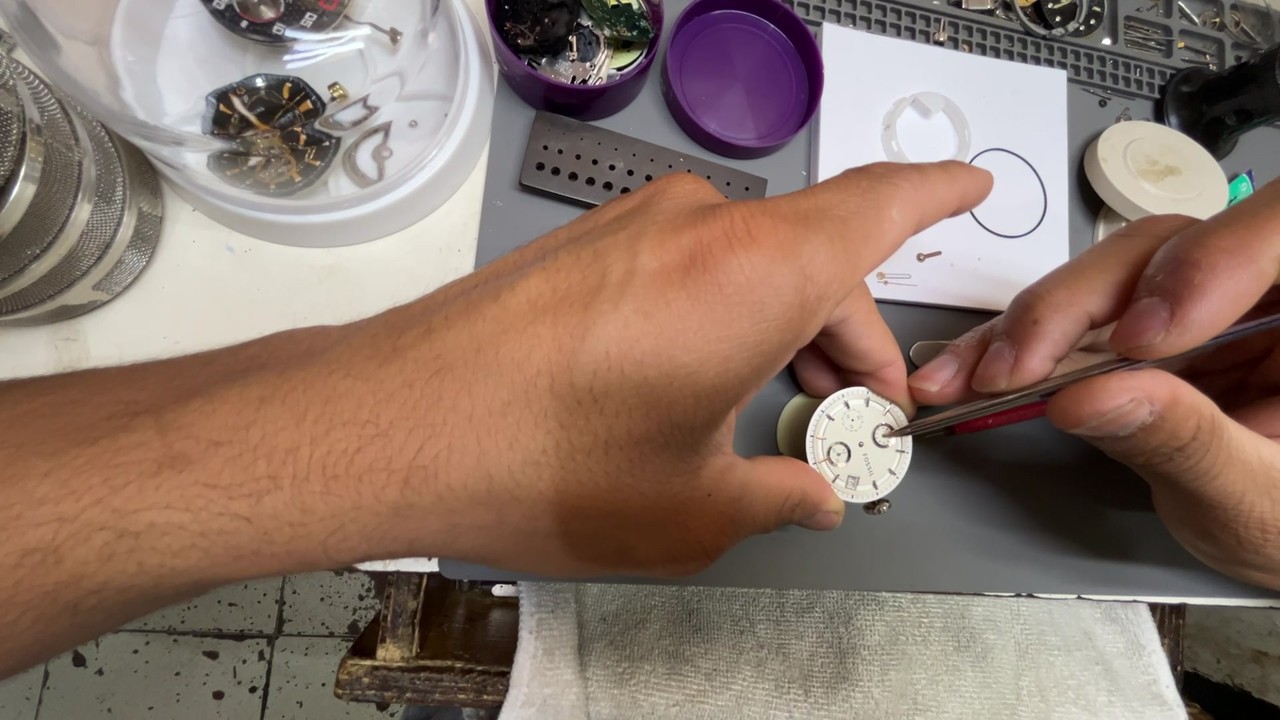} \\
1:17 & 4:02 & 6:40 & 9:09 & 11:22 & 13:54 \\
\end{tabular}

\vspace{3pt}
\includegraphics[width=\textwidth]{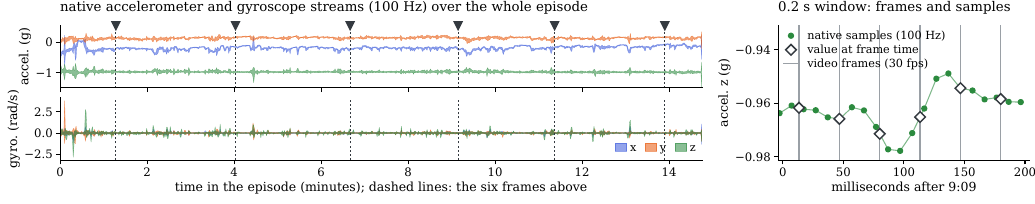}

\vspace{3pt}
\renewcommand{\arraystretch}{1.15}\setlength{\tabcolsep}{4pt}
\begin{tabularx}{\textwidth}{@{}lY@{}}
\thickrule
\textbf{Task type} & Factory \& production line work (work category: Manufacturing \& assembly) \\
\textbf{Description} & Assemble and lubricate watch movement components \\
\textbf{Instruction} & Test the watch movement with meter probes and sort cleaned parts. Assemble and lubricate train wheels and bridge using tweezers, oiler, and screwdriver; install the winding stem, date ring, jumper, springs, hour wheel, washer, and guard plate, then secure screws. \\
\textbf{Annotations} & action family: Assemble \& install; skill group: Tool Use \& Technical Manipulation; difficulty: hard; setting type: Workshop \& repair bench \\
\thickrule
\end{tabularx}
\caption{One episode of \hub{} (14.8 minutes), a watch movement assembly at a watchmaker's bench, with video, sensors and annotations on one timeline. Top: six frames with their time in the episode. Middle: the native accelerometer and gyroscope streams (min-max envelope of the 88,864 samples of each stream; dashed lines mark the six frames) and a 0.2-second window with the capture times of the video frames, the native accelerometer samples and the value at each frame time by linear interpolation, the method of the per-frame readings. Bottom: the released annotations. By model estimate, at least one hand appears in 99.8\% of the sampled frames and both hands in 90.3\%, and the on-task share is 99.7\%.}
\label{fig:episode}
\end{figure*}

\begin{table}[t]
\centering
\small
\caption{\hub{} v1.0 in numbers.}
\label{tab:stats}
\renewcommand{\arraystretch}{1.1}
\begin{tabularx}{\columnwidth}{P{2.55cm}IY}
\thickrule
\headrow \textbf{Quantity} & \textbf{Value} \\
\headrule
Hours & 10,124 \\
\thinrule
Episodes & 72,754 (median 4.9 min, at most 30 min) \\
\thinrule
Recordings & 41,688 \\
\thinrule
Frames & 1,093,194,072 \\
\thinrule
Contributors & 12,227 \\
\thinrule
Environments & 15,548, across 11 setting types \\
\thinrule
Task types & 56, in 30 work categories \\
\thinrule
Action annotations & 35 action families, 18 skill groups \\
\thinrule
Language & description and step-level instruction per episode \\
\thinrule
Video & $1920\times1080$, H.264, 30 fps, monocular \\
\thinrule
Inertial data & accelerometer and gyroscope, native median rate 99.98 Hz, aligned to every frame \\
\thinrule
Size & 43.5 TB video, 1.29 TB native sensor streams \\
\thinrule
Format & LeRobot v3, MCAP, parquet tables \\
\thinrule
License & CC BY 4.0 \\
\thickrule
\end{tabularx}
\end{table}

\begin{figure}[t]
\centering
\includegraphics[width=\columnwidth]{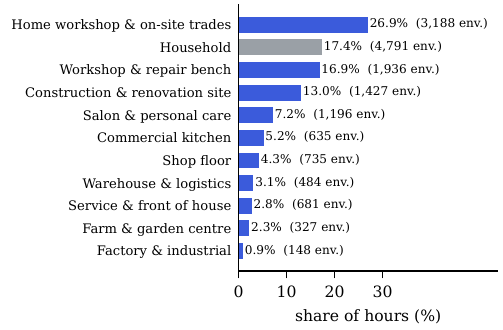}
\caption{Share of hours and number of environments per setting type. Gray: household; blue: workplace settings.}
\label{fig:settings}
\end{figure}

\paragraph{Terminology.}
\label{sec:envdef}
A \emph{recording} is one continuous capture by one contributor. A \emph{segment} is a continuous part of a recording kept after curation, and an \emph{episode}, the unit of the dataset as in LeRobot, is a segment or, for a segment longer than 30 minutes, one of its equal parts. A \emph{contributor} is the person wearing the camera, called camera wearer or participant in other egocentric datasets; contributor is the usual word for people who supply data to a crowd-sourced collection. A \emph{setting type} is one of 11 kinds of place, such as a workshop or a construction site (Fig.~\ref{fig:settings}), comparable to the scene types of DROID \citep{khazatsky2024droid}; ten of them are \emph{workplace settings} and the eleventh is the household. An \emph{environment} is a distinct pair of contributor and setting type, that is, one contributor's own setting of one type. A contributor's sites of one setting type count as one environment, and a site shared by several contributors counts once for each. v1.0 has 15,548 environments, or 1.27 per contributor; the released column \texttt{environment} holds the setting type.

\subsection{Data Collection}
\label{sec:collection}

Contributors record a named task with a head-mounted monocular camera (a phone) during their own work, in their own workplace or home (Fig.~\ref{fig:examples}). The task is named but not scripted, and there are no staged recording sessions: lighting, clutter, tools and materials are those of the workplace or home. Each recording stores the video together with the phone's sensor streams. v1.0 was recorded on 56 phone models; 95.3\% of hours come from iOS devices and the rest from Android devices. 99.1\% of episodes use the ultra-wide lens, and the median diagonal field of view is 114.3 degrees. 26.2\% of hours were recorded in the morning, 49.0\% in the afternoon, 17.2\% in the evening and 7.3\% at night (local time, 0.3\% unlabeled; Fig.~\ref{fig:diff}).

\subsection{Curation Pipeline}
\label{sec:curation}

Fig.~\ref{fig:pipeline} summarizes the nine stages from capture to release. Stages 2 to 7 run automatically on all footage, and human review decisions, where available, take precedence over the automated ones.
\begin{enumerate}
\item \textbf{Capture.} The capture app records $1920\times1080$ H.264 video at 30 or 60 fps and logs the phone's sensors at their native rates: on iOS, accelerometer, gyroscope, magnetometer and device motion at 100 Hz, with heading and barometer at lower rates; on Android, accelerometer and gyroscope at 100 to 500 Hz depending on the model. Every sensor sample carries a timestamp from the phone's monotonic sensor clock.
\item \textbf{Head-mount check.} Footage that was not recorded with the phone worn on the head is detected from motion and camera signals and removed.
\item \textbf{Dense temporal segmentation.} A vision-language model segments every recording in time, and only the segments in which the contributor's hands perform the named task are kept. The model-estimated \emph{on-task share} of each episode is stored with its annotations (\texttt{on\_task\_share}; 96.9\% hours-weighted).
\item \textbf{Hand-visibility gating.} A panel of two vision-language models describes each recording at one frame per second and labels every sampled frame as showing no hand, the left hand, the right hand, both hands, or unclear. A frame counts as showing a hand when at least one model reports one, and unclear frames count as showing none. Recordings in which the hands are mostly out of frame are dropped. Two further rules apply: a recording must reach a minimum share of sampled frames with both hands, and every long stretch without hands is cut out. The model-estimated shares of sampled frames with at least one hand and with both hands are stored in the catalog (\texttt{hands\_share}, \texttt{hands\_share\_both}).
\item \textbf{Privacy detection and cutting.} Review models inspect one frame every 2 to 6 seconds and flag children, faces of other people and private text (documents, screens, cards, license plates). Every flagged frame of a child, and every run of two or more flagged frames of a face or private text, is cut out with a 5-second margin; segments with a single flagged face or text frame are withheld (Section~\ref{sec:privacy}).
\item \textbf{Episode assembly.} Each remaining segment becomes an episode. Recordings made at 30 fps are stream-copied without re-encoding, starting at the first keyframe at or after the segment start, and their frames keep the recorded capture times. Recordings made at 60 fps are re-encoded to constant 30 fps (libx264, CRF 18, one keyframe every 30 frames); output frame $k$ is the source frame nearest to $t_0 + k/30$ s. Segments longer than 30 minutes are split into equal parts.
\item \textbf{Integrity and synchronization checks.} For every episode the pipeline measures the median and maximum frame interval, counts the intervals longer than 1.5 times the nominal interval and the frames repeated by re-encoding, and computes the drift between the real duration and the frame count at 30 fps (median absolute drift 0.017 s). Sensor timestamps are re-expressed in nanoseconds from the first video frame of the episode, so video and sensors share one timeline (Section~\ref{sec:format}). These signals are stored per episode in the catalog.
\item \textbf{Task rebalancing.} Hours are rebalanced across the 56 task types so that no task type dominates; the median task type holds 69.5 hours.
\item \textbf{Annotation and packaging.} The annotation model labels every episode (Section~\ref{sec:taxonomy}), and the episodes are written as a LeRobot v3 dataset with one MCAP file of native sensor streams per episode (Section~\ref{sec:format}).
\end{enumerate}
A recording yields one episode per kept segment or part: v1.0 contains 41,688 recordings and 72,754 episodes, 1.75 episodes per recording on average, and every episode lasts at most 30 minutes.

\subsection{Annotations}
\label{sec:taxonomy}

\begin{figure}[t]
\centering
\includegraphics[width=\columnwidth]{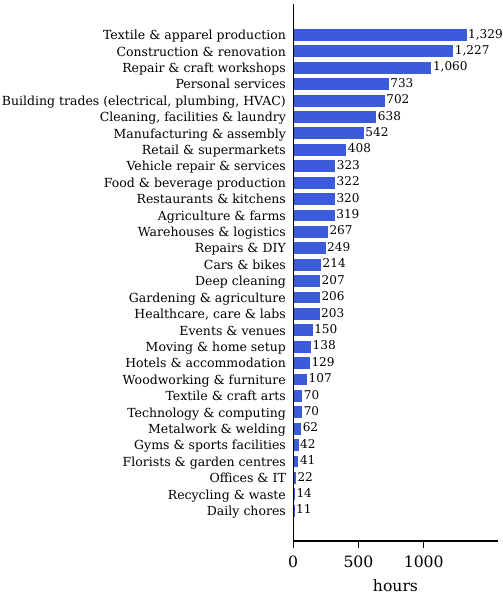}
\caption{Hours per work category (30 categories).}
\label{fig:cat}
\end{figure}

Each episode has one of 56 task types, grouped by industry into 30 work categories (Fig.~\ref{fig:cat}) in the spirit of the scenario labels of Ego4D \citep{grauman2022ego4d}; the task type is the named task the contributor recorded. A vision-language model (the annotation model) then labels each episode by dense video annotation with:
\begin{itemize}
\item one of 35 \emph{action families} (34 named action families and one residual class), a verb-level taxonomy comparable to the verb classes of EPIC-KITCHENS \citep{damen2022epic100};
\item one of 18 \emph{skill groups}, a \emph{difficulty} level (easy, medium or hard) and one of 11 \emph{setting types};
\item a short free-text \emph{description} of the episode (mean 7.4 words), for example ``Machine-sew and overlock sleeveless top seams''; it plays the role of the narrations of EPIC-KITCHENS and Ego4D \citep{damen2022epic100,grauman2022ego4d}, but is machine-generated and covers a whole episode;
\item a step-level \emph{instruction} (mean 35.2 words) that describes goal, actions and objects in order, also stored in the LeRobot subtask tables with the time span of the episode; its steps are not localized in time, unlike the keysteps of Ego-Exo4D \citep{grauman2024egoexo4d}.
\end{itemize}
Work categories describe the trade, setting types the place, and action families what the hands do. Fig.~\ref{fig:episode} shows one episode with its annotations, and Appendix~\ref{app:taxonomy} lists all task types, action families and skill groups with their hours. Each episode also carries the model-estimated share of sampled frames that show at least one hand and both hands (\texttt{hands\_share}, \texttt{hands\_share\_both}), so users can filter by hand visibility.

\subsection{Privacy and Ethics}
\label{sec:privacy}

Each contributor agreed that their recordings may be used to train and evaluate AI and robotics systems. Before release we removed:
\begin{itemize}
\item segments where a child, another person's face or private text (documents, screens, cards, license plates) was detected, cut with a 5-second margin; and recordings in which faces fill a large share of the footage;
\item episodes in which the video annotation detected a minor;
\item care tasks performed on another person;
\item audio, location, device identifiers and contributor identity. Contributor and recording identifiers in the release are pseudonymous strings; the phone make and model are kept as capture metadata.
\end{itemize}

\subsection{Data Format}
\label{sec:format}

\begin{figure}[t]
\centering
\includegraphics[width=\columnwidth]{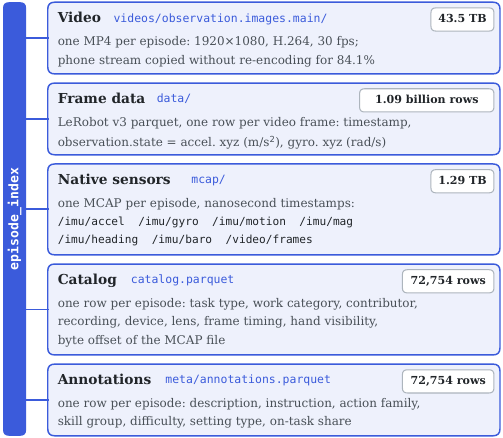}
\caption{Data format of \hub{} v1.0. Each episode has an MP4 video, per-frame rows, an MCAP file of native sensor streams, a catalog row and an annotation row, all keyed by \texttt{episode\_index}.}
\label{fig:format}
\end{figure}

The head-mounted monocular data is in the \texttt{mono/} folder, a LeRobot v3 dataset \citep{cadene2024lerobot}; later releases can add other capture setups as sibling folders. Every episode has one entry in each of five stores, all keyed by \texttt{episode\_index} (Fig.~\ref{fig:format}).

\begin{itemize}
\item \textbf{Video}: \texttt{videos/observation.images.main/}, one MP4 per episode, $1920\times1080$ H.264 at 30 fps. 84.1\% of episodes (86.1\% of hours) are the phone's original stream copied without re-encoding; the rest were recorded at 60 fps and converted to 30 fps. The median video bitrate is 8.1 Mb/s and the video totals 43.5 TB.
\item \textbf{Frame data}: \texttt{data/}, LeRobot v3 parquet, one row per video frame. \texttt{timestamp} is the capture time of the frame from the first frame of the episode, and \texttt{observation.state} (six float32 values) holds the accelerometer (m/s$^2$, gravity included) and gyroscope (rad/s) reading at that time, obtained by linear interpolation of the native samples (Fig.~\ref{fig:episode}, middle).
\item \textbf{Native sensors}: \texttt{mcap/}, one MCAP file \citep{mcap} per episode with every native sensor sample (1.29 TB), as JSON messages with JSON schemas in zstd-compressed chunks. The topics are \texttt{/imu/accel}, \texttt{/imu/gyro} and \texttt{/imu/motion} (device motion) for every episode, \texttt{/imu/mag}, \texttt{/imu/heading} and \texttt{/imu/baro} where the phone provides them, and \texttt{/video/frames}, which gives the capture time of every video frame. Message times are in nanoseconds from the first video frame of the episode, and every sensor message keeps the phone's own sensor-clock timestamp. Values are in the phone's native units (the accelerometer in g on iOS and in m/s$^2$ on Android), named in the channel metadata. The catalog stores the archive, member and byte offset of each episode's MCAP file, so a single episode can be read with one HTTP range request.
\item \textbf{Catalog}: \texttt{catalog.parquet}, one row per episode: task type, work category, contributor and recording identifiers, duration, device facts (model, lens, field of view, source frame rate), hand-visibility shares and capture-integrity signals. Episodes cut from the same recording share a \texttt{recording\_id} and carry their start and end time within it.
\item \textbf{Annotations}: \texttt{meta/annotations.parquet}: description, instruction, action family, skill group, difficulty, setting type and on-task share. \texttt{meta/subtasks.parquet} and \texttt{meta/subtask\_occurrences.parquet} store the instruction of each episode in the LeRobot subtask layout.
\end{itemize}
\begin{table*}[tp]
\centering
\footnotesize
\caption{Data properties that the literature relates to robot learning from human video, and their measurement on \hub{} v1.0 (hours-weighted unless noted). Evidence: Sections~\ref{sec:why-env} to~\ref{sec:why-imu}. Measurements: Section~\ref{sec:analysis}.}
\label{tab:evidence}
\renewcommand{\arraystretch}{1.12}
\begin{tabularx}{\textwidth}{P{2.4cm}IYIY}
\thickrule
\headrow \textbf{Property} & \textbf{Published evidence (selected)} & \textbf{\hub{} measurement} \\
\headrule
\textbf{Environment and object diversity} & Generalization follows a power law in the number of environments and objects; demonstrations per environment saturate; 32 environments with 50 demonstrations each reach about 90\% success on new environments and objects \citep{lin2025scaling}. At matched size, more diverse scenes improve out-of-distribution success \citep{khazatsky2024droid}; more training locations improve success in unseen homes \citep{black2025pi05}. & 15,548 environments across 11 setting types; median 268 environments per task type (47 of 56 task types have at least 100); median 489 per work category; effective number of setting types 7.7 of 11. \\
\thinrule
\textbf{Egocentric human video} & Ego4D pretraining beats ImageNet for manipulation (34\% to 42\%) \citep{nair2022r3m}; human data improves robot policies by 34 to 228\% relative score \citep{kareer2025egomimic}, raises out-of-distribution success from 34.7\% to 59.4\% \citep{qiu2025humanoid} and nearly doubles success on settings seen only in human data after diverse pretraining \citep{pi2025humantransfer}. & 10,124 hours, 1.09 billion frames, head-mounted ultra-wide view (median 114 degrees diagonal), 72.8\% of hours in episodes longer than 10 minutes. \\
\thinrule
\textbf{Hand visibility and curation} & On Ego4D, keeping stable clips with a visible hand (975 of 3,976 hours) turns a negative co-training effect into a positive one \citep{long2026egoguidance}; hand and contact labels mined from egocentric video raise real-robot success \citep{srirama2024hrp}. & Model-estimated shares of sampled frames: at least one hand in 94.7\%, both hands in 76.3\%; model-estimated on-task share 96.9\%; 83.0\% of hours in episodes with a hand in at least 90\% of sampled frames. \\
\thinrule
\textbf{Task breadth and long tail} & Removing 25\% of tasks hurts generalization as much as removing 49\% of data \citep{brohan2023rt1}; training on more than 100 tasks gives 44\% success on 24 unseen tasks \citep{jang2021bcz}. & 56 task types, 30 work categories, 35 action families; 1,429 verb and 3,709 noun lemmas in the machine-generated descriptions; inside v1.0, with episodes in random order, the last 10\% of hours adds about 39 verb and 84 noun lemmas; 34.5\% of hours labeled hard. \\
\thinrule
\textbf{Contributor diversity} & Diversity of people and places is a design goal of Ego4D \citep{grauman2022ego4d}; mixed-skill multi-operator data is harder to imitate at small scale \citep{mandlekar2021robomimic,belkhale2023dataquality}. & 12,227 contributors; no contributor above 0.22\% of hours; 2,824 contributors recorded one hour or more; effective number of contributors 5,666 (exponential of entropy). \\
\thinrule
\textbf{Inertial sensing} & IMU is a low-cost egocentric modality (video takes 60 times more storage and 43 times more compute) \citep{gong2023mmg}; IMU pretraining aligned with text and video nearly doubles activity-recognition F1 on Ego4D \citep{moon2023imu2clip}; a camera plus a low-cost IMU is the minimum suite for metric 6-DoF state estimation \citep{qin2018vinsmono}. & Accelerometer and gyroscope aligned to every frame; native accelerometer rate median 99.98 Hz; in MCAP, device motion for every episode, magnetometer and heading for 99.1\% and barometer for 93.6\% of episodes. \\
\thickrule
\end{tabularx}
\end{table*}

Appendix~\ref{app:fields} describes every field of the two tables.

\subsection{Release and Access}
\label{sec:release}

\hub{} is hosted at \url{https://huggingface.co/datasets/HubLabs/HubEgo-100K} and licensed under Creative Commons Attribution 4.0 (CC BY 4.0): it may be used, shared and adapted for any purpose, including commercially, with credit to ``HubEgo-100K, Hub Labs (Hub Data, Inc.)''. Access goes through a short form with automatic approval. The catalog and annotation tables can be read without downloading video, and single episodes can be fetched individually. This paper describes v1.0 (10,124 hours); further data is released in versioned increments toward 100,000 hours with the same format, annotation schema and license, and release notes and per-version statistics are published on the dataset page.

\section{Which Data Properties Matter: Evidence and Measurements}
\label{sec:why}

This section reviews the published evidence on six data properties. Table~\ref{tab:evidence} sets selected findings next to the corresponding measurements on \hub{} v1.0, which Section~\ref{sec:analysis} reports in full. Numbers about other work are quoted from the cited papers; numbers about \hub{} are computed from the released tables and are hours-weighted unless noted. Several of the cited studies vary robot or hand-held-gripper demonstrations rather than egocentric human video \citep{lin2025scaling,khazatsky2024droid,black2025pi05}; we use them to select the properties to measure.

\subsection{Environment and Object Diversity}
\label{sec:why-env}

\textbf{Evidence.} \citet{lin2025scaling} collected over 40,000 human demonstrations with hand-held grippers and executed more than 15,000 real-world rollouts to measure scaling in imitation learning. They report that ``the generalization performance of the policy follows a roughly power-law relationship with the number of environments and objects'' and that ``the diversity of environments and objects is far more important than the absolute number of demonstrations''. For 8, 16 and 32 environment-object pairs, performance plateaus at 400, 800 and 1,600 demonstrations, about 50 per pair, and collecting in 32 environments with one object each and 50 demonstrations yields a policy with about 90\% success on new environments and objects. In their environment-by-object sweep (their Fig.~6), the normalized score on Pour Water with one object per environment rises from 0.14 with one training environment to 0.74 with sixteen, while with sixteen environments, adding objects per environment no longer helps (0.74 with one object, 0.73 with four). The authors recommend collecting ``in as many diverse environments as possible''. A power-law fit against the total number of demonstrations is weak (correlation coefficients $r=-0.62$ and $-0.79$ for the two tasks in their Fig.~7).

Robot datasets point the same way. DROID \citep{khazatsky2024droid} contains 76k trajectories (350 hours) from 564 scenes and 52 buildings; co-training with DROID improves success by 22 points in distribution and 17 points out of distribution over the next best alternative, and a size-matched subset with fewer scenes performs worse out of distribution, which the authors read as ``the strength of DROID lies in its size and especially in its diversity''. Open X-Embodiment pools data from 22 robots and reports that RT-2-X, trained on the pooled data, improves emergent-skill success from 27.3\% to 75.8\% \citep{oxe2024}, which is evidence for cross-embodiment training rather than for environment diversity. For $\pi_{0.5}$, \citet{black2025pi05} train on data from 3 to 104 locations at a fixed number of unique samples and find that performance in unseen homes ``generally improves with more training locations'', with the 104-location model approaching a model trained on the test environments. \citet{xie2024decomposing} find that factors well covered in training data, such as backgrounds, stop being a bottleneck for generalization, while camera position remains the hardest factor.

\textbf{Measured in \hub{}.} See Sections~\ref{sec:diversity} and~\ref{sec:settings}.

\subsection{Learning from Egocentric Human Video}
\label{sec:why-ego}

\textbf{Evidence.} R3M \citep{nair2022r3m} pretrained on Ego4D improves success on 12 simulated manipulation tasks by over 20\% compared to training from scratch and by over 10\% compared to CLIP and MoCo; on a real Franka arm with 20 demonstrations per task, it reaches 56\% against 24\% for CLIP, and the authors attribute part of the gain to the data itself (34\% to 42\% when switching from ImageNet to Ego4D under the same objective). MVP~\mbox{\citep{radosavovic2022mvp}}, whose 4.5M pretraining images include 2.6M Ego4D frames, outperforms CLIP by up to 75\% and supervised ImageNet pretraining by up to 81\%, and notes that ``larger models require bigger datasets''. VC-1 \citep{majumdar2023vc1}, trained on more than 4,000 hours of egocentric video, finds that scaling data size and diversity helps on average but not universally, and that adding diversity helps more than adding size. \citet{dasari2023unbiased} report that ImageNet, Kinetics and 100 Days of Hands are competitive with Ego4D and that the image distribution matters more than size, a caution that motivates curation (Section~\ref{sec:why-hands}).

For policies, EgoMimic \citep{kareer2025egomimic} reports that adding human embodiment data improves task score by 34 to 228\% relative to robot data alone, and that 2 hours of robot data plus 1 hour of human data outperforms 3 hours of robot data (128 vs.\ 74 points), with 1,400 human demonstrations per hour against 135 robot demonstrations. Co-training with human data in \citet{qiu2025humanoid} raises out-of-distribution success from 34.7\% to 59.4\%, and a human demonstration takes 4.8 to 7.6 times less time to collect than humanoid teleoperation. DexWild \citep{tao2025dexwild} reaches 68.5\% success in unseen environments, nearly four times robot-only training, with human data from 93 environments; it also reports that too high a human-to-robot ratio degrades performance, so robot data remains necessary for grounding. MimicPlay \citep{wang2023mimicplay} uses 10 minutes of human play per environment, which the authors equate to about 3 hours of teleoperation. LAPA \citep{ye2025lapa} pretrains on human manipulation video through latent actions and outperforms a VLA pretrained on robot data from Bridge. Among VLA pretraining efforts, EgoVLA \citep{yang2025egovla} raises long-horizon success on unseen backgrounds from 11.21\% to 28.79\% in simulation, Being-H0 \citep{luo2025beingh0} reports real-robot gains of 10 to 45 points per task from human-video pretraining, and GR00T N1 \citep{nvidia2025gr00t} includes 2,517 hours of human egocentric video in its pretraining mix. \citet{pi2025humantransfer} find that human-to-robot transfer ``emerges once the VLA is pre-trained on sufficient scenes, tasks, and embodiments'', nearly doubling performance on settings seen only in human data (for example 32\% to 71\%).

\textbf{Measured in \hub{}.} See Section~\ref{sec:episodes}.

\subsection{Hand Visibility and Curation}
\label{sec:why-hands}

\textbf{Evidence.} Hands are the interface between a human demonstration and a robot policy. \citet{shan2020hands} built 100 Days of Hands from 131 days of internet footage with 100K annotated frames (189.6K hands) labeled with contact state. HaMeR \citep{pavlakos2024hamer} reconstructs 3D hands as MANO meshes \citep{romero2017mano} from 2.7M training examples, of which only 5\% come from in-the-wild datasets, and reports a 2 to 3 times improvement in PCK@0.05 on in-the-wild data over prior work. VISOR \citep{darkhalil2022visor} provides dense hand-object segmentation for 36 hours of kitchen video. VRB \citep{bahl2023vrb} learns affordances from human hand-object contact and reaches 57\% average success against 25\% for the next method; HRP \citep{srirama2024hrp} injects hand, contact and active-object labels mined from 150K Ego4D videos into pretrained encoders and improves real-robot success by at least 15\% across five tasks, raising an Ego4D-MAE baseline from 0.51 to 0.73.

Direct evidence on curation comes from \citet{long2026egoguidance}, who co-train a VLA on robot data and action-free Ego4D video. Of 3,976 raw hours, a camera-stability gate keeps 1,893 hours (48\%) and an additional visible-hand gate keeps 975 hours (25\%). ``Training with raw Ego4D clips underperforms the robot-only baseline, whereas the clips retained after both gates surpass it''; the stability gate alone ``provides at most a small improvement'', and adding the visible-hand gate gives a further gain. The reported differences are about one to two percentage points.

\textbf{Measured in \hub{}.} See Section~\ref{sec:hands}.

\subsection{Task Breadth and the Long Tail}
\label{sec:why-tasks}

\textbf{Evidence.} In RT-1 \citep{brohan2023rt1}, ``removing 25\% of the tasks while keeping 97\% of the data achieves an equivalent generalization performance to reducing the dataset size by as much as 49\%'', and the authors conclude that ``data diversity is more essential than data quantity''. BC-Z \citep{jang2021bcz} scaled data collection to more than 100 tasks and performed 24 unseen tasks with 44\% average success without demonstrations for them.%
\setcounter{dbltopnumber}{1}%
\begin{figure*}[t]
\centering
\includegraphics[width=\textwidth]{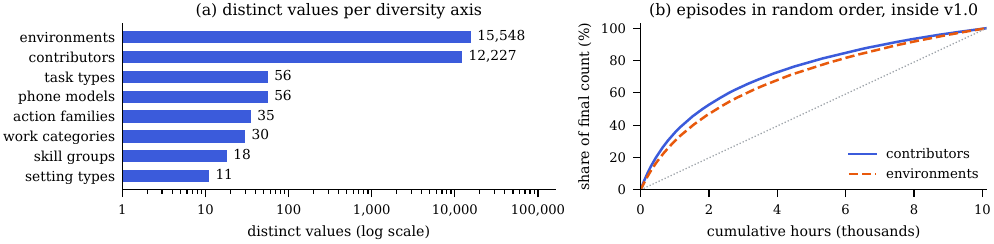}
\caption{Diversity of \hub{} v1.0. (a) Number of distinct values along each diversity axis (log scale). (b) Distinct contributors and environments as a share of their final count against cumulative hours, inside v1.0 with episodes in random order (mean of five orders).}
\label{fig:diversity}
\end{figure*}%
\begin{figure}[t]
\centering
\includegraphics[width=\columnwidth]{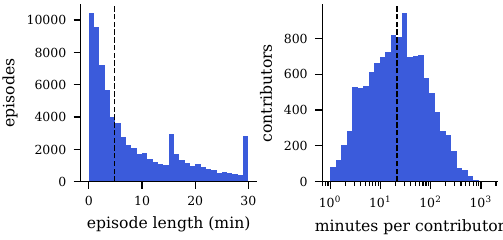}
\caption{Left: episode length. Right: minutes recorded per contributor (log scale). Dashed lines: medians (4.9 and 21.3 minutes).}
\label{fig:lengths}
\end{figure}%
\begin{figure*}[t]
\centering
\setlength{\tabcolsep}{1.5pt}\renewcommand{\arraystretch}{0.9}\scriptsize
\begin{tabular}{@{}*{5}{>{\centering\arraybackslash}p{0.194\textwidth}}@{}}
\includegraphics[width=\linewidth]{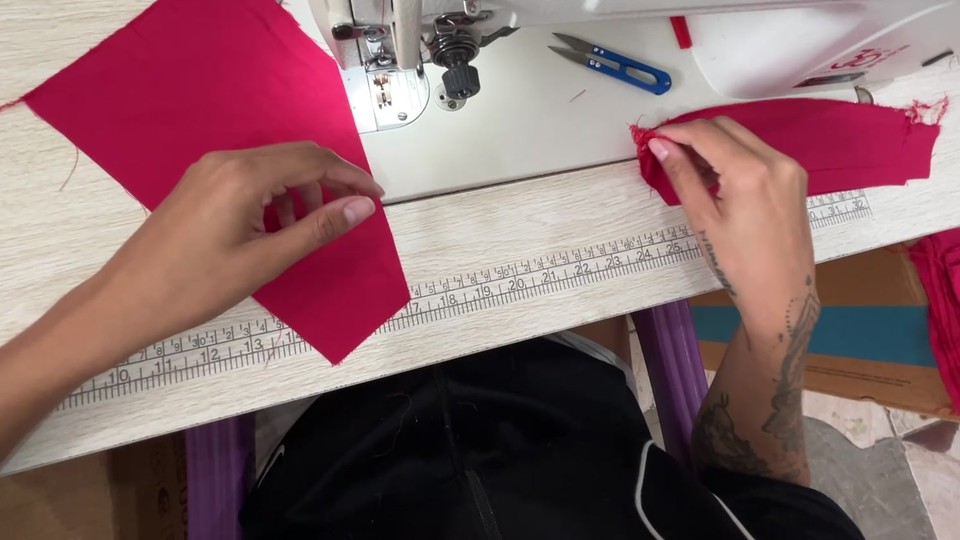} & \includegraphics[width=\linewidth]{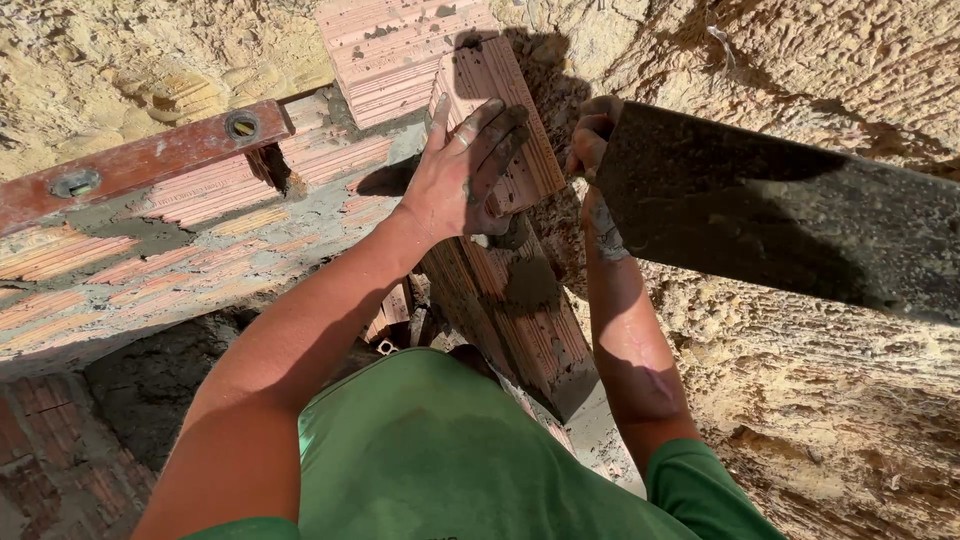} & \includegraphics[width=\linewidth]{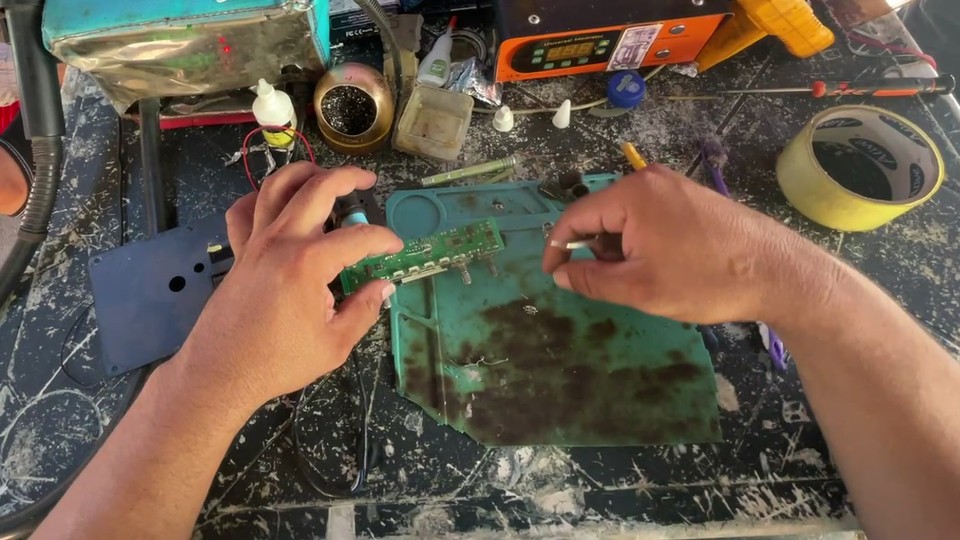} & \includegraphics[width=\linewidth]{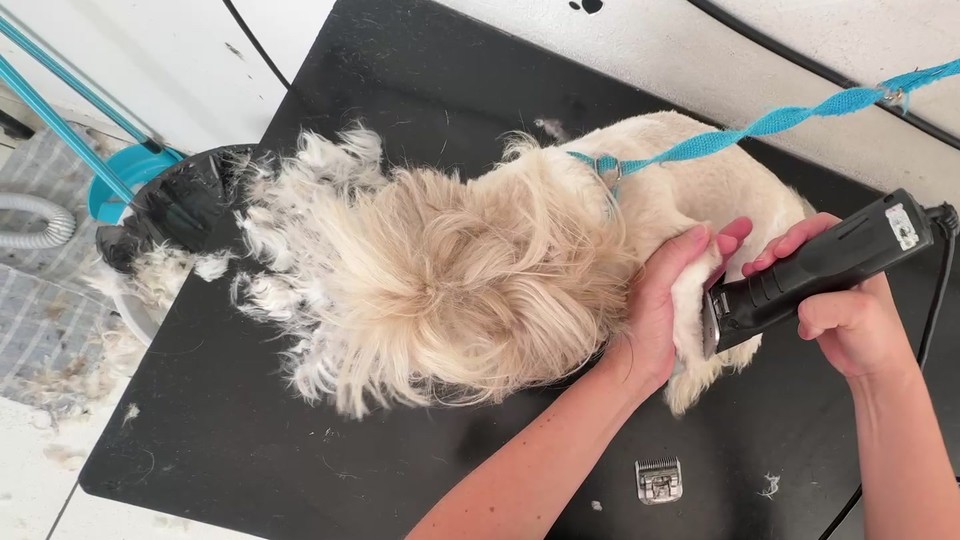} & \includegraphics[width=\linewidth]{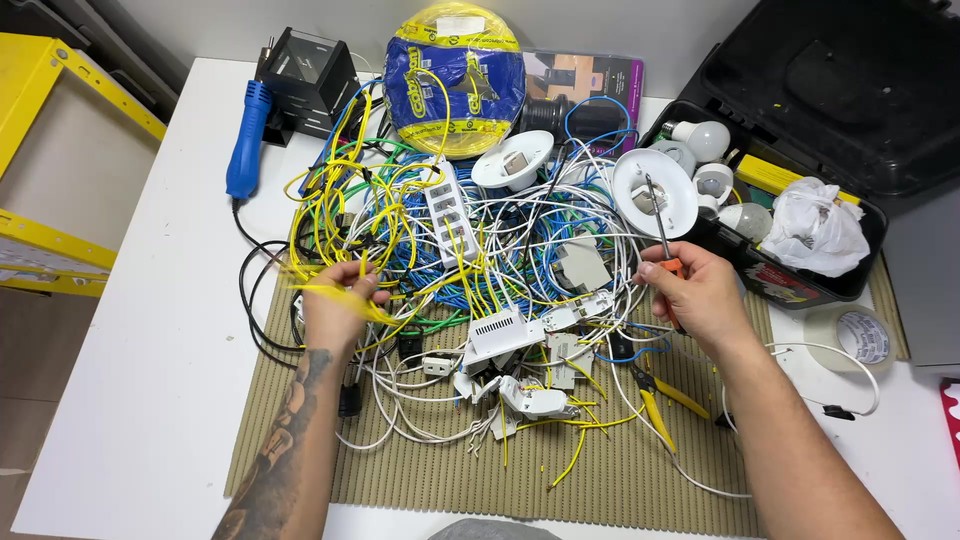} \\
Textile \& apparel production & Construction \& renovation & Repair \& craft workshops & Personal services & Building trades \\[1.5pt]
\includegraphics[width=\linewidth]{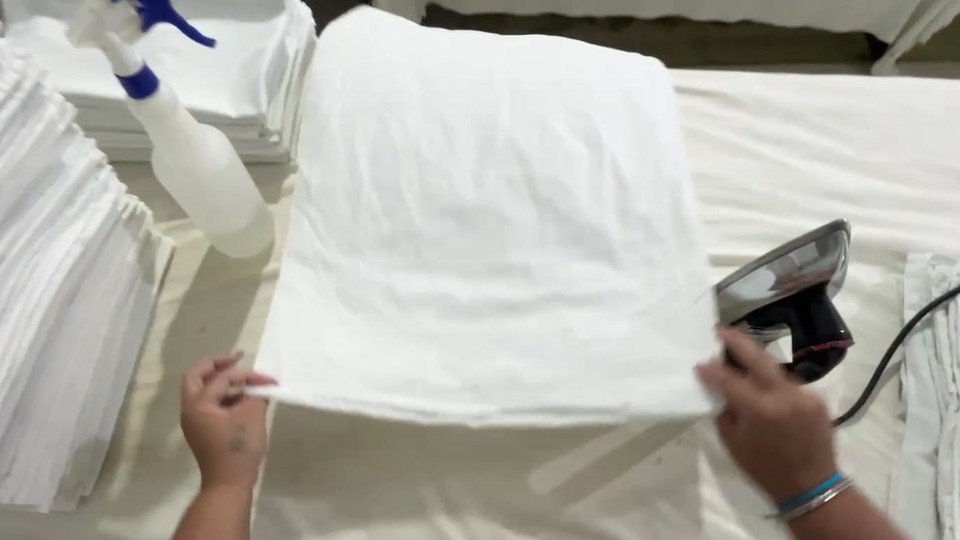} & \includegraphics[width=\linewidth]{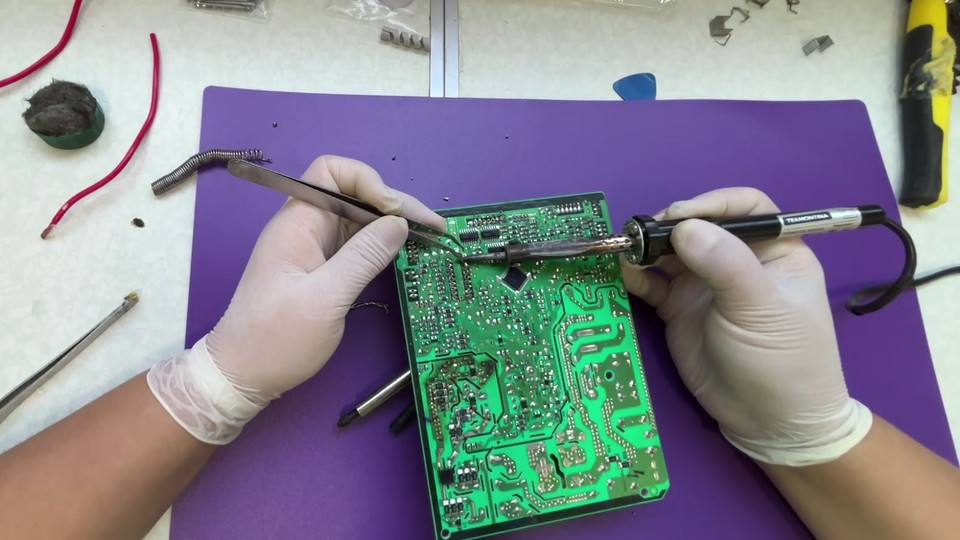} & \includegraphics[width=\linewidth]{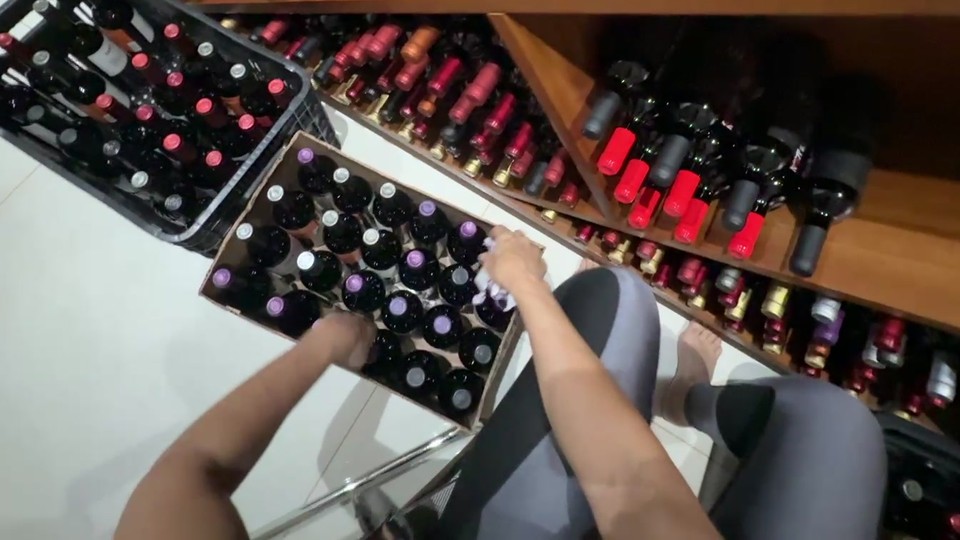} & \includegraphics[width=\linewidth]{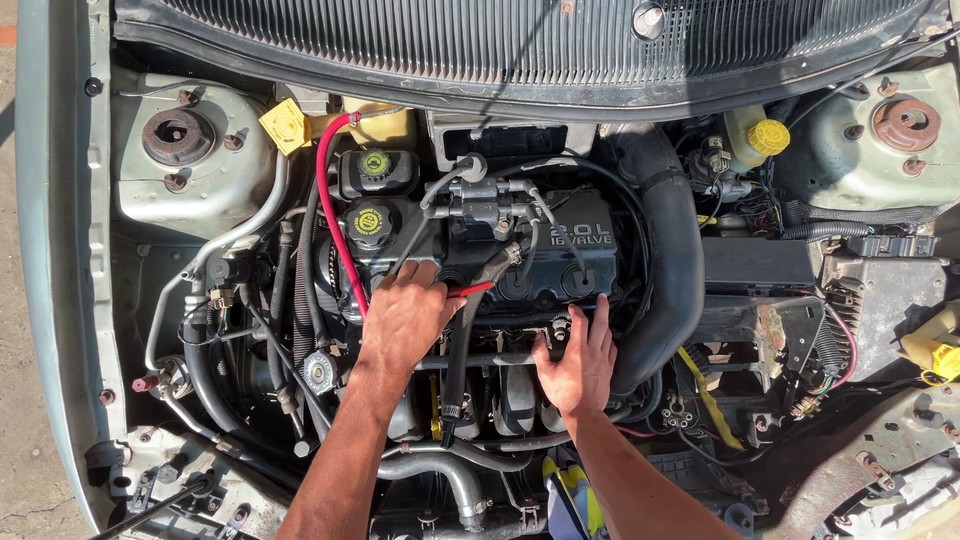} & \includegraphics[width=\linewidth]{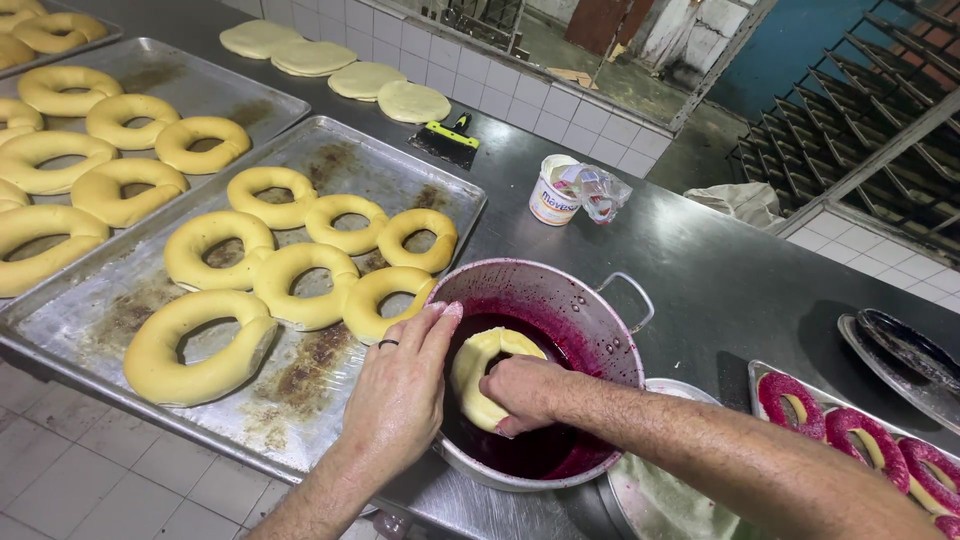} \\
Cleaning, facilities \& commercial laundry & Manufacturing \& assembly & Retail \& supermarkets & Vehicle repair \& services & Food \& beverage production \\[1.5pt]
\includegraphics[width=\linewidth]{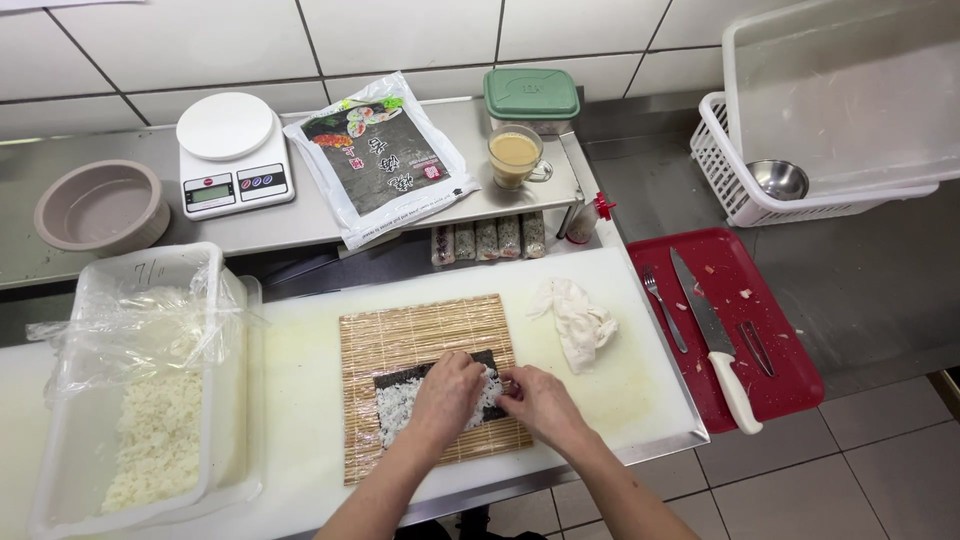} & \includegraphics[width=\linewidth]{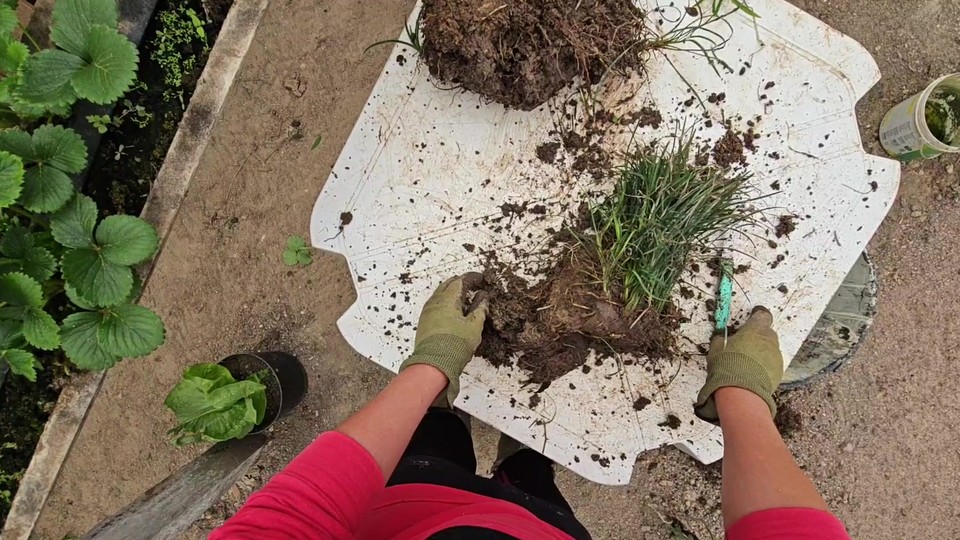} & \includegraphics[width=\linewidth]{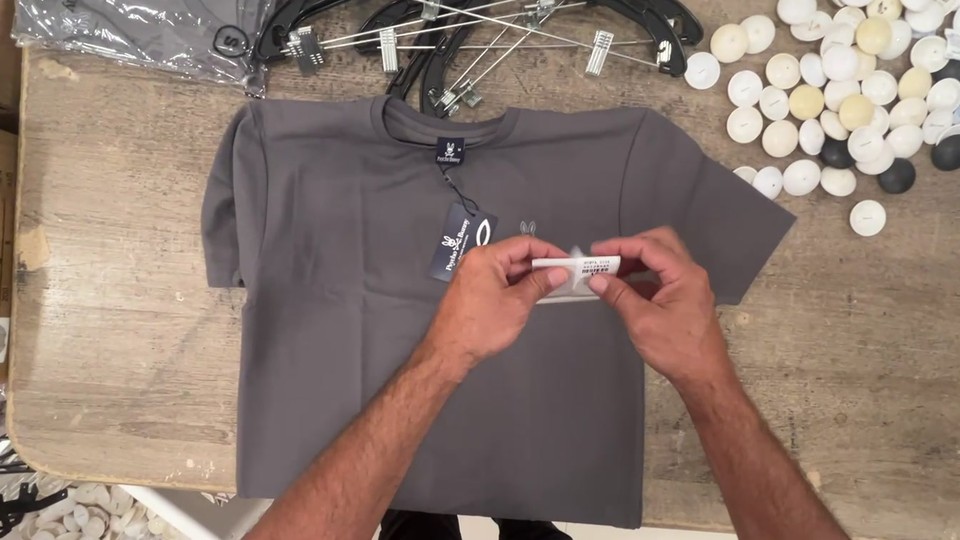} & \includegraphics[width=\linewidth]{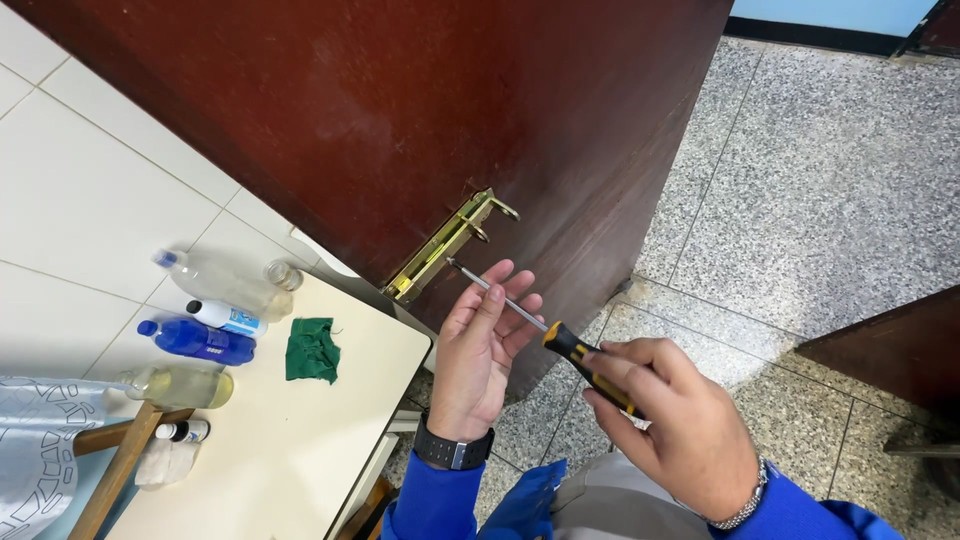} & \includegraphics[width=\linewidth]{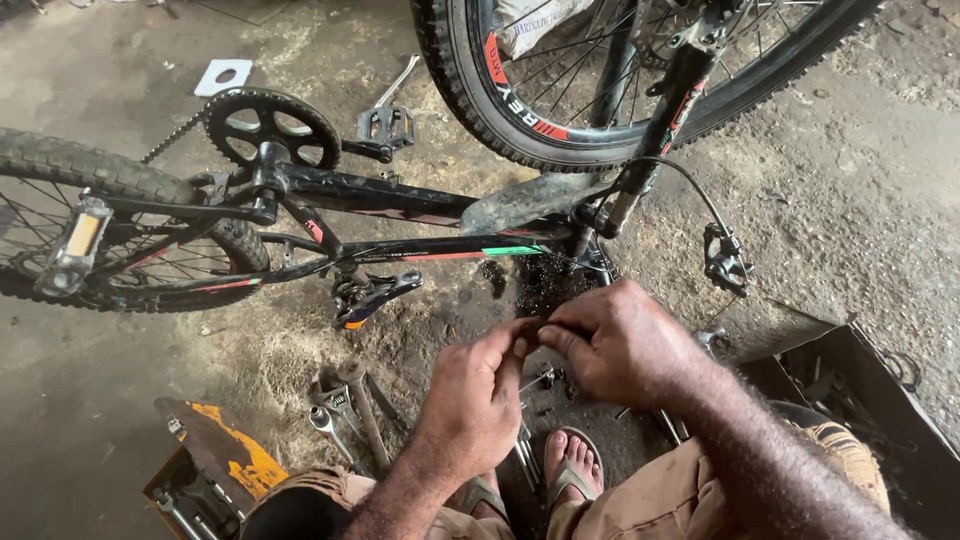} \\
Restaurants \& kitchens & Agriculture \& farms & Warehouses \& logistics & Repairs \& DIY & Cars \& bikes \\[1.5pt]
\includegraphics[width=\linewidth]{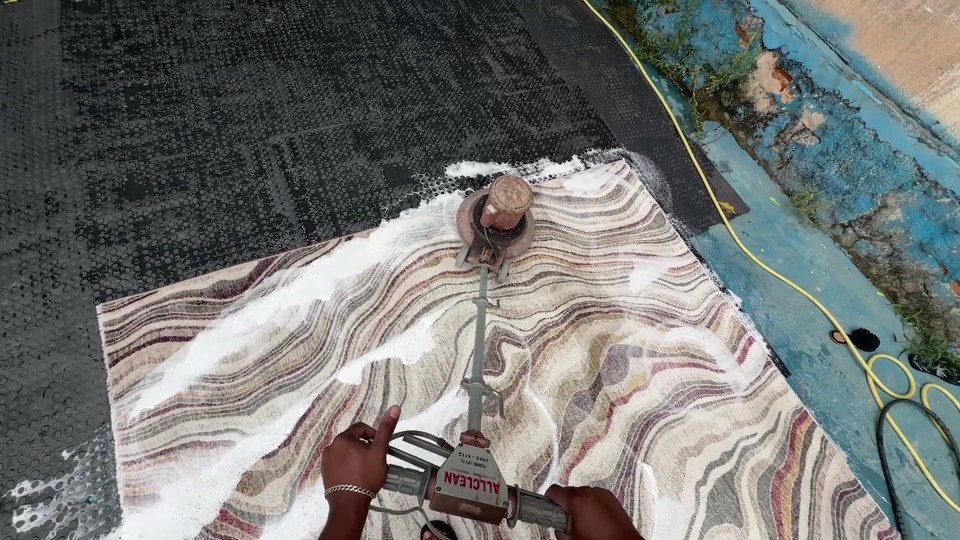} & \includegraphics[width=\linewidth]{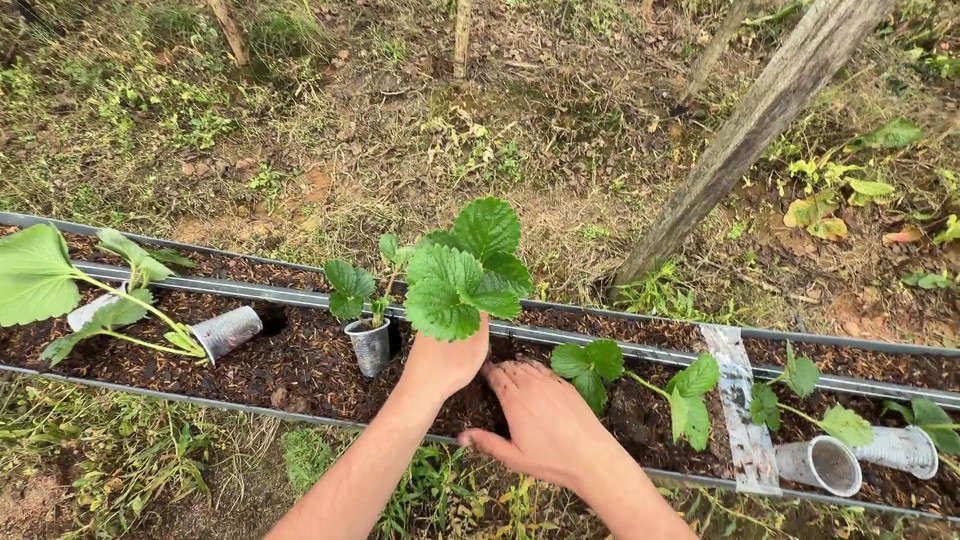} & \includegraphics[width=\linewidth]{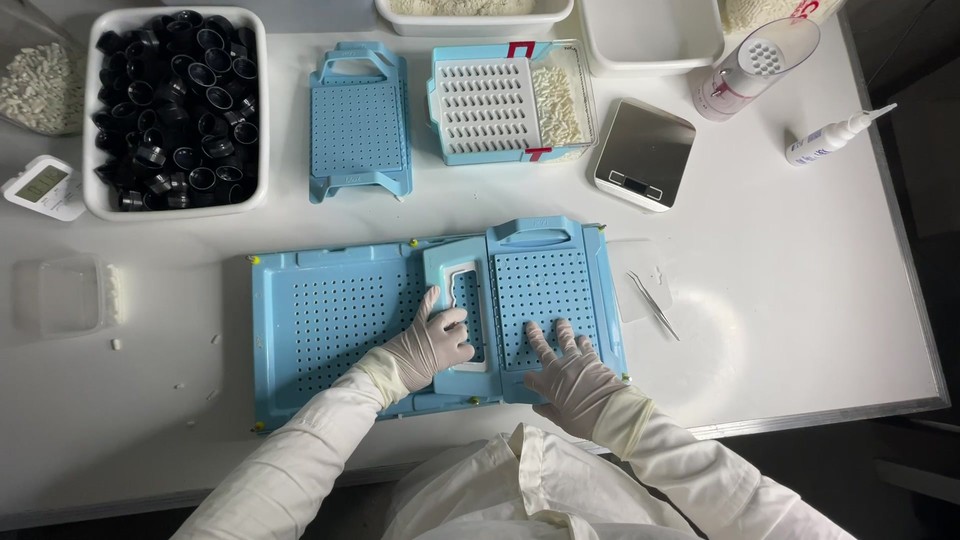} & \includegraphics[width=\linewidth]{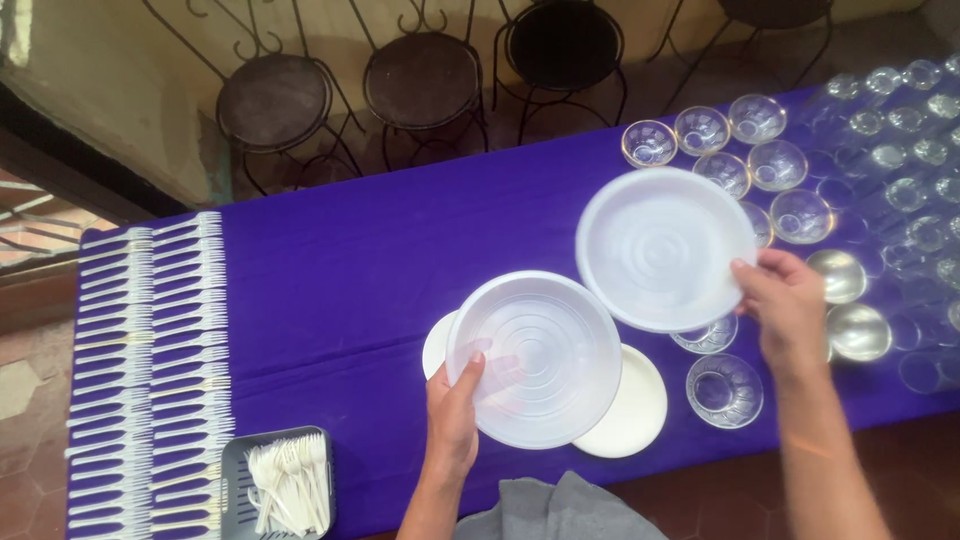} & \includegraphics[width=\linewidth]{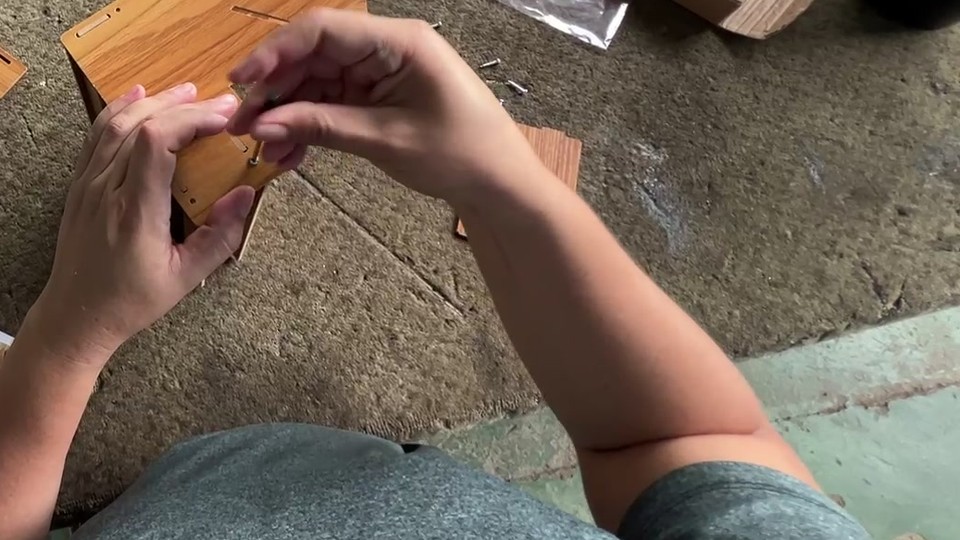} \\
Deep cleaning & Gardening \& agriculture & Healthcare, care \& labs & Events \& venues & Moving \& home setup \\[1.5pt]
\includegraphics[width=\linewidth]{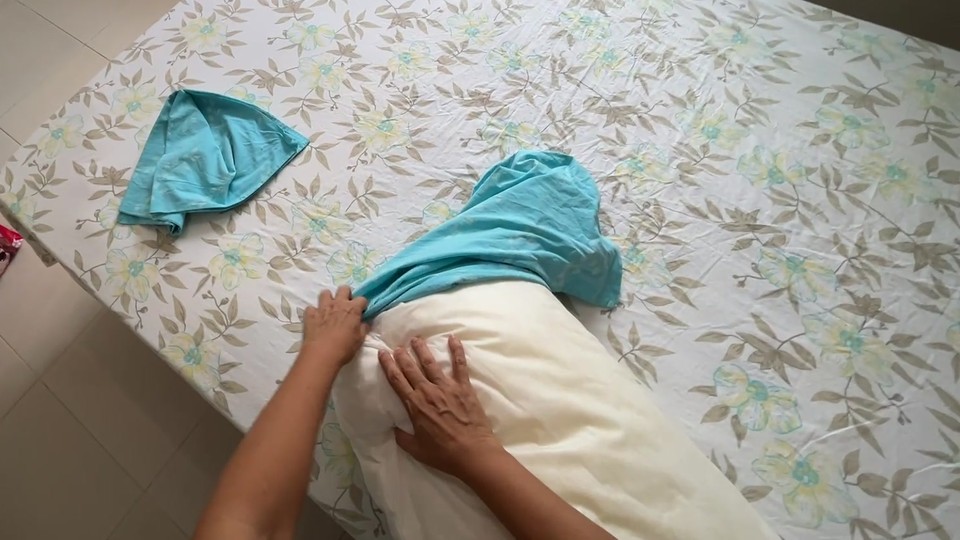} & \includegraphics[width=\linewidth]{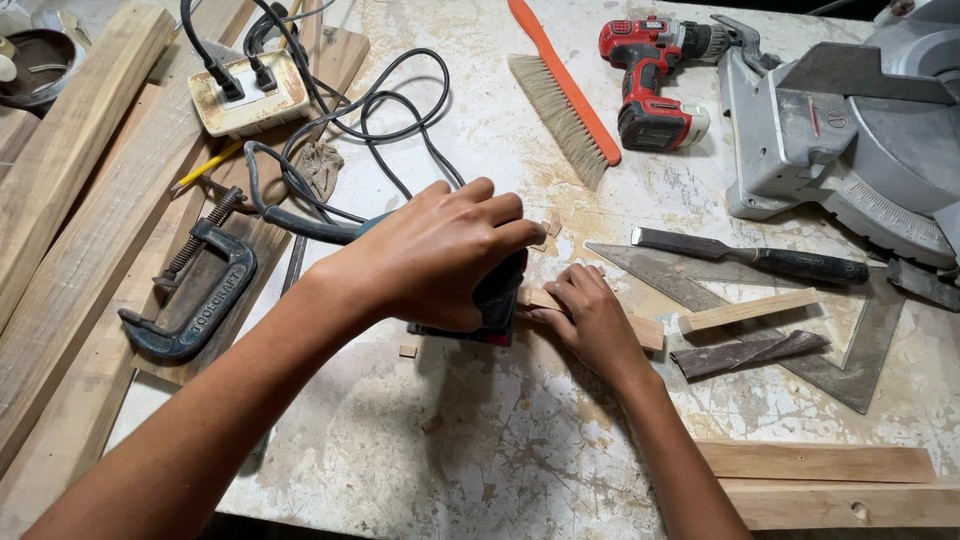} & \includegraphics[width=\linewidth]{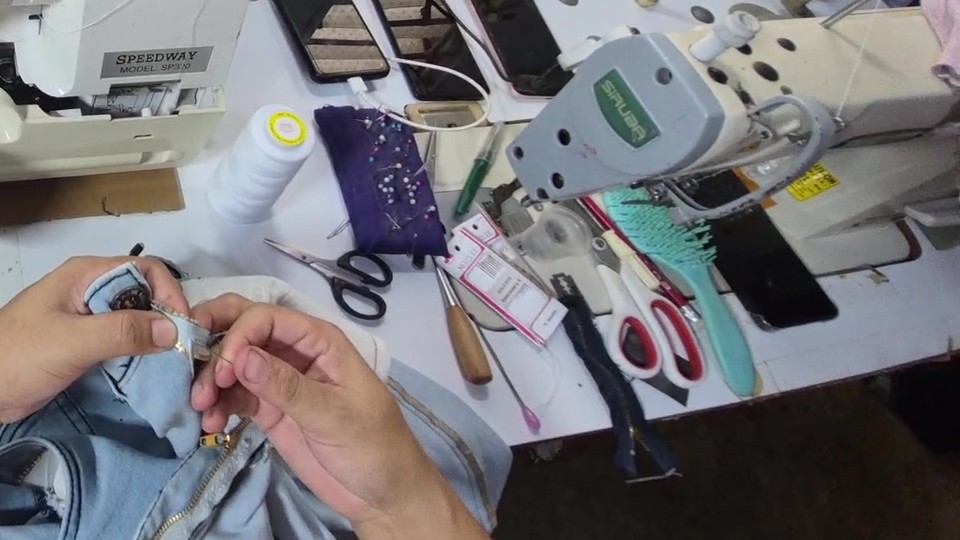} & \includegraphics[width=\linewidth]{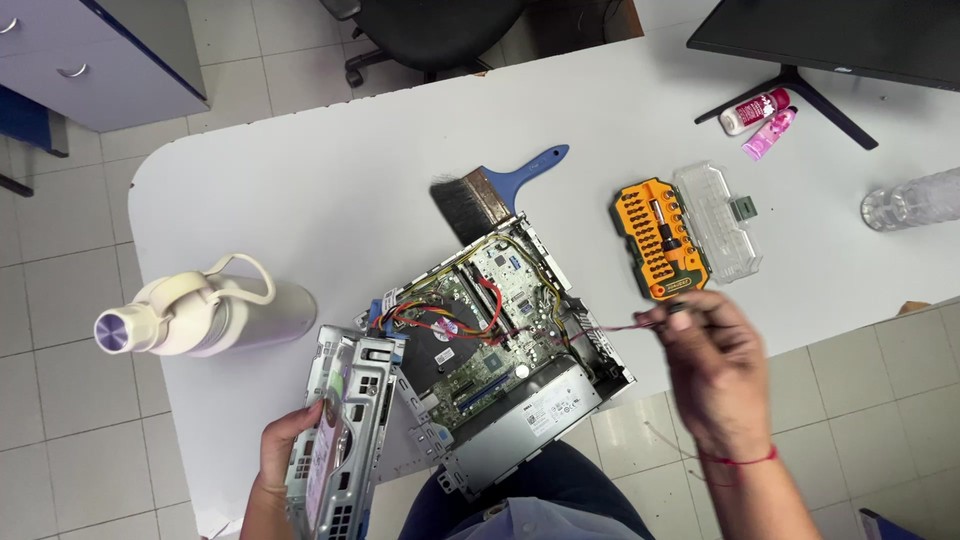} & \includegraphics[width=\linewidth]{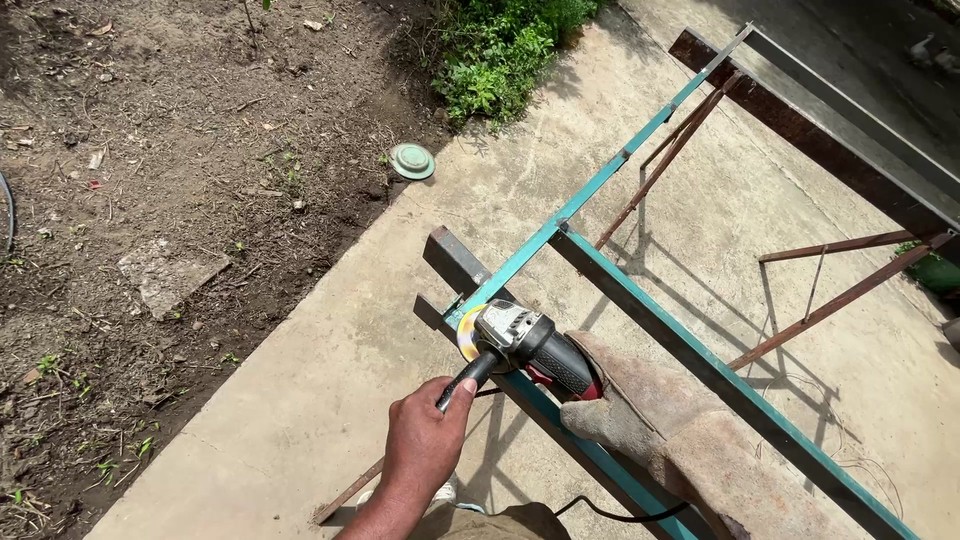} \\
Hotels \& accommodation & Woodworking \& furniture & Textile \& craft arts & Technology \& computing & Metalwork \& welding \\[1.5pt]
\includegraphics[width=\linewidth]{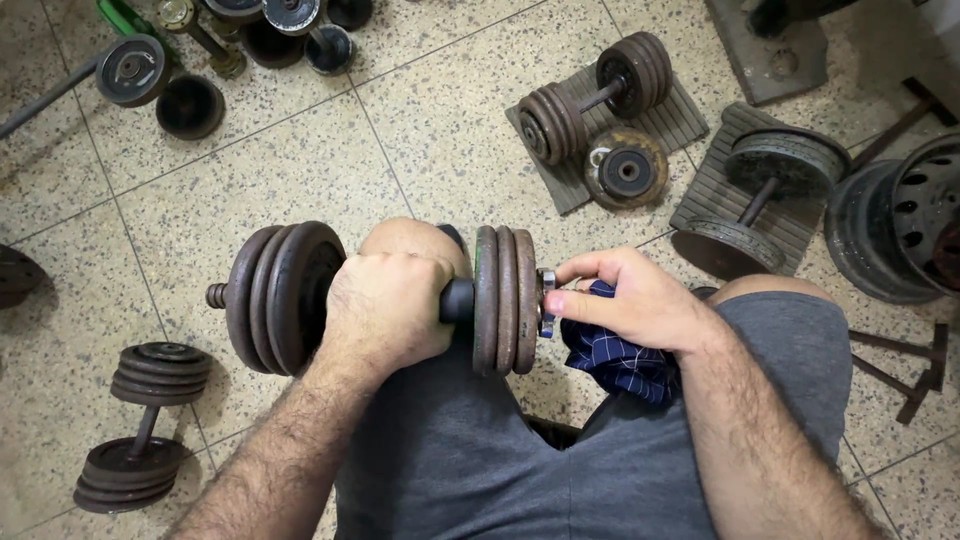} & \includegraphics[width=\linewidth]{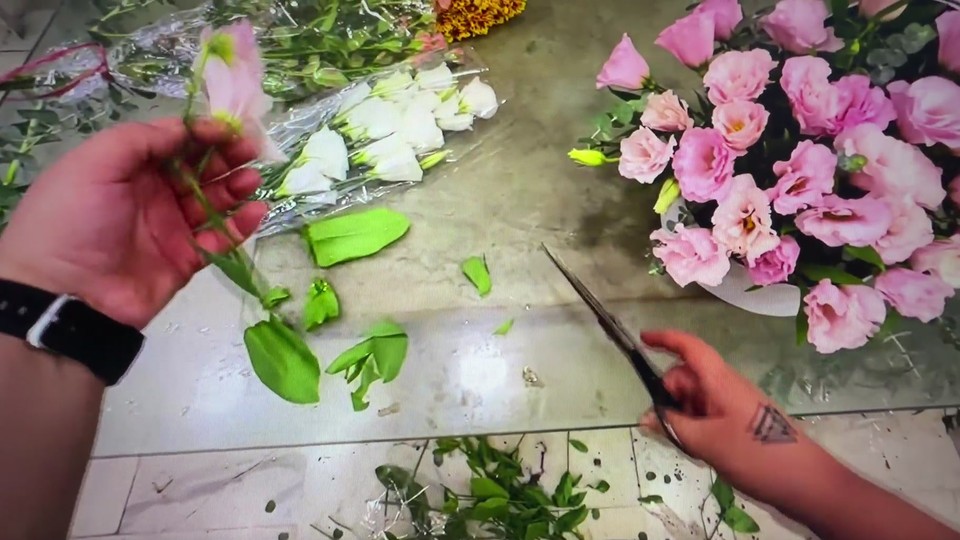} & \includegraphics[width=\linewidth]{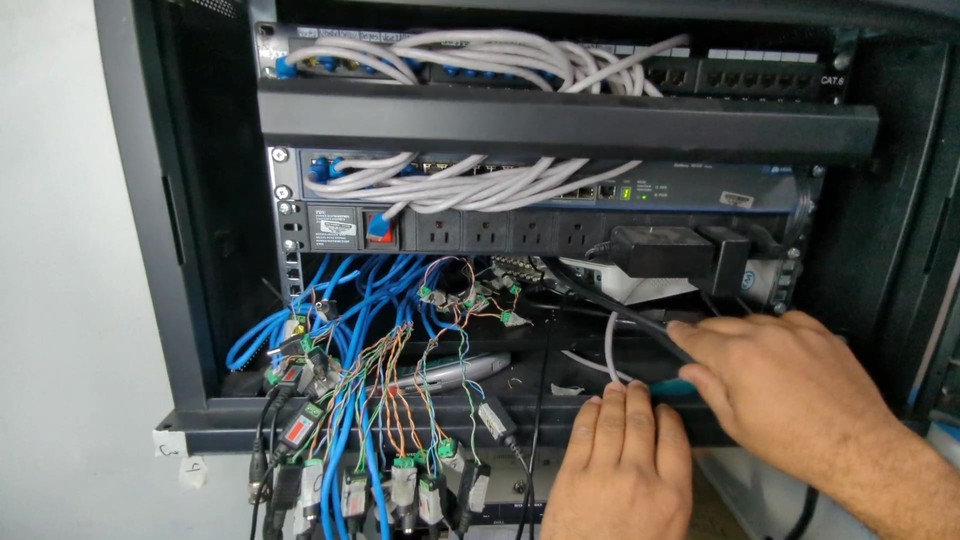} & \includegraphics[width=\linewidth]{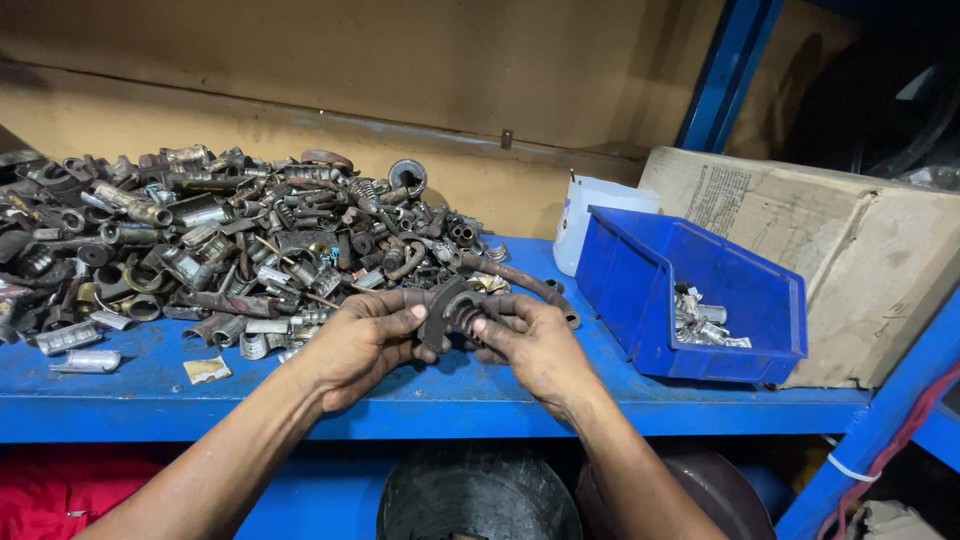} & \includegraphics[width=\linewidth]{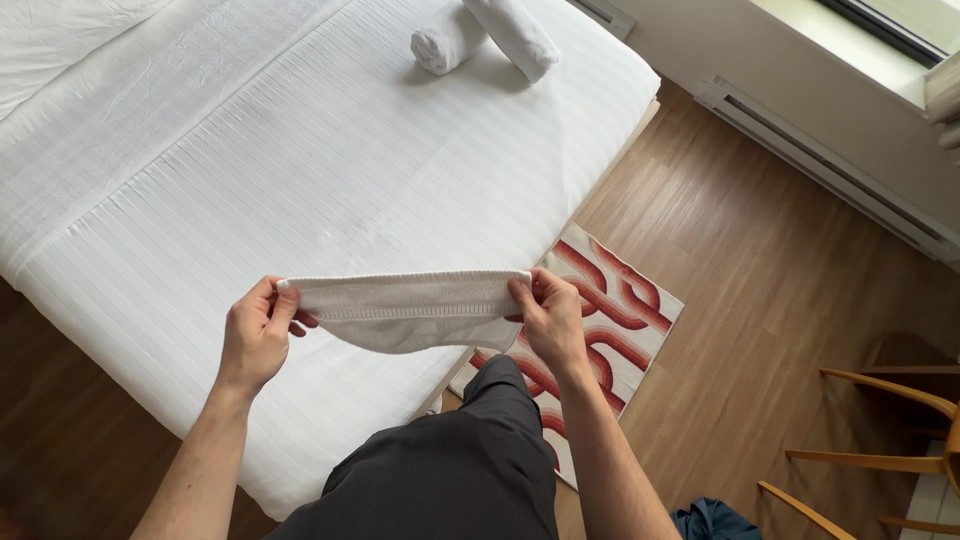} \\
Gyms \& sports facilities & Florists \& garden centres & Offices \& IT & Recycling \& waste & Daily chores \\
\end{tabular}
\caption{One frame from each of the 30 work categories of \hub{} v1.0, ordered row by row by hours per work category. All 30 episodes were recorded in workplaces; 15 are labeled hard and 15 medium.}
\label{fig:catgrid}
\end{figure*}%
\begin{figure}[t]
\centering
\includegraphics[width=\columnwidth]{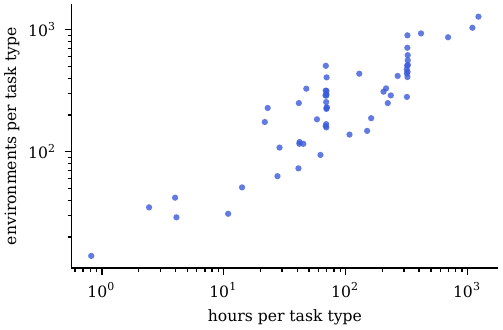}
\caption{Environments per task type against hours per task type (56 points, log scales).}
\label{fig:taskenv}
\end{figure}%
\begin{figure*}[t]
\centering
\includegraphics[width=\textwidth]{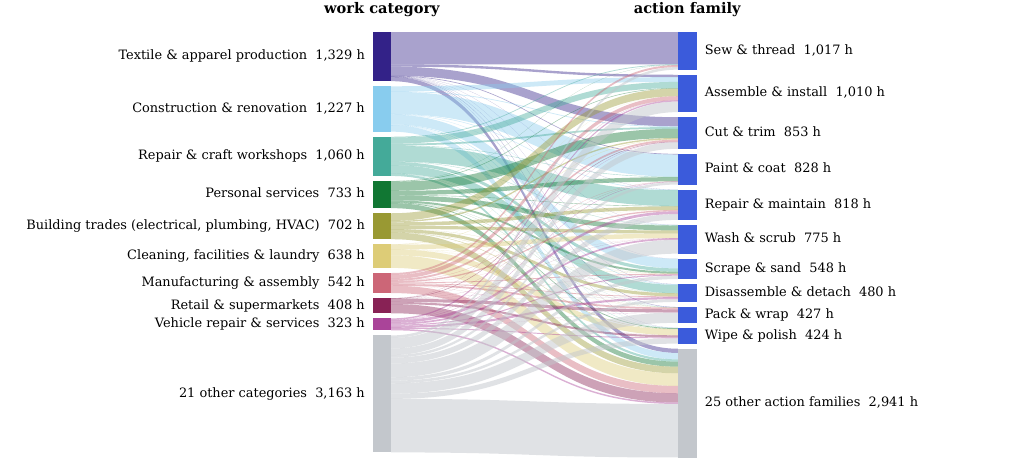}
\caption{Hours from work categories (left) to action families (right); band width is proportional to hours. The nine work categories and ten action families with the most hours are shown individually; the remaining 21 work categories and 25 action families are grouped (gray).}
\label{fig:taxonomy}
\end{figure*}%
\begin{figure}[t]
\centering
\includegraphics[width=\columnwidth]{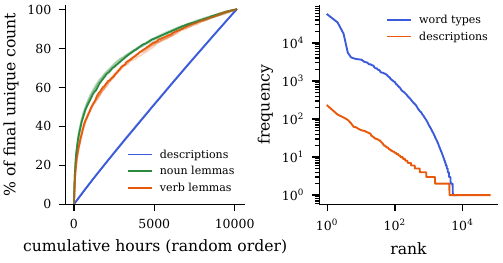}
\caption{Left: unique descriptions, noun lemmas and verb lemmas as a share of their final count against cumulative hours, over five random episode orders. Right: rank-frequency of word types and of descriptions (log scales).}
\label{fig:lang}
\end{figure}%
\begin{figure*}[t]
\centering
\includegraphics[width=0.92\textwidth]{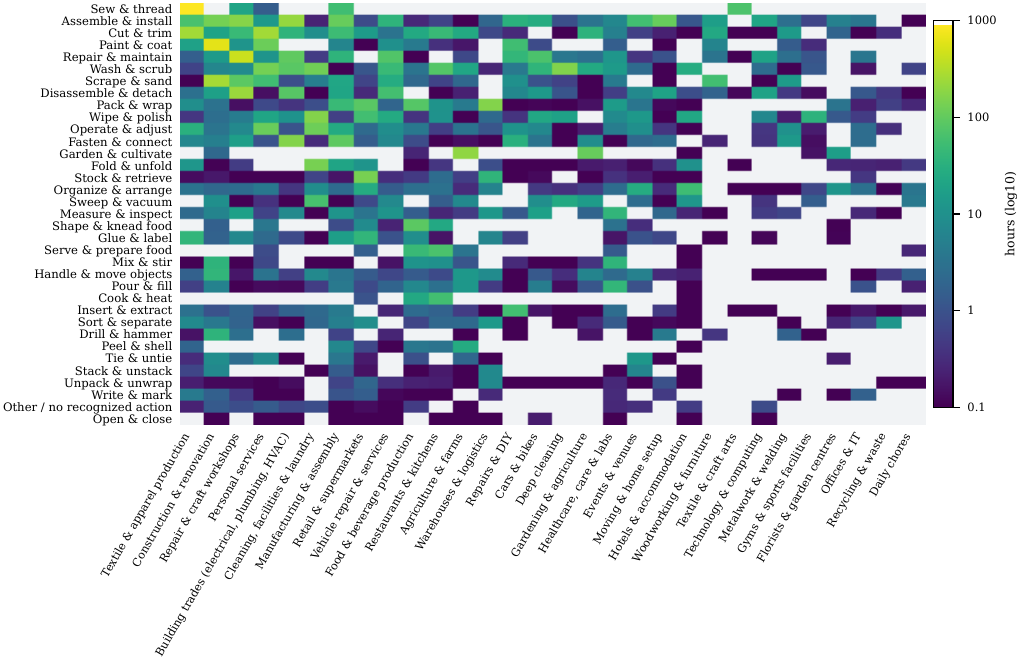}
\caption{Hours per action family (rows) and work category (columns), log scale. Light gray cells contain no data. Rows and columns are sorted by total hours.}
\label{fig:afcat}
\end{figure*}%
\begin{figure*}[t]
\centering
\includegraphics[width=\textwidth]{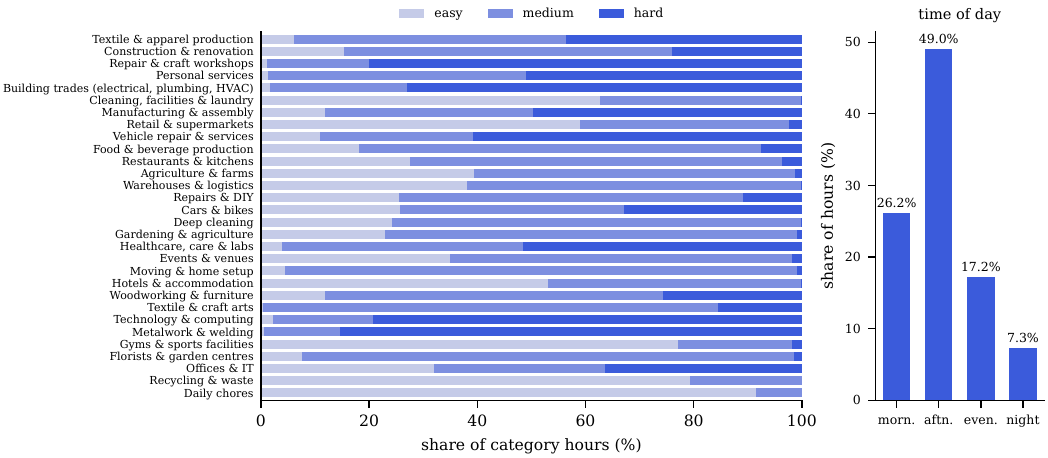}
\caption{Left: difficulty mix per work category, sorted by category hours. Right: share of hours by local time of day.}
\label{fig:diff}
\end{figure*}%
\begin{figure}[t]
\centering
\includegraphics[width=\columnwidth]{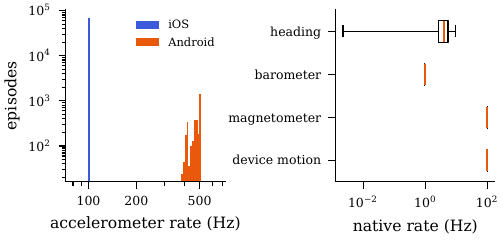}
\caption{Left: native accelerometer rate per episode by operating system (log scales). Right: native rates of the device-motion, magnetometer, barometer and heading streams (box: quartiles; whiskers: 1.5 interquartile range).}
\label{fig:imu}
\end{figure}
 VC-1 finds that adding a different kind of data (navigation) to Ego4D helps more than adding more manipulation data \citep{majumdar2023vc1}, and EgoVLA reports that increasing the scale and diversity of human pretraining data ``consistently improves downstream performance'' \citep{yang2025egovla}. In the opposite direction, at a fixed budget of 200K clips, \citet{long2026egoguidance} did not observe a consistent gain from increasing scene, object or interaction diversity alone.

\textbf{Measured in \hub{}.} See Sections~\ref{sec:tasks}, \ref{sec:language} and~\ref{sec:difficulty}.

\subsection{Contributor Diversity}
\label{sec:why-contrib}

\textbf{Evidence.} The evidence on the diversity of demonstrators is less direct than for environments. Ego4D made diversity of ``people, places, objects, and activities'' an explicit design goal and recruited 931 camera wearers, including occupations such as bakers, carpenters, landscapers and mechanics \citep{grauman2022ego4d}. For robot imitation at small scale, the robomimic study \citep{mandlekar2021robomimic} finds that 300 demonstrations from six operators of varying proficiency yield lower success than 200 demonstrations from one experienced operator, and \citet{belkhale2023dataquality} find that less consistent actions at a given state are often associated with lower policy success. BC-Z \citep{jang2021bcz} similarly observes that pooling data from several robots and operators makes a single task harder to learn, and that pooling across many tasks overcomes this. No controlled study has yet varied the number of human demonstrators at fixed hours for egocentric pretraining; the per-contributor identifiers of \hub{} make such a study possible, for example with contributor-disjoint splits.

\textbf{Measured in \hub{}.} See Section~\ref{sec:diversity}.

\subsection{Inertial Sensing}
\label{sec:why-imu}

\textbf{Evidence.} In Ego4D, 836 of the 3,670 hours include IMU data \citep{grauman2022ego4d}. MMG-Ego4D \citep{gong2023mmg} builds a video, audio and IMU benchmark from about 202 hours of Ego4D and notes that video consumes 60 times more storage and 43 times more compute than IMU. IMU2CLIP \citep{moon2023imu2clip} aligns IMU with video and text and raises IMU-based activity recognition on Ego4D from 23.23 to 45.15 F1. Project Aria glasses \citep{engel2023aria}, used by Ego-Exo4D \citep{grauman2024egoexo4d}, record two IMUs at 800 and 1000 Hz with a magnetometer and a barometer. Visual-inertial estimators such as VINS-Mono \citep{qin2018vinsmono} establish that a monocular camera with a low-cost IMU is the minimum sensor suite for metric six-degree-of-freedom state estimation, and head motion has been used as an intermediate representation for body pose \citep{li2023egoego}. Inertial data costs little to store and supports ego-motion estimation; its effect on robot policies has not been studied yet.

\textbf{Measured in \hub{}.} See Section~\ref{sec:imu}.

\section{Dataset Analysis}
\label{sec:analysis}

This section reports the statistics of v1.0, including the measurements of Table~\ref{tab:evidence}. They are computed from the released tables and are hours-weighted unless noted. It starts with diversity, the main property of the dataset.

\begin{table}[t]
\centering
\small
\caption{Diversity axes of \hub{} v1.0 and the spread of hours over each. $K$: number of distinct values; $e^{H}$: effective number; $J$: evenness; $D^{-1}$: inverse Simpson index.}
\label{tab:div}
\renewcommand{\arraystretch}{1.12}\setlength{\tabcolsep}{4.5pt}
\begin{tabular}{lIrIrIrIr}
\thickrule
\headrow \textbf{Axis} & $\boldsymbol{K}$ & $\boldsymbol{e^{H}}$ & $\boldsymbol{J}$ & $\boldsymbol{D^{-1}}$ \\
\headrule
Contributors & 12,227 & 5,666 & 0.92 & 3,385 \\
\thinrule
Environments & 15,548 & 7,244 & 0.92 & 4,197 \\
\thinrule
Setting types & 11 & 7.7 & 0.85 & 6.2 \\
\thinrule
Work categories & 30 & 18.9 & 0.86 & 14.4 \\
\thinrule
Task types & 56 & 31.0 & 0.85 & 20.9 \\
\thinrule
Action families & 35 & 20.8 & 0.85 & 16.4 \\
\thinrule
Skill groups & 18 & 11.3 & 0.84 & 9.4 \\
\thinrule
Phone models & 56 & 15.8 & 0.69 & 9.7 \\
\thickrule
\end{tabular}
\end{table}

\subsection{Diversity}
\label{sec:diversity}

Table~\ref{tab:div} and Fig.~\ref{fig:diversity}a gather the axes along which v1.0 varies. For contributors and places we also quote the values that the other datasets in Table~\ref{tab:positioning} report.

\paragraph{Contributors.} v1.0 was recorded by 12,227 contributors (Fig.~\ref{fig:glance}d), with a median of 250.5 contributors per task type. No contributor accounts for more than 0.22\% of hours, and 2,824 contributors recorded one hour or more. The effective number of contributors (exponential of the Shannon entropy of the hours distribution) is 5,666 and the inverse Simpson index is 3,385. The median contributor recorded 21.3 minutes, the mean is 49.7 minutes, the 90th percentile 2.1 hours and the maximum 22.3 hours (Fig.~\ref{fig:lengths}, right). Contributors record on average 1.33 task types and 1.27 setting types, and 21.7\% of contributors appear in more than one setting type. Among the datasets in Table~\ref{tab:positioning}, Ego4D reports 931 camera wearers, Ego-Exo4D 740 participants, EgoVerse 2,087 demonstrators and DROID 50 data collectors \citep{grauman2022ego4d,grauman2024egoexo4d,egoverse2026,khazatsky2024droid}.

\paragraph{Environments.} Each contributor records in the wild, in their own workplace or home. v1.0 contains 15,548 environments across 11 setting types (Section~\ref{sec:settings}). The median task type is recorded in 268 environments, and 47 of the 56 task types have at least 100 (Fig.~\ref{fig:taskenv}); the median work category has 489 and the minimum is 51. Among the datasets in Table~\ref{tab:positioning}, Ego4D reports 74 locations in 9 countries and Ego-Exo4D 123 scene contexts in 13 cities; EPIC-KITCHENS-100 was recorded in 45 kitchens in 4 cities, HOI4D in 610 indoor rooms, EgoVerse in 240 scenes and DROID in 564 scenes in 52 buildings \citep{grauman2022ego4d,grauman2024egoexo4d,damen2022epic100,liu2022hoi4d,egoverse2026,khazatsky2024droid}. Each source defines its unit of place in its own way; ours is defined in Section~\ref{sec:dataset}.

\paragraph{Work, tasks and actions.} The recordings span 30 work categories and 56 task types, all unscripted, and the annotation model assigns 35 action families and 18 skill groups. A task type names a kind of job, a broader unit than the single manipulation tasks counted by task-oriented datasets, which report 20 guided tasks for HoloAssist, 194 tabletop tasks for EgoDex and 1,965 tasks for EgoVerse \citep{wang2023holoassist,hoque2026egodex,egoverse2026}; within a task type, the finer grain is carried by the descriptions.

\paragraph{Language, objects and devices.} The verb and noun vocabulary of the descriptions has a long tail (Section~\ref{sec:language}), and its nouns reflect the diversity of objects. The footage comes from 56 phone models, which behave like 15.8 equally used ones.

\paragraph{Evenness.} To summarize how evenly hours are spread along each axis, Table~\ref{tab:div} reports the effective number $e^{H}$, where $H=-\sum_i p_i \ln p_i$ is the Shannon entropy of the hours distribution (the number of equally sized classes with the same entropy), Pielou's evenness $J=H/\ln K$ and the inverse Simpson index $1/\sum_i p_i^2$. Evenness lies between 0.84 and 0.86 for work categories, task types, action families, skill groups and setting types, and reaches 0.92 for contributors and environments. For task types this is partly by design, since hours are rebalanced across task types (Section~\ref{sec:curation}). The 56 task types behave like 31 equally sized ones, the 35 action families like 20.8 and the 30 work categories like 18.9.

\paragraph{New contributors and words inside v1.0.} Inside v1.0, with episodes in random order, new contributors and words keep appearing as hours accumulate (Fig.~\ref{fig:diversity}b). Half of the hours contain 79.7\% of the contributors and 75.7\% of the environments, and the last 10\% of hours adds 356 contributors and 570 environments (mean of five orders). Section~\ref{sec:language} gives the same measurement for verbs and nouns.

\subsection{Tasks and Actions}
\label{sec:tasks}

Fig.~\ref{fig:catgrid} shows one frame from each of the 30 work categories, and Fig.~\ref{fig:cat} gives the hours per category. Eighteen work categories each hold more than 200 hours and the top one holds 13.1\%; 52 of the 56 task types hold at least 10 hours each. Fig.~\ref{fig:taxonomy} shows how hours flow from work categories to action families: each large category spreads over many action families (26 action families with at least one hour in Construction \& renovation, 22 in Manufacturing \& assembly, 18 in Textile \& apparel production), and each large action family draws on many categories.

\subsection{Setting Types}
\label{sec:settings}

Fig.~\ref{fig:settings} shows the 11 setting types by hours together with their numbers of environments. The household setting accounts for 17.4\% of hours (Fig.~\ref{fig:glance}a). The remaining 82.6\% are workplace settings: 55.7\% in dedicated workplaces, led by workshops and repair benches (16.9\%) and construction and renovation sites (13.0\%), and 26.9\% as trades done in homes (a home sewing table or repair bench, or a tradesperson working in a client's home). Across task types, the number of environments is roughly proportional to hours (Fig.~\ref{fig:taskenv}).

\subsection{Actions Across Work Categories}
\label{sec:afcat}

Action families recur across work categories (Fig.~\ref{fig:afcat}). Of the 1,050 cells of the action family by category matrix, 663 contain data and 331 contain at least one hour. Counting only cells with at least one hour, Assemble \& install appears in 20 categories, Wash \& scrub in 19 and Cut \& trim in 17, and 16 of the 35 action families appear in 10 or more categories. The top action family holds 10.0\% of hours and the top six cover 52.4\%. The mutual information between action family and category is 1.27 nats, 42\% of the action-family entropy, so knowing the category leaves an effective 5.8 action families per category. The same action family is recorded in many trades, each with its own tools, materials and environments.

\subsection{Language}
\label{sec:language}

The annotation model writes one free-text description per episode, so most episodes receive their own description (the 72,754 episodes carry 62,605 distinct descriptions, 93.6\% of which occur once; 80.5\% of episodes have a description shared with no other episode), and we measure breadth on their vocabulary rather than on the number of distinct descriptions. Verb and noun lemmas are extracted from the descriptions with a part-of-speech tagger (spaCy \texttt{en\_core\_web\_sm}): the descriptions use 1,429 verb lemmas and 3,709 noun lemmas, of which 685 verbs and 2,110 nouns appear in at least five descriptions. The most frequent verbs are trim, wipe, assemble, cut, apply and pack, and the most frequent nouns wall, car, floor, machine and brush. Word frequencies follow a Zipf law with slope $-0.99$ between ranks 10 and 1,000 (Fig.~\ref{fig:lang}, right). To measure how quickly new content appears inside v1.0, we shuffle episodes at random and count distinct descriptions, verbs and nouns as hours accumulate (Fig.~\ref{fig:lang}, left, shows five such orders). The number of distinct descriptions rises almost linearly along such an order (a power-law fit beyond the first 100 hours gives exponent 0.97, and new descriptions appear at 5.8 per hour after 90\% of hours), as expected for free text written per episode. Verb and noun lemmas accumulate sublinearly (exponents 0.34 and 0.29), and new words keep appearing: the last 10\% of hours adds about 39 verb lemmas and about 84 noun lemmas on average over 20 random episode orders.

\subsection{Difficulty}
\label{sec:difficulty}

The annotation model labels 34.5\% of hours hard and 47.3\% medium; in seven work categories, more than half of the hours are labeled hard (Fig.~\ref{fig:diff}). Episodes labeled hard cluster in the trades: 85.4\% of hours are labeled hard in Metalwork \& welding, 80.0\% in Repair \& craft workshops, 79.3\% in Technology \& computing and 73.0\% in the building trades, against less than 5\% in categories such as retail, kitchens and warehouses. Among action families, 87.0\% of Repair \& maintain hours and 78.3\% of Fasten \& connect hours are labeled hard.

\subsection{Episodes and Recordings}
\label{sec:episodes}

v1.0 contains 10,124 hours and 1,093,194,072 frames of $1920\times1080$ video at 30 fps, recorded from the contributor's head with an ultra-wide lens (Section~\ref{sec:collection}). Episodes are long: the median is 4.9 minutes, the 90th percentile 21.4 minutes, and 72.8\% of hours lie in episodes longer than 10 minutes (Fig.~\ref{fig:lengths}, left). Episodes cut from one recording share a recording identifier and carry their start and end time within it, so longer procedures can be reassembled. The step-level instruction of each episode provides a language target of the kind used in VLA pretraining.

\subsection{Hand Visibility and On-Task Share}
\label{sec:hands}

Hand visibility and on-task share are model estimates: hand visibility is the share of sampled frames in which the model panel of the curation pipeline reports a hand, and the on-task share comes from the temporal segmentation model (Section~\ref{sec:curation}). The pipeline keeps the segments in which the contributor's hands perform the task. Over all hours, at least one hand appears in a model-estimated 94.7\% of sampled frames and both hands in 76.3\% (Fig.~\ref{fig:glance}b); for the median episode the share with at least one hand is 95.8\%. 70.6\% of episodes, holding 83.0\% of hours, show a hand in at least 90\% of their sampled frames. The model-estimated on-task share is 96.9\% hours-weighted (median episode 98.9\%; 39.1\% of episodes are estimated to be entirely on task), and 92.2\% of hours lie in episodes with an on-task share of at least 90\% (Fig.~\ref{fig:glance}c). Both estimates are stored per episode, so users can apply stricter thresholds. The model-estimated share of sampled frames with both hands varies by action: 91.0\% for Peel \& shell, 87.2\% for Fasten \& connect and 86.3\% for Sew \& thread, against 52.2\% for Sweep \& vacuum.

\subsection{Inertial Data}
\label{sec:imu}

All 72,754 episodes include inertial data. The native accelerometer and gyroscope streams, measured as samples divided by episode duration, have a median rate of 99.98 Hz (10th to 90th percentile 99.43 to 100.54 Hz) and 95.6\% of episodes lie between 95 and 105 Hz; iOS devices log at 100 Hz, and episodes recorded on Android devices log at a median of about 497 Hz (Fig.~\ref{fig:imu}). The release contains about 4.28 billion accelerometer samples. Magnetometer and device-motion streams run at about 100 Hz, heading at about 4 Hz and the barometer at about 0.9 Hz; device motion is present in every episode, magnetometer and heading in 99.1\% and the barometer in 93.6\% of episodes. The LeRobot \texttt{observation.state} holds the accelerometer and gyroscope reading resampled to each frame's capture time (Section~\ref{sec:format}).

\section{Applications}
\label{sec:uses}

GR00T N1 organizes its training data as a ``data pyramid'' \citep{nvidia2025gr00t}: web data and human video form the base layer, which provides ``broad visual and behavioral priors'', synthetic data the middle, and real-robot data the top, which grounds the model in real-robot execution. \hub{} is intended for that base layer. The data is action-free; established ways to use such data for policy learning are co-training with a small robot dataset, pretraining on human data followed by fine-tuning on robot data, and training a visual encoder on the human data \citep{hoque2026egodex}. A contributor-disjoint train, validation and test split, stratified by task type, is available on request, so that evaluation measures generalization to unseen contributors and environments.
\begin{itemize}
\item \textbf{Visual representation pretraining} for manipulation, following R3M, MVP and VC-1 \citep{nair2022r3m,radosavovic2022mvp,majumdar2023vc1}, evaluated on established robot benchmarks.
\item \textbf{Hand and hand-object pose.} Pseudo-labeling 3D hands with HaMeR \citep{pavlakos2024hamer} as MANO parameters \citep{romero2017mano} and contact with the 100DOH detector \citep{shan2020hands}, which in turn supports affordance learning \citep{bahl2023vrb,srirama2024hrp} and hand-motion pretraining \citep{luo2025beingh0}.
\item \textbf{Action recognition} over the 35 action families or the 56 task types, with video, IMU or both \citep{gong2023mmg,moon2023imu2clip}.
\item \textbf{VLA pretraining and co-training} with latent actions \citep{ye2025lapa,nvidia2025gr00t}, future visual-state prediction \citep{long2026egoguidance} or human-to-robot co-training \citep{kareer2025egomimic,pi2025humantransfer}, using the step-level instructions as language targets.
\item \textbf{Video prediction and world models.} Minutes-long episodes of real work with step-level language can serve language-conditioned video prediction and world modeling, a use that EgoDex also names for egocentric manipulation data \citep{hoque2026egodex}; GR00T N1 uses video generation models to synthesize additional training trajectories \citep{nvidia2025gr00t}.
\item \textbf{Head motion and visual-inertial odometry} from the aligned camera and IMU \citep{qin2018vinsmono}. The catalog gives the lens and diagonal field of view of every episode.
\end{itemize}

\section{Broader Impact}
\label{sec:impact}

\hub{} is meant to help robots learn manual skills from people who practice them. Models trained on recordings of trade work may automate parts of the trades shown. Contributors agreed to this use of their recordings, and the privacy measures of Section~\ref{sec:privacy} are applied to every episode before release. After rebalancing, the task mix of v1.0 does not reflect how often each task occurs in the world, so models trained on it should be evaluated in the conditions in which they will be used.

\section{Conclusion}
\label{sec:conclusion}

We presented \hub{}, an in-the-wild egocentric dataset of skilled work whose main property is diversity, and described its first release, v1.0: 10,124 hours of unscripted work recorded by 12,227 contributors in 15,548 environments across 11 setting types, 30 work categories and 56 task types. We measured this diversity axis by axis; inside v1.0, with episodes in random order, new contributors and words keep appearing. We related the design of the dataset to published evidence on environment and object diversity, egocentric human video, hand visibility, task breadth, contributor diversity and inertial sensing, and reported the corresponding measurement for each property from the released tables. The dataset, its catalog and its annotations are available under CC~BY~4.0.

\paragraph{Discussion and future work.} Subsequent releases will extend \hub{} to 100,000 hours (Section~\ref{sec:release}). The release layout is built to grow: later releases can add other capture setups as sibling folders of \texttt{mono/} (Section~\ref{sec:format}), and annotation layers such as 3D hand pose and hand-object contact (Section~\ref{sec:uses}) can be added per episode under the same \texttt{episode\_index} key. Our next step is to train baselines on the data: visual encoders pretrained on \hub{} and policies co-trained with robot data, evaluated on contributor-disjoint splits, to measure how the diversity reported here carries over to policy performance. Later versions of this paper will add these experiments and cover subsequent releases.

\section*{Acknowledgments}
We thank the 12,227 contributors who recorded their work for this release.

\appendix
\section{Datasheet for \hub{}}
\label{app:datasheet}

This datasheet follows the questions of \citet{gebru2021datasheets} and describes \hub{} v1.0 (10,124 hours), the first release of a dataset planned to reach 100,000 hours.

\subsection*{Motivation}

\dsq{For what purpose was the dataset created?}
Robots that manipulate objects in homes and workplaces need training data that covers the variety of those places, objects and tasks. Existing egocentric datasets focus mostly on daily-life activity, kitchens, or scripted and instrumented tasks. \hub{} was created to provide egocentric video of skilled manual work (for example bricklaying, wiring, plumbing, garment sewing, microsoldering, hair cutting or engine repair), recorded by contributors in their own workplaces and homes. It is intended for research and development of perception, representation learning, action understanding and robot learning systems (Section~\ref{sec:uses}).

\dsq{Who created the dataset and on behalf of which entity?}
Hub Labs (Hub Data, Inc., San Francisco, CA, USA). The authors of this paper are listed on its first page. All recordings are first-party: every hour was recorded by a Hub contributor, and the release contains no third-party episodes.

\dsq{Who funded the creation of the dataset?}
Hub Data, Inc.

\subsection*{Composition}

\dsq{What do the instances that comprise the dataset represent?}
An instance is an \emph{episode}: a continuous segment of head-mounted, egocentric video in which a contributor performs a named manual task, together with the phone's sensor data and a set of annotations. Episodes are cut from \emph{recordings}; one recording can yield several episodes. Recordings and contributors are referenced by identifier (\texttt{recording\_id}, \texttt{contributor\_id}).

\dsq{How many instances are there in total?}
v1.0 contains 72,754 episodes from 41,688 recordings (1.75 episodes per recording on average), for 10,124 hours and 1,093,194,072 frames, recorded by 12,227 contributors. We count 15,548 environments, defined as distinct pairs of contributor and setting type (Section~\ref{sec:envdef}). The video totals 43.5 TB and the native sensor logs 1.29 TB.

\dsq{Does the dataset contain all possible instances or is it a sample of instances from a larger set?}
It is a curated subset of what contributors recorded. The curation pipeline keeps the segments in which the contributor's hands perform the task, privacy filtering removes content (see Preprocessing), and hours are rebalanced across task types so that no task type dominates.

\dsq{What data does each instance consist of?}
One MP4 video ($1920\times1080$, H.264, 30 fps, monocular), the accelerometer and gyroscope reading at each frame in LeRobot v3 parquet, the native sensor streams in MCAP, one row in \texttt{catalog.parquet} and one row in \texttt{meta/annotations.parquet} (Section~\ref{sec:format}; fields in Appendix~\ref{app:fields}). Episodes last at most 30 minutes (median 4.9 minutes).

\dsq{Is there a label or target associated with each instance?}
Yes. Each episode has a task type (56 types) in a work category (30 categories); the task type is the task the contributor was asked to record. The annotation model adds, by dense video annotation, a short free-text description, a step-level instruction, an action family (35 families), a skill group (18 groups), a difficulty level (easy, medium or hard), a setting type (11 types) and the share of the episode spent on task. \texttt{meta/subtasks.parquet} and \texttt{meta/subtask\_occurrences.parquet} store the instruction of each episode in the LeRobot subtask layout.

\dsq{Is any information missing from individual instances?}
Four of the 72,754 episodes have empty annotation fields, 22 have no setting type and 209 have no time-of-day label. All episodes include accelerometer, gyroscope and device-motion data; the magnetometer and heading streams are present in 99.1\% of episodes and the barometer in 93.6\%. Audio, location, device identifiers and contributor identity were removed on purpose (see Preprocessing).

\dsq{Are relationships between individual instances made explicit?}
Yes. All tables are keyed by \texttt{episode\_index}, the same identifier used in the LeRobot data and the videos. \texttt{recording\_id} links episodes cut from the same recording, and \texttt{recording\_start\_s} and \texttt{recording\_end\_s} give the position of the episode inside it. \texttt{contributor\_id} is stable across episodes and links all episodes of one contributor.

\dsq{Are there recommended data splits?}
We recommend contributor-disjoint train, validation and test splits, so that evaluation measures generalization to unseen contributors and environments; the identifiers needed to build them are in the catalog. A contributor-disjoint split stratified by task type is available on request.

\dsq{Are there any errors, sources of noise, or redundancies in the dataset?}
(i) The annotations are machine-generated and have not been audited by human reviewers; hand visibility and on-task share are model estimates. (ii) 5.5\% of episodes contain at least one frame interval longer than 1.5 times the nominal interval; the median absolute drift against the nominal frame rate is 0.017 s over an episode and the 95th percentile 0.155 s; re-encoded episodes can contain duplicated frames. All three are logged per episode in the catalog. (iii) Native inertial rates differ by operating system: about 100 Hz on iOS devices and a median of about 497 Hz on Android devices. (iv) Recordings come from 56 phone models, so the field of view varies across episodes, and camera calibration is not provided. (v) Video and inertial data share one timeline (Section~\ref{sec:curation}); the residual time offset between the camera and the inertial sensors has not been measured. Redundancy: several episodes can come from one recording, and 6.4\% of the distinct descriptions are used by more than one episode.

\dsq{Is the dataset self-contained, or does it link to or otherwise rely on external resources?}
It is self-contained and hosted in one repository (\url{https://huggingface.co/datasets/HubLabs/HubEgo-100K}).%
\begin{table*}[t]
\centering
\scriptsize
\renewcommand{\arraystretch}{0.92}
\setlength{\tabcolsep}{3pt}
\begin{minipage}[t]{0.67\textwidth}
\centering
\caption{The 35 action families of \hub{} v1.0, sorted by hours. Ep.: episodes. Cat.: number of work categories in which the family has at least one hour.}
\label{tab:families}
\vspace{2pt}
\begin{tabular}[t]{@{}lIrIrIr@{}}
\thickrule
\headrow \textbf{Action family} & \textbf{Hours} & \textbf{Ep.} & \textbf{Cat.} \\
\headrule
Sew \& thread & 1,017 & 4,077 & 5 \\
\thinrule
Assemble \& install & 1,010 & 6,138 & 20 \\
\thinrule
Cut \& trim & 853 & 5,638 & 17 \\
\thinrule
Paint \& coat & 828 & 4,925 & 11 \\
\thinrule
Repair \& maintain & 818 & 4,172 & 17 \\
\thinrule
Wash \& scrub & 775 & 5,869 & 19 \\
\thinrule
Scrape \& sand & 548 & 3,720 & 12 \\
\thinrule
Disassemble \& detach & 480 & 3,363 & 14 \\
\thinrule
Pack \& wrap & 427 & 3,635 & 12 \\
\thinrule
Wipe \& polish & 424 & 3,968 & 17 \\
\thinrule
Operate \& adjust & 362 & 3,838 & 18 \\
\thinrule
Fasten \& connect & 360 & 2,441 & 15 \\
\thinrule
Garden \& cultivate & 317 & 2,038 & 4 \\
\thinrule
Fold \& unfold & 193 & 1,254 & 5 \\
\thinrule
Stock \& retrieve & 182 & 1,684 & 3 \\
\thinrule
Organize \& arrange & 165 & 2,085 & 17 \\
\thinrule
Sweep \& vacuum & 158 & 1,472 & 12 \\
\thinrule
Measure \& inspect & 137 & 1,732 & 14 \\
\thickrule
\end{tabular}\hfill
\begin{tabular}[t]{@{}lIrIrIr@{}}
\thickrule
\headrow \textbf{Action family} & \textbf{Hours} & \textbf{Ep.} & \textbf{Cat.} \\
\headrule
Shape \& knead food & 131 & 1,071 & 6 \\
\thinrule
Glue \& label & 125 & 876 & 8 \\
\thinrule
Serve \& prepare food & 121 & 1,123 & 5 \\
\thinrule
Mix \& stir & 98 & 869 & 5 \\
\thinrule
Handle \& move objects & 98 & 1,615 & 14 \\
\thinrule
Pour \& fill & 84 & 947 & 9 \\
\thinrule
Cook \& heat & 80 & 694 & 3 \\
\thinrule
Insert \& extract & 79 & 1,060 & 9 \\
\thinrule
Sort \& separate & 62 & 614 & 11 \\
\thinrule
Drill \& hammer & 50 & 481 & 5 \\
\thinrule
Peel \& shell & 43 & 203 & 5 \\
\thinrule
Tie \& untie & 39 & 386 & 6 \\
\thinrule
Stack \& unstack & 23 & 242 & 4 \\
\thinrule
Unpack \& unwrap & 13 & 242 & 2 \\
\thinrule
Write \& mark & 12 & 196 & 5 \\
\thinrule
Other / no recognized action & 7.8 & 45 & 2 \\
\thinrule
Open \& close & 0.5 & 37 & 0 \\
\thickrule
\end{tabular}
\end{minipage}\hfill
\begin{minipage}[t]{0.31\textwidth}
\centering
\caption{The 18 skill groups of \hub{} v1.0, sorted by hours.}
\label{tab:skills}
\vspace{2pt}
\begin{tabular}[t]{@{}lIr@{}}
\thickrule
\headrow \textbf{Skill group} & \textbf{Hours} \\
\headrule
Tool Use \& Technical Manipulation & 1,797 \\
\thinrule
Clothing \& Laundry & 1,502 \\
\thinrule
Construction \& Building & 1,282 \\
\thinrule
Cleaning \& Sanitation & 1,166 \\
\thinrule
Assembly \& Installation & 851 \\
\thinrule
Repair \& Maintenance & 704 \\
\thinrule
Electronics \& Diagnostics & 626 \\
\thinrule
Food Preparation \& Cooking & 507 \\
\thinrule
Mechanical / Automotive Work & 488 \\
\thinrule
Packing \& Bagging & 436 \\
\thinrule
Inventory \& Stock Management & 272 \\
\thinrule
Healthcare \& Caregiving & 123 \\
\thinrule
Organization \& Tidying & 100 \\
\thinrule
Retail \& Service Operations & 97 \\
\thinrule
Pick and Place / Object Handling & 79 \\
\thinrule
Dish Handling & 60 \\
\thinrule
Home Appliance Interaction & 34 \\
\thinrule
Human Interaction \& Handoffs & 0.1 \\
\thickrule
\end{tabular}
\end{minipage}
\end{table*}%
\begin{table*}[t]
\centering
\scriptsize
\renewcommand{\arraystretch}{0.90}
\caption{The 56 task types of \hub{} v1.0, sorted by hours, with their work category, episodes, contributors (Contrib.) and environments (Env.).}
\label{tab:tasks}
\vspace{2pt}
\setlength{\tabcolsep}{3.2pt}
\begin{tabular}{@{}lIlIrIrIrIr@{}}
\thickrule
\headrow \textbf{Task type} & \textbf{Work category} & \textbf{Hours} & \textbf{Episodes} & \textbf{Contrib.} & \textbf{Env.} \\
\headrule
Building site \& renovation crew & Construction \& renovation & 1,227 & 8,317 & 1,247 & 1,276 \\
\thinrule
Garment \& shoe workshop & Textile \& apparel production & 1,094 & 4,752 & 868 & 1,035 \\
\thinrule
Building systems trades & Building trades (electrical, plumbing, HVAC) & 691 & 3,293 & 690 & 865 \\
\thinrule
Salon, barber \& nail services & Personal services & 415 & 5,218 & 927 & 930 \\
\thinrule
Phone \& electronics repair bench & Repair \& craft workshops & 323 & 1,538 & 418 & 513 \\
\thinrule
Auto shop \& fuel station work & Vehicle repair \& services & 323 & 4,622 & 474 & 560 \\
\thinrule
Shop floor \& stockroom work & Retail \& supermarkets & 323 & 2,280 & 520 & 617 \\
\thinrule
Food \& drink production line & Food \& beverage production & 322 & 2,653 & 363 & 451 \\
\thinrule
Factory \& production line work & Manufacturing \& assembly & 321 & 1,955 & 307 & 409 \\
\thinrule
Facility cleaning crew & Cleaning, facilities \& commercial laundry & 321 & 2,564 & 736 & 897 \\
\thinrule
Restaurant, caf\'e \& bar staff work & Restaurants \& kitchens & 320 & 3,094 & 594 & 710 \\
\thinrule
Farm \& grounds crew work & Agriculture \& farms & 319 & 2,097 & 406 & 507 \\
\thinrule
Pet grooming salon & Personal services & 318 & 2,147 & 262 & 281 \\
\thinrule
Appliance, bike \& shoe repair shop & Repair \& craft workshops & 318 & 1,503 & 361 & 437 \\
\thinrule
Commercial laundry floor & Cleaning, facilities \& commercial laundry & 317 & 1,864 & 439 & 473 \\
\thinrule
Warehouse, dock \& delivery work & Warehouses \& logistics & 267 & 2,820 & 307 & 417 \\
\thinrule
Print \& sign shop work & Textile \& apparel production & 234 & 1,873 & 232 & 289 \\
\thinrule
Electronics \& wiring assembly & Manufacturing \& assembly & 221 & 1,055 & 220 & 250 \\
\thinrule
Workshop making \& finishing & Repair \& craft workshops & 214 & 1,062 & 274 & 330 \\
\thinrule
Jewelry, watch, lock \& instrument repair & Repair \& craft workshops & 204 & 1,192 & 297 & 310 \\
\thinrule
Laboratory bench work & Healthcare, care \& labs & 162 & 1,266 & 151 & 188 \\
\thinrule
Venue setup \& breakdown & Events \& venues & 150 & 844 & 113 & 148 \\
\thinrule
Hotel \& guesthouse staff work & Hotels \& accommodation & 129 & 1,391 & 395 & 435 \\
\thinrule
Carpentry, upholstery \& furniture making & Woodworking \& furniture & 107 & 331 & 118 & 138 \\
\thinrule
Hand sewing \& clothing repair & Textile \& craft arts & 70 & 334 & 220 & 231 \\
\thinrule
Laptop or PC upgrade & Technology \& computing & 70 & 458 & 177 & 228 \\
\thinrule
Appliance filter service & Deep cleaning & 70 & 540 & 405 & 406 \\
\thinrule
Wall mounting: frames, shelves, curtains \& blinds & Moving \& home setup & 70 & 801 & 306 & 315 \\
\thinrule
Door, hinge \& handle fixes & Repairs \& DIY & 69 & 505 & 293 & 296 \\
\thinrule
Lawn \& hedge care & Gardening \& agriculture & 69 & 471 & 221 & 224 \\
\thinrule
Sealant \& wall touch-ups & Repairs \& DIY & 69 & 445 & 147 & 158 \\
\thinrule
Home car wash \& interior detail & Cars \& bikes & 69 & 1,272 & 261 & 288 \\
\thinrule
Bike puncture, brake \& chain fixes & Cars \& bikes & 69 & 377 & 152 & 168 \\
\thinrule
Sofa, mattress or rug deep clean & Deep cleaning & 69 & 467 & 254 & 255 \\
\thinrule
Bulb, battery \& alarm swaps & Repairs \& DIY & 69 & 894 & 491 & 505 \\
\thinrule
Flat-pack furniture assembly \& take-down & Moving \& home setup & 69 & 445 & 153 & 162 \\
\thinrule
Repotting \& planting & Gardening \& agriculture & 69 & 508 & 314 & 316 \\
\thinrule
Kitchen appliance deep clean & Deep cleaning & 68 & 472 & 289 & 289 \\
\thinrule
Metalwork, welding \& machine shop & Metalwork \& welding & 62 & 410 & 83 & 94 \\
\thinrule
Checkout, bagging \& order packing & Retail \& supermarkets & 58 & 593 & 145 & 184 \\
\thinrule
Car checks \& small part swaps & Cars \& bikes & 47 & 709 & 302 & 328 \\
\thinrule
Home pool cleaning \& water test & Gardening \& agriculture & 45 & 265 & 116 & 116 \\
\thinrule
Gym \& sports facility upkeep & Gyms \& sports facilities & 42 & 266 & 103 & 120 \\
\thinrule
Veterinary clinic work & Healthcare, care \& labs & 42 & 411 & 98 & 116 \\
\thinrule
Sink, shower \& toilet fixes & Repairs \& DIY & 41 & 461 & 247 & 250 \\
\thinrule
Florist \& garden centre work & Florists \& garden centres & 41 & 252 & 67 & 73 \\
\thinrule
Spare tire change & Cars \& bikes & 29 & 275 & 96 & 108 \\
\thinrule
Butcher, fish, bakery \& market stalls & Retail \& supermarkets & 28 & 202 & 53 & 63 \\
\thinrule
Hose, sprinkler \& drip line setup & Gardening \& agriculture & 23 & 391 & 227 & 228 \\
\thinrule
Office, mailroom \& IT hardware work & Offices \& IT & 22 & 370 & 141 & 175 \\
\thinrule
Recycling \& waste sorting & Recycling \& waste & 14 & 133 & 46 & 51 \\
\thinrule
Solar, fiber \& telecom installs & Building trades (electrical, plumbing, HVAC) & 11 & 89 & 26 & 31 \\
\thinrule
Making the Bed & Daily chores & 4.1 & 59 & 29 & 29 \\
\thinrule
Sweeping Floors & Daily chores & 4.0 & 71 & 41 & 42 \\
\thinrule
Taking out the Trash & Daily chores & 2.4 & 56 & 35 & 35 \\
\thinrule
Feed your Pets & Daily chores & 0.8 & 21 & 14 & 14 \\
\thickrule
\end{tabular}
\end{table*}
 Reading it relies only on standard formats (MP4 with H.264, parquet, LeRobot v3, MCAP).

\dsq{Does the dataset contain data that might be considered confidential?}
The footage shows the inside of workplaces and homes. Segments in which private text was detected (documents, screens, cards, license plates) were cut out before release.

\dsq{Does the dataset contain data that, if viewed directly, might be offensive, insulting, threatening, or might otherwise cause anxiety?}
No. The footage shows manual work.

\dsq{Does the dataset identify any subpopulations (e.g., by age, gender)?}
No. The release contains no demographic attributes; contributors are identified only by a pseudonymous identifier.

\dsq{Does the dataset contain data that might be considered sensitive in any way?}
The data is video of people at work in their own workplace or home. It contains no location data, no audio and no contributor identity. Care tasks performed on another person, episodes with a detected minor and segments with private text were removed.

\subsection*{Collection Process}

\dsq{How was the data associated with each instance acquired?}
Video and sensor data were recorded by the contributor's phone. The task type is the named task the contributor was asked to record. Descriptions, instructions, action family, skill group, difficulty, setting type and on-task share were produced by the annotation model, and hand-visibility shares were estimated by the model panel of the curation pipeline.

\dsq{What mechanisms or procedures were used to collect the data?}
Contributors record a named, unscripted task with a head-mounted phone during their own work, in their own workplace or home (Section~\ref{sec:collection}). v1.0 was recorded on 56 phone models.

\dsq{If the dataset is a sample from a larger set, what was the sampling strategy?}
Curation followed by rebalancing: the pipeline keeps on-task segments, applies privacy filtering, and rebalances hours across task types.

\dsq{Who was involved in the data collection process and how were they compensated?}
12,227 Hub contributors recorded the data. Contributors are compensated by Hub for their recordings.

\dsq{Over what timeframe was the data collected?}
The recording date of every episode is given in the catalog (\texttt{recorded\_date}).

\dsq{Were any ethical review processes conducted?}
The dataset was collected by a company and not as an institutional human-subjects study. Each contributor agreed to the use of their recordings (see below), and the privacy measures of Section~\ref{sec:privacy} are applied to every episode before release.

\dsq{Did you collect the data from the individuals in question directly, or obtain it via third parties or other sources?}
Directly. Contributors recorded themselves at work; no third-party footage is included.

\dsq{Were the individuals in question notified about the data collection?}
Contributors made the recordings themselves for this purpose.

\dsq{Did the individuals in question consent to the collection and use of their data?}
Yes. Each contributor agreed that their recordings may be used to train and evaluate AI and robotics systems.

\dsq{Has an analysis of the potential impact of the dataset and its use on data subjects been conducted?}
The privacy measures of Section~\ref{sec:privacy} limit the impact on contributors and on other people present during recording. Section~\ref{sec:impact} discusses the broader impact.

\subsection*{Preprocessing, Cleaning and Labeling}

\dsq{Was any preprocessing, cleaning or labeling of the data done?}
Yes, in the stages of Section~\ref{sec:curation} (Fig.~\ref{fig:pipeline}): a head-mount check that removes footage not worn on the head, dense temporal segmentation that keeps the segments where the contributor's hands perform the task, hand-visibility gating, privacy detection and cutting (Section~\ref{sec:privacy}), episode assembly, integrity and synchronization checks, and rebalancing of hours across task types; human review decisions take precedence where available. Episodes recorded at 60 fps (15.9\%) are re-encoded to 30 fps; the others keep the phone's original video stream. Episodes are then annotated by the annotation model (Section~\ref{sec:taxonomy}).

\dsq{Was the ``raw'' data saved in addition to the preprocessed, cleaned or labeled data?}
The release keeps the phone's original video stream for 84.1\% of episodes and every native sensor sample of every episode in MCAP. The footage removed by curation and privacy filtering is not part of the release.

\dsq{Is the software that was used to preprocess, clean or label the data available?}
The curation and annotation pipeline is not released. Every statistic in this paper is computed from the released catalog and annotation tables, and the analysis script is available from the authors on request.

\subsection*{Uses}

\dsq{Has the dataset been used for any tasks already?}
This paper reports the statistical analysis of Section~\ref{sec:analysis} and no model training results.

\dsq{Is there a repository that links to any or all papers or systems that use the dataset?}
Not yet.

\dsq{What (other) tasks could the dataset be used for?}
Those of Section~\ref{sec:uses}: visual representation pretraining, hand and hand-object pose, action recognition, VLA pretraining and co-training, video prediction and world models, and visual-inertial odometry.

\dsq{Is there anything about the composition of the dataset or the way it was collected and preprocessed, cleaned or labeled that might impact future uses?}
See the answer on errors and sources of noise above. In addition, recorded time varies across contributors (the median contributor recorded 21.3 minutes), and task types were rebalanced, so their shares do not reflect how often the tasks occur in the world.

\dsq{Are there tasks for which the dataset should not be used?}
No restrictions beyond the license.

\subsection*{Distribution}

\dsq{Will the dataset be distributed to third parties outside of the entity on behalf of which the dataset was created?}
Yes. It is distributed publicly, free of charge. Access goes through a short form (name, organization, country, intended use) in which users agree to credit the dataset and may opt in to dataset updates.

\dsq{How will the dataset be distributed? Does the dataset have a digital object identifier (DOI)?}
Through the Hugging Face Hub at \url{https://huggingface.co/datasets/HubLabs/HubEgo-100K}, as a LeRobot v3 dataset with MCAP sensor archives, a per-episode catalog and annotation tables. Single episodes can be fetched without downloading the whole dataset. The dataset page is its reference address.

\dsq{When will the dataset be distributed?}
v1.0 is available on the dataset page.

\dsq{Will the dataset be distributed under a copyright or other intellectual property (IP) license, and/or under applicable terms of use (ToU)?}
Creative Commons Attribution 4.0 (CC BY 4.0, \url{https://creativecommons.org/licenses/by/4.0/}): free to use, share and adapt for any purpose, including commercially, with credit. Credit line: ``HubEgo-100K, Hub Labs (Hub Data, Inc.), \url{https://huggingface.co/datasets/HubLabs/HubEgo-100K}, CC BY 4.0''.

\dsq{Have any third parties imposed IP-based or other restrictions on the data associated with the instances?}
No. All recordings are first-party.

\dsq{Do any export controls or other regulatory restrictions apply to the dataset or to individual instances?}
None known.

\subsection*{Maintenance}

\dsq{Who will be supporting, hosting and maintaining the dataset?}
Hub Labs (Hub Data, Inc.) maintains the dataset; it is hosted on the Hugging Face Hub.

\dsq{How can the owner, curator or manager of the dataset be contacted?}
Through \url{https://hub.xyz}.

\dsq{Is there an erratum?}
No. Corrections are recorded in the release notes on the dataset page.

\dsq{Will the dataset be updated?}
Yes. Additional data is released in versioned increments toward the full 100,000 hours; each release keeps the same format, annotation schema and license, and release notes and per-version statistics are published on the dataset page.

\dsq{If the dataset relates to people, are there applicable limits on the retention of the data associated with the instances?}
The dataset is distributed without a fixed end date.

\dsq{Will older versions of the dataset continue to be supported, hosted and maintained?}
Releases are versioned, and the statistics in this paper refer to v1.0.

\dsq{If others want to extend, augment, build on or contribute to the dataset, is there a mechanism for them to do so?}
The CC BY 4.0 license allows anyone to adapt and build on the dataset with credit. Corrections and suggestions can be sent to the maintainers through the contact above.

\section{Task Types, Action Families and Skill Groups}
\label{app:taxonomy}

Tables~\ref{tab:families} and \ref{tab:skills} list the action families and skill groups of \hub{} v1.0 with their hours, and Table~\ref{tab:tasks} lists the 56 task types. All values are computed from the released tables.

\section{Data Fields}
\label{app:fields}

Two tables describe every episode, both keyed by \texttt{episode\_index}.

\paragraph{\texttt{catalog.parquet}.}
\begin{itemize}
\item \texttt{recording\_id}: pseudonymous identifier of the source recording, one continuous capture by one contributor; a recording yields one or more episodes.
\item \texttt{recording\_start\_s}, \texttt{recording\_end\_s}: where the episode starts and ends inside its recording, in seconds; \texttt{recording\_start\_s} gives the order of the episodes of a recording.
\item \texttt{segment\_index}: counter of the kept segments of a recording.
\item \texttt{part}: $[i, n]$ when one segment longer than 30 minutes was cut into $n$ equal parts ($i$ counts from 0); $[0, 1]$ or null when the segment was not split.
\item \texttt{task}, \texttt{category}: task type the contributor was asked to record (56 task types) and its work category (30 categories).
\item \texttt{contributor\_id}: pseudonymous contributor identifier, stable across episodes.
\item \texttt{recorded\_date}, \texttt{time\_of\_day}: date of recording, and morning, afternoon, evening or night in local time.
\item \texttt{device\_make}, \texttt{device\_model}, \texttt{os}: capture phone and operating system.
\item \texttt{lens}, \texttt{fov\_diag\_deg}: camera lens (ultra-wide or wide) and its diagonal field of view in degrees.
\item \texttt{frames}, \texttt{duration\_s}: number of video frames and episode length in seconds.
\item \texttt{video\_method}, \texttt{source\_fps}: stream copy of the phone's 30 fps video, or 60 fps video re-encoded to 30 fps.
\item \texttt{frames\_repeated}: frames duplicated to keep a constant 30 fps (re-encoded episodes only).
\item \texttt{median\_frame\_interval\_s}, \texttt{max\_frame\_interval\_s}, \texttt{frame\_gaps\_over\_1\_5x}: typical and maximum time between consecutive frames, and the number of intervals longer than 1.5 times the nominal interval.
\item \texttt{drift\_vs\_nominal\_fps\_s}: difference in seconds between the real duration and the frame count divided by 30.
\item \texttt{imu\_samples\_json}: number of native sensor samples per stream (accelerometer, gyroscope, magnetometer, device motion, heading, barometer).
\item \texttt{hands\_share}, \texttt{hands\_share\_both}: model-estimated share of sampled frames that show at least one hand, and both hands, of the contributor.
\item \texttt{video\_path}, \texttt{video\_bytes}: location and size of the episode video.
\item \texttt{mcap\_tar}, \texttt{mcap\_member}, \texttt{mcap\_offset}, \texttt{mcap\_bytes}: archive, member name, byte offset and size of the episode's native sensor file, so that one episode can be read without extracting the archive.
\end{itemize}

\paragraph{\texttt{meta/annotations.parquet}.}
\begin{itemize}
\item \texttt{description}: what the contributor does in the episode, in one sentence.
\item \texttt{task\_instruction}: step-level instruction for the episode, written as a sequence of actions.
\item \texttt{action\_family}: main type of hand action (35 families).
\item \texttt{skill\_group}: skill area of the task (18 groups).
\item \texttt{difficulty}: easy, medium or hard.
\item \texttt{environment}: setting type of the episode (11 types); in this paper, an environment is a distinct pair of contributor and setting type.
\item \texttt{on\_task\_share}: model-estimated share of the episode spent on the task, between 0 and 1.
\end{itemize}


\begin{thebibliography}{54}
\providecommand{\natexlab}[1]{#1}
\providecommand{\url}[1]{\texttt{#1}}
\expandafter\ifx\csname urlstyle\endcsname\relax
  \providecommand{\doi}[1]{doi: #1}\else
  \providecommand{\doi}{doi: \begingroup \urlstyle{rm}\Url}\fi

\bibitem[Bahl et~al.(2022)Bahl, Gupta, and Pathak]{bahl2022whirl}
Shikhar Bahl, Abhinav Gupta, and Deepak Pathak.
\newblock Human-to-robot imitation in the wild.
\newblock In \emph{Proceedings of Robotics: Science and Systems (RSS)}, 2022.

\bibitem[Bahl et~al.(2023)Bahl, Mendonca, Chen, Jain, and Pathak]{bahl2023vrb}
Shikhar Bahl, Russell Mendonca, Lili Chen, Unnat Jain, and Deepak Pathak.
\newblock Affordances from human videos as a versatile representation for
  robotics.
\newblock In \emph{Proceedings of the IEEE/CVF Conference on Computer Vision
  and Pattern Recognition (CVPR)}, 2023.

\bibitem[Belkhale et~al.(2023)Belkhale, Cui, and
  Sadigh]{belkhale2023dataquality}
Suneel Belkhale, Yuchen Cui, and Dorsa Sadigh.
\newblock Data quality in imitation learning.
\newblock In \emph{Advances in Neural Information Processing Systems
  (NeurIPS)}, 2023.

\bibitem[Black et~al.(2025{\natexlab{a}})Black, Brown, Darpinian, Dhabalia,
  Driess, Esmail, et~al.]{black2025pi05}
Kevin Black, Noah Brown, James Darpinian, Karan Dhabalia, Danny Driess, Adnan
  Esmail, et~al.
\newblock $\pi_{0.5}$: a {Vision-Language-Action} model with open-world
  generalization.
\newblock In \emph{Proceedings of the Conference on Robot Learning (CoRL)},
  volume 305 of \emph{Proceedings of Machine Learning Research},
  2025{\natexlab{a}}.

\bibitem[Black et~al.(2025{\natexlab{b}})Black, Brown, Driess, Esmail, Equi,
  Finn, Fusai, Groom, Hausman, Ichter, et~al.]{black2025pi0}
Kevin Black, Noah Brown, Danny Driess, Adnan Esmail, Michael Equi, Chelsea
  Finn, Niccolo Fusai, Lachy Groom, Karol Hausman, Brian Ichter, et~al.
\newblock $\pi_0$: A {Vision-Language-Action} flow model for general robot
  control.
\newblock In \emph{Proceedings of Robotics: Science and Systems (RSS)},
  2025{\natexlab{b}}.

\bibitem[Brohan et~al.(2023)Brohan, Brown, Carbajal, Chebotar, Dabis, Finn,
  et~al.]{brohan2023rt1}
Anthony Brohan, Noah Brown, Justice Carbajal, Yevgen Chebotar, Joseph Dabis,
  Chelsea Finn, et~al.
\newblock {RT-1}: Robotics transformer for real-world control at scale.
\newblock In \emph{Proceedings of Robotics: Science and Systems (RSS)}, 2023.

\bibitem[{Build AI}(2025{\natexlab{a}})]{buildai2025egocentric100k}
{Build AI}.
\newblock {Egocentric-100K}.
\newblock \url{https://huggingface.co/datasets/builddotai/Egocentric-100K},
  2025{\natexlab{a}}.
\newblock Hugging Face dataset, accessed November 2026.

\bibitem[{Build AI}(2025{\natexlab{b}})]{buildai2025egocentric10k}
{Build AI}.
\newblock {Egocentric-10K}.
\newblock \url{https://huggingface.co/datasets/builddotai/Egocentric-10K},
  2025{\natexlab{b}}.
\newblock Hugging Face dataset, accessed November 2026.

\bibitem[Cadene et~al.(2024)Cadene, Alibert, Soare, Gallouedec, Zouitine,
  Palma, Kooijmans, Aractingi, Shukor, Aubakirova, Russi, Capuano, Pascal,
  Choghari, Meftah, Ellerbach, Moss, and Wolf]{cadene2024lerobot}
Remi Cadene, Simon Alibert, Alexander Soare, Quentin Gallouedec, Adil Zouitine,
  Steven Palma, Pepijn Kooijmans, Michel Aractingi, Mustafa Shukor, Dana
  Aubakirova, Martino Russi, Francesco Capuano, Caroline Pascal, Jade Choghari,
  Khalil Meftah, Maxime Ellerbach, Jess Moss, and Thomas Wolf.
\newblock {LeRobot}: State-of-the-art machine learning for real-world robotics
  in {PyTorch}.
\newblock \url{https://github.com/huggingface/lerobot}, 2024.

\bibitem[Chi et~al.(2023)Chi, Feng, Du, Xu, Cousineau, Burchfiel, and
  Song]{chi2023diffusion}
Cheng Chi, Siyuan Feng, Yilun Du, Zhenjia Xu, Eric Cousineau, Benjamin
  Burchfiel, and Shuran Song.
\newblock Diffusion policy: Visuomotor policy learning via action diffusion.
\newblock In \emph{Proceedings of Robotics: Science and Systems (RSS)}, 2023.

\bibitem[Damen et~al.(2018)Damen, Doughty, Farinella, Fidler, Furnari, Kazakos,
  Moltisanti, Munro, Perrett, Price, and Wray]{damen2018epic}
Dima Damen, Hazel Doughty, Giovanni~Maria Farinella, Sanja Fidler, Antonino
  Furnari, Evangelos Kazakos, Davide Moltisanti, Jonathan Munro, Toby Perrett,
  Will Price, and Michael Wray.
\newblock Scaling egocentric vision: The {EPIC-KITCHENS} dataset.
\newblock In \emph{Proceedings of the European Conference on Computer Vision
  (ECCV)}, 2018.

\bibitem[Damen et~al.(2022)Damen, Doughty, Farinella, Furnari, Kazakos, Ma,
  Moltisanti, Munro, Perrett, Price, and Wray]{damen2022epic100}
Dima Damen, Hazel Doughty, Giovanni~Maria Farinella, Antonino Furnari,
  Evangelos Kazakos, Jian Ma, Davide Moltisanti, Jonathan Munro, Toby Perrett,
  Will Price, and Michael Wray.
\newblock Rescaling egocentric vision: Collection, pipeline and challenges for
  {EPIC-KITCHENS-100}.
\newblock \emph{International Journal of Computer Vision (IJCV)}, 130, 2022.

\bibitem[Darkhalil et~al.(2022)Darkhalil, Shan, Zhu, Ma, Kar, Higgins, Fidler,
  Fouhey, and Damen]{darkhalil2022visor}
Ahmad Darkhalil, Dandan Shan, Bin Zhu, Jian Ma, Amlan Kar, Richard Higgins,
  Sanja Fidler, David Fouhey, and Dima Damen.
\newblock {EPIC-KITCHENS VISOR} benchmark: {VI}deo segmentations and object
  relations.
\newblock In \emph{Advances in Neural Information Processing Systems (NeurIPS),
  Datasets and Benchmarks Track}, volume~35, 2022.

\bibitem[Dasari et~al.(2023)Dasari, Srirama, Jain, and
  Gupta]{dasari2023unbiased}
Sudeep Dasari, Mohan~Kumar Srirama, Unnat Jain, and Abhinav Gupta.
\newblock An unbiased look at datasets for visuo-motor pre-training.
\newblock In \emph{Proceedings of the Conference on Robot Learning (CoRL)},
  volume 229 of \emph{Proceedings of Machine Learning Research}, 2023.

\bibitem[Engel et~al.(2023)Engel, Somasundaram, Goesele, Sun, Gamino, Turner,
  Talattof, Yuan, Souti, Meredith, et~al.]{engel2023aria}
Jakob Engel, Kiran Somasundaram, Michael Goesele, Albert Sun, Alexander Gamino,
  Andrew Turner, Arjang Talattof, Arnie Yuan, Bilal Souti, Brighid Meredith,
  et~al.
\newblock Project {Aria}: A new tool for egocentric multi-modal {AI} research.
\newblock \emph{arXiv preprint arXiv:2308.13561}, 2023.

\bibitem[{Foxglove}(2026)]{mcap}
{Foxglove}.
\newblock {MCAP} format specification.
\newblock \url{https://mcap.dev/spec}, 2026.
\newblock Accessed November 2026.

\bibitem[Gebru et~al.(2021)Gebru, Morgenstern, Vecchione, Vaughan, Wallach,
  {Daum\'e III}, and Crawford]{gebru2021datasheets}
Timnit Gebru, Jamie Morgenstern, Briana Vecchione, Jennifer~Wortman Vaughan,
  Hanna Wallach, Hal {Daum\'e III}, and Kate Crawford.
\newblock Datasheets for datasets.
\newblock \emph{Communications of the ACM}, 64\penalty0 (12), December 2021.
\newblock \doi{10.1145/3458723}.

\bibitem[Gong et~al.(2023)Gong, Mohan, Dhingra, Bazin, Li, Wang, and
  Ranjan]{gong2023mmg}
Xinyu Gong, Sreyas Mohan, Naina Dhingra, Jean-Charles Bazin, Yilei Li,
  Zhangyang Wang, and Rakesh Ranjan.
\newblock {MMG-Ego4D}: Multimodal generalization in egocentric action
  recognition.
\newblock In \emph{Proceedings of the IEEE/CVF Conference on Computer Vision
  and Pattern Recognition (CVPR)}, 2023.

\bibitem[Grauman et~al.(2022)Grauman, Westbury, Byrne, Chavis, Furnari,
  Girdhar, Hamburger, Jiang, Liu, Liu, et~al.]{grauman2022ego4d}
Kristen Grauman, Andrew Westbury, Eugene Byrne, Zachary Chavis, Antonino
  Furnari, Rohit Girdhar, Jackson Hamburger, Hao Jiang, Miao Liu, Xingyu Liu,
  et~al.
\newblock {Ego4D}: Around the world in 3,000 hours of egocentric video.
\newblock In \emph{Proceedings of the IEEE/CVF Conference on Computer Vision
  and Pattern Recognition (CVPR)}, 2022.

\bibitem[Grauman et~al.(2024)Grauman, Westbury, Torresani, Kitani, Malik,
  Afouras, Ashutosh, Baiyya, Bansal, Boote, et~al.]{grauman2024egoexo4d}
Kristen Grauman, Andrew Westbury, Lorenzo Torresani, Kris Kitani, Jitendra
  Malik, Triantafyllos Afouras, Kumar Ashutosh, Vijay Baiyya, Siddhant Bansal,
  Bikram Boote, et~al.
\newblock {Ego-Exo4D}: Understanding skilled human activity from first- and
  third-person perspectives.
\newblock In \emph{Proceedings of the IEEE/CVF Conference on Computer Vision
  and Pattern Recognition (CVPR)}, 2024.

\bibitem[Hoque et~al.(2026)Hoque, Huang, Yoon, Sivapurapu, and
  Zhang]{hoque2026egodex}
Ryan Hoque, Peide Huang, David~J. Yoon, Mouli Sivapurapu, and Jian Zhang.
\newblock {EgoDex}: Learning dexterous manipulation from large-scale egocentric
  video.
\newblock In \emph{International Conference on Learning Representations
  (ICLR)}, 2026.

\bibitem[Jang et~al.(2021)Jang, Irpan, Khansari, Kappler, Ebert, Lynch,
  et~al.]{jang2021bcz}
Eric Jang, Alex Irpan, Mohi Khansari, Daniel Kappler, Frederik Ebert, Corey
  Lynch, et~al.
\newblock {BC-Z}: Zero-shot task generalization with robotic imitation
  learning.
\newblock In \emph{Proceedings of the Conference on Robot Learning (CoRL)},
  volume 164 of \emph{Proceedings of Machine Learning Research}, 2021.

\bibitem[Kareer et~al.(2025)Kareer, Patel, Punamiya, Mathur, Cheng, Wang,
  Hoffman, and Xu]{kareer2025egomimic}
Simar Kareer, Dhruv Patel, Ryan Punamiya, Pranay Mathur, Shuo Cheng, Chen Wang,
  Judy Hoffman, and Danfei Xu.
\newblock {EgoMimic}: Scaling imitation learning via egocentric video.
\newblock In \emph{Proceedings of the IEEE International Conference on Robotics
  and Automation (ICRA)}, 2025.

\bibitem[Kareer et~al.(2026)Kareer, Pertsch, Darpinian, Hoffman, Xu, Levine,
  et~al.]{pi2025humantransfer}
Simar Kareer, Karl Pertsch, James Darpinian, Judy Hoffman, Danfei Xu, Sergey
  Levine, et~al.
\newblock Emergence of human to robot transfer in vision-language-action
  models.
\newblock In \emph{Proceedings of Robotics: Science and Systems (RSS)}, 2026.

\bibitem[Khazatsky et~al.(2024)Khazatsky, Pertsch, Nair, Balakrishna, Dasari,
  Karamcheti, Nasiriany, Srirama, Chen, Ellis, et~al.]{khazatsky2024droid}
Alexander Khazatsky, Karl Pertsch, Suraj Nair, Ashwin Balakrishna, Sudeep
  Dasari, Siddharth Karamcheti, Soroush Nasiriany, Mohan~Kumar Srirama,
  Lawrence~Yunliang Chen, Kirsty Ellis, et~al.
\newblock {DROID}: A large-scale in-the-wild robot manipulation dataset.
\newblock In \emph{Proceedings of Robotics: Science and Systems (RSS)}, 2024.

\bibitem[Kim et~al.(2024)Kim, Pertsch, Karamcheti, Xiao, Balakrishna, Nair,
  Rafailov, Foster, Lam, Sanketi, Vuong, Kollar, Burchfiel, Tedrake, Sadigh,
  Levine, Liang, and Finn]{kim2024openvla}
Moo~Jin Kim, Karl Pertsch, Siddharth Karamcheti, Ted Xiao, Ashwin Balakrishna,
  Suraj Nair, Rafael Rafailov, Ethan Foster, Grace Lam, Pannag Sanketi, Quan
  Vuong, Thomas Kollar, Benjamin Burchfiel, Russ Tedrake, Dorsa Sadigh, Sergey
  Levine, Percy Liang, and Chelsea Finn.
\newblock {OpenVLA}: An open-source vision-language-action model.
\newblock In \emph{Proceedings of the Conference on Robot Learning (CoRL)},
  volume 270 of \emph{Proceedings of Machine Learning Research}, 2024.

\bibitem[Li et~al.(2023)Li, Liu, and Wu]{li2023egoego}
Jiaman Li, C.~Karen Liu, and Jiajun Wu.
\newblock Ego-body pose estimation via ego-head pose estimation.
\newblock In \emph{Proceedings of the IEEE/CVF Conference on Computer Vision
  and Pattern Recognition (CVPR)}, 2023.

\bibitem[Li et~al.(2018)Li, Liu, and Rehg]{li2018egtea}
Yin Li, Miao Liu, and James~M. Rehg.
\newblock In the eye of beholder: Joint learning of gaze and actions in first
  person video.
\newblock In \emph{Proceedings of the European Conference on Computer Vision
  (ECCV)}, 2018.

\bibitem[Lin et~al.(2025)Lin, Hu, Sheng, Wen, You, and Gao]{lin2025scaling}
Fanqi Lin, Yingdong Hu, Pingyue Sheng, Chuan Wen, Jiacheng You, and Yang Gao.
\newblock Data scaling laws in imitation learning for robotic manipulation.
\newblock In \emph{International Conference on Learning Representations
  (ICLR)}, 2025.

\bibitem[Liu et~al.(2022)Liu, Liu, Jiang, Lyu, Wan, Shen, Liang, Fu, Wang, and
  Yi]{liu2022hoi4d}
Yunze Liu, Yun Liu, Che Jiang, Kangbo Lyu, Weikang Wan, Hao Shen, Boqiang
  Liang, Zhoujie Fu, He~Wang, and Li~Yi.
\newblock {HOI4D}: A {4D} egocentric dataset for category-level human-object
  interaction.
\newblock In \emph{Proceedings of the IEEE/CVF Conference on Computer Vision
  and Pattern Recognition (CVPR)}, 2022.

\bibitem[Long et~al.(2026)Long, Ma, Chu, Wu, Cao, Ye,
  et~al.]{long2026egoguidance}
Lin Long, Tianren Ma, Zhong Chu, Di~Wu, Maosong Cao, Qixiang Ye, et~al.
\newblock {Ego Guidance}: Enhancing robot policy learning with action-free
  human egocentric videos.
\newblock arXiv preprint arXiv:2611.13685, 2026.

\bibitem[Luo et~al.(2026)Luo, Feng, Zhang, Zheng, Wang, Yuan,
  et~al.]{luo2025beingh0}
Hao Luo, Yicheng Feng, Wanpeng Zhang, Sipeng Zheng, Ye~Wang, Haoqi Yuan, et~al.
\newblock {Being-H0}: Vision-language-action pretraining from large-scale human
  videos.
\newblock In \emph{Proceedings of the International Conference on Machine
  Learning (ICML)}, 2026.

\bibitem[Majumdar et~al.(2023)Majumdar, Yadav, Arnaud, Ma, Chen, Silwal, Jain,
  Berges, Wu, Vakil, Abbeel, Malik, Batra, Lin, Maksymets, Rajeswaran, and
  Meier]{majumdar2023vc1}
Arjun Majumdar, Karmesh Yadav, Sergio Arnaud, Yecheng~Jason Ma, Claire Chen,
  Sneha Silwal, Aryan Jain, Vincent-Pierre Berges, Tingfan Wu, Jay Vakil,
  Pieter Abbeel, Jitendra Malik, Dhruv Batra, Yixin Lin, Oleksandr Maksymets,
  Aravind Rajeswaran, and Franziska Meier.
\newblock Where are we in the search for an artificial visual cortex for
  embodied intelligence?
\newblock In \emph{Advances in Neural Information Processing Systems
  (NeurIPS)}, 2023.

\bibitem[Mandlekar et~al.(2021)Mandlekar, Xu, Wong, Nasiriany, Wang, Kulkarni,
  et~al.]{mandlekar2021robomimic}
Ajay Mandlekar, Danfei Xu, Josiah Wong, Soroush Nasiriany, Chen Wang, Rohun
  Kulkarni, et~al.
\newblock What matters in learning from offline human demonstrations for robot
  manipulation.
\newblock In \emph{Proceedings of the Conference on Robot Learning (CoRL)},
  volume 164 of \emph{Proceedings of Machine Learning Research}, 2021.

\bibitem[Moon et~al.(2023)Moon, Madotto, Lin, Saraf, Bearman, and
  Damavandi]{moon2023imu2clip}
Seungwhan Moon, Andrea Madotto, Zhaojiang Lin, Aparajita Saraf, Amy Bearman,
  and Babak Damavandi.
\newblock {IMU2CLIP}: Language-grounded motion sensor translation with
  multimodal contrastive learning.
\newblock In \emph{Findings of the Association for Computational Linguistics:
  EMNLP 2023 (Findings of EMNLP)}, 2023.
\newblock \doi{10.18653/v1/2023.findings-emnlp.883}.

\bibitem[Nair et~al.(2022)Nair, Rajeswaran, Kumar, Finn, and
  Gupta]{nair2022r3m}
Suraj Nair, Aravind Rajeswaran, Vikash Kumar, Chelsea Finn, and Abhinav Gupta.
\newblock {R3M}: A universal visual representation for robot manipulation.
\newblock In \emph{Proceedings of the Conference on Robot Learning (CoRL)},
  volume 205 of \emph{Proceedings of Machine Learning Research}, 2022.

\bibitem[{NVIDIA} et~al.(2025){NVIDIA}, Bjorck, Casta{\~n}eda, Cherniadev, Da,
  Ding, et~al.]{nvidia2025gr00t}
{NVIDIA}, Johan Bjorck, Fernando Casta{\~n}eda, Nikita Cherniadev, Xingye Da,
  Runyu Ding, et~al.
\newblock {GR00T} {N1}: An open foundation model for generalist humanoid
  robots.
\newblock arXiv preprint arXiv:2503.14734, 2025.

\bibitem[{Open X-Embodiment Collaboration}(2024)]{oxe2024}
{Open X-Embodiment Collaboration}.
\newblock Open {X}-embodiment: Robotic learning datasets and {RT-X} models.
\newblock In \emph{Proceedings of the IEEE International Conference on Robotics
  and Automation (ICRA)}, 2024.

\bibitem[Pavlakos et~al.(2024)Pavlakos, Shan, Radosavovic, Kanazawa, Fouhey,
  and Malik]{pavlakos2024hamer}
Georgios Pavlakos, Dandan Shan, Ilija Radosavovic, Angjoo Kanazawa, David
  Fouhey, and Jitendra Malik.
\newblock Reconstructing hands in {3D} with transformers.
\newblock In \emph{Proceedings of the IEEE/CVF Conference on Computer Vision
  and Pattern Recognition (CVPR)}, 2024.

\bibitem[Punamiya et~al.(2026)Punamiya, Kareer, Liu, Citron, Qiu, Cai,
  Gavryushin, Chen, Liconti, Zhu, et~al.]{egoverse2026}
Ryan Punamiya, Simar Kareer, Zeyi Liu, Joshua Citron, Ri-Zhao Qiu, Xiongyi Cai,
  Alexey Gavryushin, Jiaqi Chen, Davide Liconti, Lawrence~Y. Zhu, et~al.
\newblock {EgoVerse}: An egocentric human dataset for robot learning from
  around the world.
\newblock In \emph{Proceedings of Robotics: Science and Systems (RSS)}, 2026.
\newblock \doi{10.15607/RSS.2026.XXII.092}.

\bibitem[Qin et~al.(2018)Qin, Li, and Shen]{qin2018vinsmono}
Tong Qin, Peiliang Li, and Shaojie Shen.
\newblock {VINS-Mono}: A robust and versatile monocular visual-inertial state
  estimator.
\newblock \emph{IEEE Transactions on Robotics (T-RO)}, 34\penalty0 (4), 2018.
\newblock \doi{10.1109/TRO.2018.2853729}.

\bibitem[Qin et~al.(2022)Qin, Wu, Liu, Jiang, Yang, Fu, and Wang]{qin2022dexmv}
Yuzhe Qin, Yueh-Hua Wu, Shaowei Liu, Hanwen Jiang, Ruihan Yang, Yang Fu, and
  Xiaolong Wang.
\newblock {DexMV}: Imitation learning for dexterous manipulation from human
  videos.
\newblock In \emph{Proceedings of the European Conference on Computer Vision
  (ECCV)}, 2022.

\bibitem[Qiu et~al.(2025)Qiu, Yang, Cheng, Chawla, Li, He,
  et~al.]{qiu2025humanoid}
Ri-Zhao Qiu, Shiqi Yang, Xuxin Cheng, Chaitanya Chawla, Jialong Li, Tairan He,
  et~al.
\newblock Humanoid policy {$\sim$} human policy.
\newblock In \emph{Proceedings of the Conference on Robot Learning (CoRL)},
  volume 305 of \emph{Proceedings of Machine Learning Research}, 2025.

\bibitem[Radosavovic et~al.(2022)Radosavovic, Xiao, James, Abbeel, Malik, and
  Darrell]{radosavovic2022mvp}
Ilija Radosavovic, Tete Xiao, Stephen James, Pieter Abbeel, Jitendra Malik, and
  Trevor Darrell.
\newblock Real-world robot learning with masked visual pre-training.
\newblock In \emph{Proceedings of the Conference on Robot Learning (CoRL)},
  volume 205 of \emph{Proceedings of Machine Learning Research}, 2022.

\bibitem[Romero et~al.(2017)Romero, Tzionas, and Black]{romero2017mano}
Javier Romero, Dimitrios Tzionas, and Michael~J. Black.
\newblock Embodied hands: Modeling and capturing hands and bodies together.
\newblock \emph{ACM Transactions on Graphics (TOG), Proceedings of SIGGRAPH
  Asia}, 36\penalty0 (6), 2017.
\newblock \doi{10.1145/3130800.3130883}.

\bibitem[Sener et~al.(2022)Sener, Chatterjee, Shelepov, He, Singhania, Wang,
  and Yao]{sener2022assembly101}
Fadime Sener, Dibyadip Chatterjee, Daniel Shelepov, Kun He, Dipika Singhania,
  Robert Wang, and Angela Yao.
\newblock {Assembly101}: A large-scale multi-view video dataset for
  understanding procedural activities.
\newblock In \emph{Proceedings of the IEEE/CVF Conference on Computer Vision
  and Pattern Recognition (CVPR)}, 2022.

\bibitem[Shan et~al.(2020)Shan, Geng, Shu, and Fouhey]{shan2020hands}
Dandan Shan, Jiaqi Geng, Michelle Shu, and David~F. Fouhey.
\newblock Understanding human hands in contact at internet scale.
\newblock In \emph{Proceedings of the IEEE/CVF Conference on Computer Vision
  and Pattern Recognition (CVPR)}, 2020.

\bibitem[Srirama et~al.(2024)Srirama, Dasari, Bahl, and Gupta]{srirama2024hrp}
Mohan~Kumar Srirama, Sudeep Dasari, Shikhar Bahl, and Abhinav Gupta.
\newblock {HRP}: Human affordances for robotic pre-training.
\newblock In \emph{Proceedings of Robotics: Science and Systems (RSS)}, Delft,
  Netherlands, 2024.
\newblock \doi{10.15607/RSS.2024.XX.068}.

\bibitem[Tao et~al.(2025)Tao, Srirama, Liu, Shaw, and Pathak]{tao2025dexwild}
Tony Tao, Mohan~Kumar Srirama, Jason~Jingzhou Liu, Kenneth Shaw, and Deepak
  Pathak.
\newblock {DexWild}: Dexterous human interactions for in-the-wild robot
  policies.
\newblock In \emph{Proceedings of Robotics: Science and Systems (RSS)}, 2025.

\bibitem[Wang et~al.(2023{\natexlab{a}})Wang, Fan, Sun, Zhang, Fei-Fei, Xu,
  et~al.]{wang2023mimicplay}
Chen Wang, Linxi Fan, Jiankai Sun, Ruohan Zhang, Li~Fei-Fei, Danfei Xu, et~al.
\newblock {MimicPlay}: Long-horizon imitation learning by watching human play.
\newblock In \emph{Proceedings of the Conference on Robot Learning (CoRL)},
  volume 229 of \emph{Proceedings of Machine Learning Research},
  2023{\natexlab{a}}.

\bibitem[Wang et~al.(2023{\natexlab{b}})Wang, Kwon, Rad, Pan, Chakraborty,
  Andrist, Bohus, Feniello, Tekin, Frujeri, Joshi, and
  Pollefeys]{wang2023holoassist}
Xin Wang, Taein Kwon, Mahdi Rad, Bowen Pan, Ishani Chakraborty, Sean Andrist,
  Dan Bohus, Ashley Feniello, Bugra Tekin, Felipe~Vieira Frujeri, Neel Joshi,
  and Marc Pollefeys.
\newblock {HoloAssist}: an egocentric human interaction dataset for interactive
  {AI} assistants in the real world.
\newblock In \emph{Proceedings of the IEEE/CVF International Conference on
  Computer Vision (ICCV)}, 2023{\natexlab{b}}.

\bibitem[Xie et~al.(2024)Xie, Lee, Xiao, and Finn]{xie2024decomposing}
Annie Xie, Lisa Lee, Ted Xiao, and Chelsea Finn.
\newblock Decomposing the generalization gap in imitation learning for visual
  robotic manipulation.
\newblock In \emph{Proceedings of the IEEE International Conference on Robotics
  and Automation (ICRA)}, 2024.

\bibitem[Yang et~al.(2025)Yang, Yu, Wu, Yan, Li, Cheng, et~al.]{yang2025egovla}
Ruihan Yang, Qinxi Yu, Yecheng Wu, Rui Yan, Borui Li, An-Chieh Cheng, et~al.
\newblock {EgoVLA}: Learning vision-language-action models from egocentric
  human videos.
\newblock arXiv preprint arXiv:2507.12440, 2025.

\bibitem[Ye et~al.(2025)Ye, Jang, Jeon, Joo, Yang, Peng, et~al.]{ye2025lapa}
Seonghyeon Ye, Joel Jang, Byeongguk Jeon, Sejune Joo, Jianwei Yang, Baolin
  Peng, et~al.
\newblock Latent action pretraining from videos.
\newblock In \emph{International Conference on Learning Representations
  (ICLR)}, 2025.

\end{thebibliography}


\footnotesize
\end{document}